\documentclass[11pt]{article}

\PassOptionsToPackage{square,numbers}{natbib}
\usepackage[preprint]{neurips_2023}

\usepackage[utf8]{inputenc} % allow utf-8 input
\usepackage[T1]{fontenc}    % use 8-bit T1 fonts
\usepackage{hyperref}       % hyperlinks
\usepackage{url}            % simple URL typesetting
\usepackage{booktabs}       % professional-quality tables
\usepackage{amsfonts}       % blackboard math symbols
\usepackage{nicefrac}       % compact symbols for 1/2, etc.
\usepackage{microtype}      % microtypography
\usepackage{xcolor}         % colors

\usepackage{amsmath}
\usepackage{amsthm}
\usepackage{acronym}
\usepackage{array}
\usepackage{wrapfig}
\usepackage{multirow}
\usepackage{tabu}

\usepackage[ruled]{algorithm2e}
\usepackage{microtype}  % For better text fitting
\usepackage{etoolbox}   % For patching commands

\SetAlFnt{\small}       % Smaller font in algorithms
\SetAlCapFnt{\small}    % Smaller font for captions
\SetNlSty{textbf}{}{:}  % Numbering style

\makeatletter
\patchcmd{\itemize}{\parsep4\p@}{\parsep1\p@}{}{}
\patchcmd{\itemize}{\leftmargin\leftmargini}{\leftmargin0.8\leftmargini}{}{}
\makeatother

\theoremstyle{plain}
\newtheorem{theorem}{Theorem}
\newtheorem{lemma}[theorem]{Lemma}

\theoremstyle{definition}
\newtheorem{definition}{Definition}

\usepackage{graphicx}
\usepackage{caption}

\usepackage{float}
\usepackage{subcaption}
\usepackage{morefloats}
\usepackage{placeins}

\usepackage{comment}

\usepackage{listings}
\usepackage{color}
 
\definecolor{codegreen}{rgb}{0,0.6,0}
\definecolor{codegray}{rgb}{0.5,0.5,0.5}
\definecolor{codepurple}{rgb}{0.58,0,0.82}
\definecolor{backcolour}{rgb}{0.95,0.95,0.92}
 
\usepackage[colorinlistoftodos]{todonotes}

\title{
ETHER: Aligning Emergent Communication for Hindsight Experience Replay
}

\author{%
  Kevin Denamganaï\\
  University of York\\
  \texttt{\small kevin.denamganai@york.ac.uk} \\
  \And
  Daniel Hernandez\\
  Sony AI\\
  \texttt{\small daniel.hernandez@sony.com}\\
  \AND 
  Ozan Vardal\\
  Fudan University\\
  \texttt{\small ozanvardal@fudan.edu.cn}\\
  \And
  Sondess Missaoui\\
  University of York\\
  \texttt{\small sondess.missaoui@york.ac.uk}\\
  \And 
  James Alfred Walker \\
  University of York\\
  \texttt{\small james.walker@york.ac.uk} \\
}

\begin{document}

\maketitle

%\listoftodos

\begin{abstract}

    % Old
    % Hindsight Experience Replay (HER) enhances sample efficiency in goal-conditioned reinforcement learning (RL) by relabelling failed trajectories with goals that were actually achieved. However, HER assumes access to a goal relabelling function and a predicate function that determines whether a goal has been satisfied. These assumptions break down in instruction-following tasks, where goals are expressed in natural language and differ from the state space. We formalize this as the Hindsight Reinforcement Learning problem, which shows the need to jointly learn these functions alongside the RL policy. To address it, we propose ETHER (Emergent Textual Hindsight Experience Replay), an agent that leverages Emergent Communication. ETHER uses a referential game to train a speaker and a listener to develop a grounded, artificial language describing environment states. It then aligns this emergent language with instruction language using co-occurrence patterns between task-relevant objects and visual observations. In experiments on BabyAI’s PickupDist task, we show that ETHER’s learned speaker and listener can function as the goal relabelling and predicate components of HER, improving sample efficiency despite imperfect language alignment. Our work bridges Emergent Communication and goal-conditioned RL, expanding HER’s applicability to natural language instruction settings.

Hindsight Experience Replay (HER) enhances sample efficiency in goal-conditioned reinforcement learning (RL) by relabelling failed trajectories with goals that were actually achieved. 
However, HER implicitly assumes access to a goal relabelling function and a predicate function that determines whether a goal has been satisfied. 
These assumptions break down in instruction-following tasks, where goals are expressed in natural language and differ from the state space. 
We formalize this as the Hindsight Reinforcement Learning problem, which shows the need to jointly learn these functions alongside the RL policy. 
To address it, we propose ETHER (Emergent Textual Hindsight Experience Replay), an agent that leverages Emergent Communication to learn the goal-relabelling and predicate functions. 
ETHER uses a referential game (RG) to train a speaker and a listener to develop a grounded, artificial language describing environment states. 
It partially aligns this emergent language with instruction language using co-occurrence patterns between task instructions and RL observations. 
We prove that the relabelling and predicate functions that ETHER derives from the RG avoid the degenerate solutions of the Hindsight RL problem, namely trivial predicates and collapsed relabelling functions. 
Experiments on BabyAI’s PickupDist task show that ETHER’s learned RG speaker and listener can function as the goal relabelling and predicate functions of HER, improving sample efficiency despite imperfect language alignment. 
Our work bridges Emergent Communication and goal-conditioned RL, opening the door to wider applications of HER. 
    
\end{abstract}
%Then, the listener agent learns in an unsupervised fashion a predicate function that can be used in place of the oracle predicate function that HiGHER relied on.

\section{Introduction}
\label{sec:introduction}

Natural languages have been harnessed by humans as powerful tools to describe not only reality as one perceives it, but also as one imagines it (e.g. via the poetic function of languages~\cite{jakobson1960linguistics}). 
%Through properties such as compositionality and recursive syntax natural languages become flexible interfaces that allow humans to express arbitrarily complex meanings. 
%Beyond being immensely useful for inter-human communication, natural languages can also be a fruitful means of communication between humans and AI models, as recently showed by the advent of large language models \cite{touvron2023llama}.
Due to its flexibility, the use of natural languages to condition the behaviour of reinforcement learning (RL) agents remains an enticing research question with an untapped potential~\cite{luketina2019survey}. 
This endeavour began in goal-conditioned RL, which asks how to train RL agents capable of achieving arbitrary goals.
Natural languages become appealing in order to efficiently communicate such arbitrary goals to RL agents. 
Thus, (natural) language-conditioned RL addresses the challenge of training agents to attain a broad array of objectives, utilizing natural languages as a flexible tool to define those objectives as instructions/goals.

% However, similarly to goal-conditioned RL agents, language-conditioned RL agents suffer from low sample-efficiency during training due to sparse rewards. This issue also negatively scales with the number of goals. % since they also suffer from low-generalisation capabilities
A fundamental problem in RL is that of training agents in sparse reward settings, with goal-conditioned RL being no exception. In order to improve the sample-efficiency of goal-conditioned RL agents, Hindsight Experience Replay (HER) \cite{andrychowicz2017hindsight}) proposes to augment the dataset of harvested RL trajectories by relabelling unsuccessful goal-conditioned trajectories with alternative goals \textbf{that are known to be fulfilled in the last state of said unsuccessful trajectories}.
% The relabelling also involves re-computing rewards along the said trajectories, directly tackling sparse rewards by manually changing observed rewards. This trick yields an augmented RL replay buffer with much more feedback (non-null rewards) for a goal-conditioned agent to train on, thus improving the sample-efficiency of the resulting agent's training.
HER's relabelling injects feedback on an agent's replay buffer (non-null rewards) by re-computing rewards along said trajectory, directly tackling the sparse reward problem. 
%For the purpose of goal-relabelling and reward re-computing, HER uses both a mapping/goal-relabelling function and a predicate function that rely on some oracles and expert supervision.
Yet HER relies on two limiting assumptions.
Firstly, it assumes the goal space and state space to be the same, thus preventing its application to, for instance, instruction-following/language-conditioned environments where the goal space consists of natural language instructions in a robotics navigation domain.
In such environments, HER would require a goal-relabelling function that returns a fulfilled goal when provided an RL state/trajectory as input.
Secondly, HER assumes access to an oracle predicate function that specifies whether a given goal/instruction has been fulfilled in a given RL trajectory/state. 

%\todo[]{move the commented elements below to the Related Work appendix, maybe?}
% \citet{Chan2019-ACTRCE}'s ACTRCE improves upon HER by mainly proposing language-conditioning as a way to introduce more flexibility into goal specification, but the limitations remain.
% Some of those limitations are addressed by HIGhER~\citep{cideron2020higher}, which improves upon HER by, firstly, and similarly to ACTRCE, also proposing language-conditioning, and, secondly, by proposing to learn the goal-relabelling function using successful trajectories in parallel of the RL policy training. 
% HIGhER still suffers from reliance on an oracle predicate function and some early/random successful trajectories being harvested in order to bootstrap its learning of the goal-relabelling function. 

This work tackles the challenges of language-conditioned RL and the limitations of HER (and subsequent improvements of ACTRCE~\citep{Chan2019-ACTRCE}' and HIGhER~\citep{cideron2020higher}). 
We present Emergent Textual Hindsight Experience Replay (ETHER). 
%ETHER learns both the goal-relabelling function and the predicate function in the form of, respectively, a speaker and a listener agent in Referential Games, a common framework for Emergent Communication.
ETHER learns both the goal-relabelling and predicate functions in parallel with policy learning by using two unsupervised auxiliary tasks inspired by Emergent Communication (EmeCom).
Firstly, ETHER uses a discriminative visual referential game (RG -~\cite{lewis1969convention}) that trains a speaker and a listener to communicate about and discriminate between states/observations of an environment using an emergent artificial language. 
Secondly, ETHER aligns this emergent language with the (synthetic) natural language from the tasks instructions/goals.

This is performed by exploiting semantic co-occurrences between objects appearing in the tasks' goal/instructions, through their referring words, and objects appearing in the agent's observations, through their state representation.
Thus, we show experimentally in the BabyAI's PickupDist task that the emergent languages created by the speaker and listener agents can be aligned to some extent with the natural-like language of the tasks, and therefore allowing both agents to be used, respectively, as goal-relabelling and predicate functions of HER.
Despite the alignment being imperfect, the sample-efficiency of ETHER still improves over baselines.
This alignment property is at the core of the ability of ETHER to not necessitate any oracle or expert supervision.
%We show that the speaker and listener agents of the referential game make an artificial language emerge that is aligned with the natural-like language used to describe goals in the BabyAI benchmark and that it is expressive enough so as to also describe unsuccessful RL trajectories and thus provide feedback to the RL agent to leverage the linguistic, structured information contained in all trajectories.

This paper makes three contributions. 
Firstly, in Section~\ref{sec:background}, we further the understanding of the limitations of HER, highlighting critical components that previous works had left implicit, and we leverage the resulting insights into a novel problem formulation, referred to as the \textbf{Hindsight RL (HRL) optimisation problem}, emphasising the importance of learning both the goal-relabelling and predicate functions in parallel of the RL policy, as detailed in Section~\ref{sec:hrl-problem-formulation}.
Secondly, we propose \textbf{ETHER} to address this novel problem formulation, using two unsupervised auxiliary tasks~\citep{jaderberg2016UNREAL} inspired from EmeCom, as detailed in Section~\ref{sec:method}. %thesis:ether:method} and Section~\ref{ether:sec:rg-as-unsupervised-rl-task}.
Moreover, we prove in Theorem~\ref{thm:non_degeneracy} that the goal-relabelling and predicate functions that ETHER derives from a descriptive RG are guaranteed to avoid the degenerate solutions of the HRL problem, so that the trivial predicate and collapsed relabelling cases identified in Section~\ref{sec:hrl-problem-formulation} are ruled out by construction rather than by tuning. 
%unlike HER/ACTRCE/HIGhER that learns solely from successful RL trajectories, we show how ETHER can leverage both successful and unsuccessful trajectories towards acquiring relevant learned representations.
%and aligning its internal emergent language with the (natural) language used to define the language-conditioned tasks. 
Finally, considering the emergent language at the core of the ETHER architecture, we provide insights about the problem of aligning the emergent language with the (synthetic) natural-like languages that is used by the environment to specify instructions/goals.
We provide experimental evidence in Section~\ref{sec:experiments} that it is possible to perform this alignment to some extent by leveraging the semantic co-occurrences of objects between the agent's observations and its task' instructions/goals.

%\todo[]{maybe put this below-commented element as conclusion rather than trying to open in the introduction}
% Taken altogether, we conclude in Section~\ref{ether:sec:conclusion} about our work showing that EmeCom provides viable unsupervised auxiliary tasks for goal-conditioned RL, and opens the door to making HER more widely applicable if the alignment between the emergent and natural-like task languages can be improved.

% \begin{figure}[th]
%     \centering
%     \includegraphics[width=1.0\linewidth]{figures/ether-arch.png}
%     \label{fig:ether}
%     \caption{
%     ETHER architecture: all of the agent trajectories are added to the RL replay buffer and the goal mapper dataset. Failed trajectories are re-labelled by the instruction generator and both original and relabaled trajectories are appended to the RL replay buffer. 
%     }
% \end{figure}

% \begin{figure}[th]
%     \centering
%     \includegraphics[width=1.0\linewidth]{figures/ETHER-Discriminative-RG.png}
%     \label{fig:ether-rg}
%     \caption{
%     Visual Discriminative Referential Game with $K$ distractor stimuli and using a Straight-Through Gumbel-Softmax (STGS) communication channel and the STGS-LazImpa loss function. Stimuli are passed through a data augmentation scheme (e.g. adding Gaussian Blur, and/or Color Jitter, and/or undergoing some Affine Transformation, etc...) in order to enforce object-centricism. Note that the target stimulus undergoes different data augmentation depending on whether it is fed to the speaker agent or the listener agent. 
%     }
% \end{figure}

\section{Background}
\label{sec:background}

\begin{figure*}[t]
    \centering
    \begin{subfigure}[t]{0.45\linewidth}
        %\vspace{-100pt}
        \includegraphics[width=1.0\linewidth]{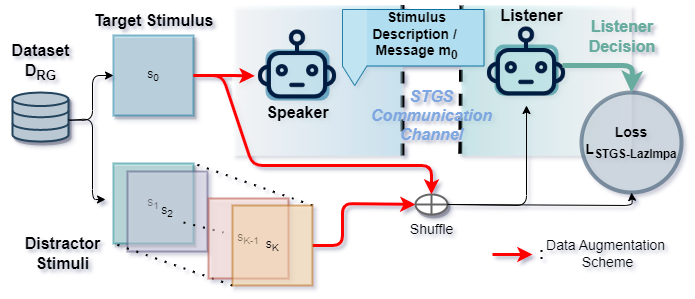}
        \caption{}
        \label{fig:ether-rg}
    \end{subfigure}
    \begin{subfigure}{0.26\textwidth}
        \centering
        \includegraphics[width=\textwidth]{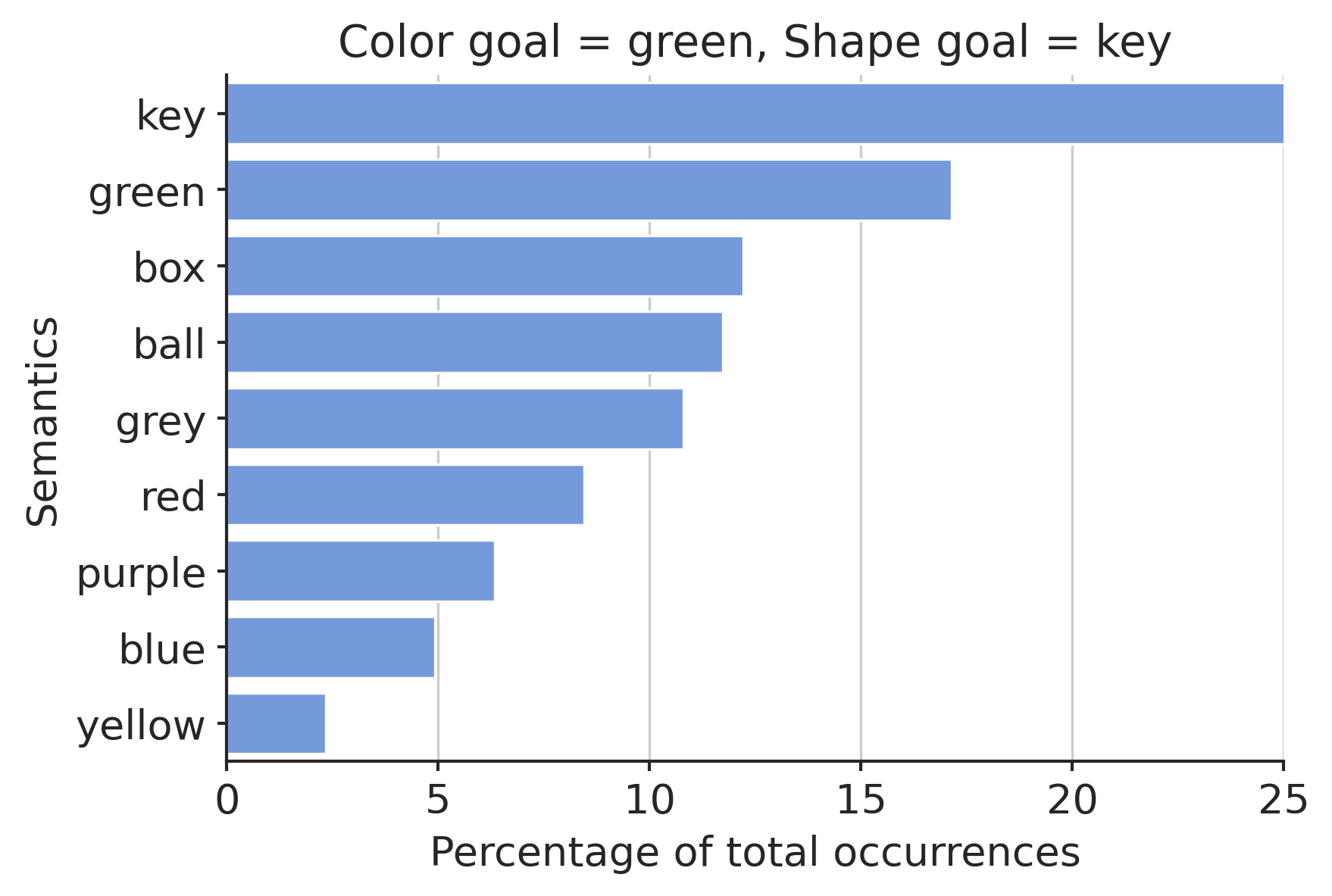}
        \caption{Expert agent}
        \label{fig:co_occurrence:expert}
    \end{subfigure}
    \begin{subfigure}{0.26\textwidth}
        \centering
        \includegraphics[width=\textwidth]{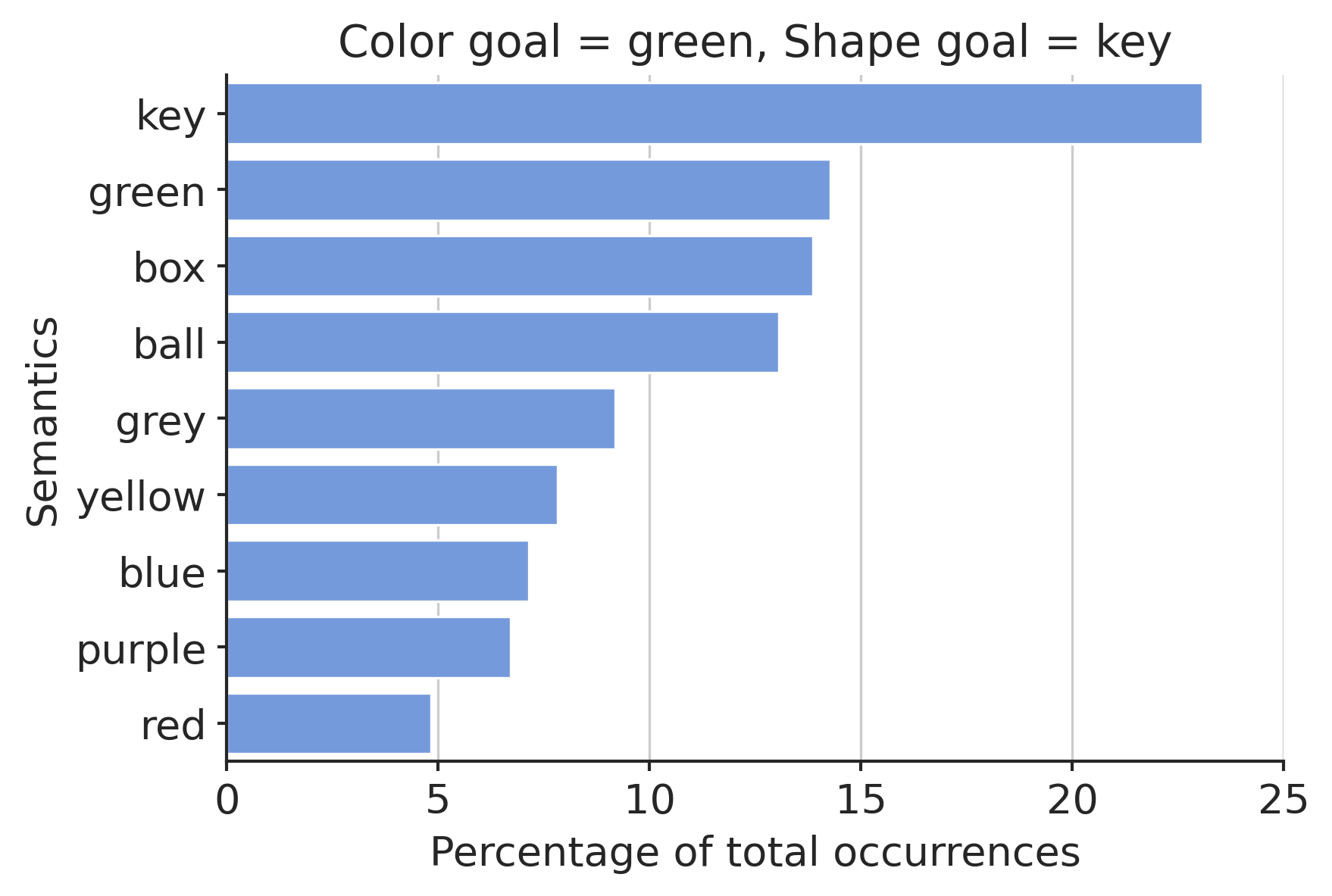}
        \caption{Random agent}
        \label{fig:co_occurrence:random}
    \end{subfigure}
    %\textbf{(a~\citep{Havrylov2017} and a loss function which adapts the STGS communication channel to LazImpa~\citep{rita2020lazimpa}, that we refer to as the STGS-LazImpa loss (cf. Appendix~\ref{subsec:stgs-lazimpa}). Stimuli are passed through the data augmentation scheme described in~\cite{Dessi2021-emecom_as_ssl} to enforce object-centricism \citep{Choi2018,DenamganaiAndWalker2020a}, which differ between s \textbf{(bc:} Histograms of observation semantic counts from goal conditioned trajectories from either BabyAI's built-in expert agent which always succeeds or a Random agent. In both cases the semantics of the goal are within the most observed features.% for any trajectory, which is less noticeable in the random agent.
    \caption{\textbf{(a):} Schema of a \textit{discriminative object-centric/2-player/$L$-signal/$N=0$-round/$K=64$-distractor} \textit{visual referential game} \citep{DenamganaiAndWalker2020a} using a Straight-Through Gumbel-Softmax (STGS) communication channel~\citep{Havrylov2017}, with a modified STGS loss adapted to LazImpa~\citep{rita2020lazimpa}, termed STGS-LazImpa (see Appendix~\ref{subsec:stgs-lazimpa}). Stimuli undergo data augmentation from~\cite{Dessi2021-emecom_as_ssl} to enforce object-centricism~\citep{Choi2018,DenamganaiAndWalker2020a}, differing between speaker and listener. \textbf{(b, c):} Histograms of semantic counts from goal-conditioned trajectories by BabyAI’s expert agent and a Random agent, showing that goal semantics align with the most frequently observed features.}

\end{figure*}

\subsection{Emergent Communication} 
\label{subsec:emecom}

Emergent Communication is at the interface of language grounding and language emergence. 
While language emergence raises the question of how to make artificial languages emerge possibly with similar properties to natural languages, such as  compositionality \citep{Baroni2019, Guo2019, Li&Bowling2019, Ren2020}, language grounding is concerned with the ability to ground the meaning of (natural) language utterances into some sensory processes, e.g. the  visual modality.  
Yet, emerging languages are far from being `natural-like' protolanguages \citep{Kottur2017,Chaabouni2019a,Chaabouni2019b}, and the questions of how to constraint them to a specific semantic or a specific syntax remain open problems, which we tackle in Section~\ref{subsec:co-occurrence}.

Stemming from the \textit{Signalling Game} or \textit{Referential Game (RG)} by \citet{lewis1969convention}, EmeCom studies commonly train a speaker agent to communicate efficiently with a listener agent about a given \textit{state/stimulus} of the world that it observes, so that the listener agent can then acts upon the observation of the message by choosing an \textit{action} that maximises a notion of utility or interest that is common to both players.
In discriminative RGs, typically, the listener action is to identify a target stimulus from a set of candidate stimuli, based solely on the speaker's message describing said target stimulus.
Following the nomenclature proposed in \citet{DenamganaiAndWalker2020a}, we will focus primarily on a \textit{descriptive object-centric (partially-observable) $2$-players/$L=10$-signal/$N=0$-round/$K=64$-distractor} RG variant, as illustrated in Figure~\ref{fig:ether-rg}, where the \textit{descriptive} aspect~\citep{Choi2018} implies that the target may not be part of the candidates observed by the listener.

\begin{definition}[Discriminative Referential Game]
\label{def:rg}
Let $\mathcal{D}_{RG}\subset \mathcal{S}$ be a finite dataset of stimuli and $\Sigma^\star$ be the set of all language utterances using vocabulary $\Sigma$. 
A discriminative RG with $K \geq 1$ distractors involves two agents, a speaker $Sp: \mathcal{S} \mapsto \Sigma^\star$ that maps stimuli to emergent language utterances, and a listener $Li: \mathcal{S} \times \Sigma^\star \mapsto \mathbb{R}$ that evaluates compatibility between stimuli and utterances.
At each round, a target stimulus $s_0$ and $K$ distractor stimuli $(s_1, \ldots, s_K)$ are drawn independently from $\mathcal{D}_{RG}$.
\end{definition}

The optimal speaker and listener functions are found solving:
\begin{comment}
\begin{align} 
Sp^\star, Li^\star &= \underset{Sp,Li}{\mathrm{argmin}} \,\, J_{RG}(Sp, Li) \\
\text{s.t. } \, \, &J_{RG}(Sp, Li) = \nonumber \\ 
%\mathop{\mathbb{E}}_{(s_0,\ldots,s_K)\sim\mathcal{D}^{K+1}} &\left[ 
%J_{RG}(Sp, Li) = \mathop{\mathbb{E}}_{(s_i)_{i\in[0,K]}\sim\mathcal{D}^{K+1}} &\left[ 
\mathop{\mathbb{E}}_{(s_i)_{i\in[0,K]}\sim\mathcal{D}^{K+1}} &\left[ 
\mathcal{L}( (Li(s_i, Sp(s_0)))_{i\in[0,K]} ) \right] 
\end{align}
\end{comment}
\begin{align}
Sp^\star, Li^\star &= \underset{Sp,Li}{\mathrm{argmin}} \,\, J_{RG}(Sp, Li) \\
\label{eq:j_rg}
\text{s.t. } \, \, J_{RG}(Sp, Li) &=
\mathop{\mathbb{E}}_{(s_i)_{i\in[0,K]}\sim\mathcal{D}_{RG}^{K+1}}  \left[
\mathcal{L}( (Li(s_i, Sp(s_0)))_{i\in[0,K]} ) \right]
\end{align}
% \begin{align} 
% Sp^\star, Li^\star &= \underset{Sp,Li}{\mathrm{argmin}} \,\, J_{RG}(Sp, Li) \\
% %\text{s.t. } \, \, 
% J_{RG}(Sp, Li) &= \mathbb{E}_{\substack{s_0\sim\mathcal{D} \\ (s_i)_{i\in[1,K]})\sim\mathcal{D}^K}}
% \left[ L( (Li(s_i, Sp(s_0)))_{i\in[0,K]} ) \right]
% \end{align}
where $\mathcal{L}$ is a loss function, such as Hinge loss. 
Writing $\mathcal{C}$ for the set of classes that the listener agent's decision module scores, $\text{score}(c)$ for the score it attributes to class $c\in\mathcal{C}$, and $c^\star\in\mathcal{C}$ for the correct class of the round, the Hinge loss reads:
\begin{equation}
\label{eq:hinge_loss}
\mathcal{L} = \max\Big(0, \; \nu - \text{score}(c^\star) + \underset{c\in\mathcal{C}\setminus\{c^\star\}}{\max} \, \text{score}(c) \Big)
\end{equation}
with margin hyperparameter $\nu$ (typically $1$). 
In a discriminative RG, $\mathcal{C}$ indexes the $K+1$ candidate stimuli, $\text{score}(i) = Li(s_i, Sp(s_0))$ with target stimulus $s_0$ and distractor stimuli $(s_1, \ldots, s_K)$, and $c^\star$ is the index of the target stimulus. 
Definition~\ref{def:descriptive_rg} extends $\mathcal{C}$ with one further class, which is what allows the listener agent to be used as an HER predicate function, as detailed below.
%For inference, scores are converted to probabilities via softmax:
%p(d_i|(d_j)_{j∈[0,K]}; l) = exp(Li(d_i, l))/∑_{j=0}^{K} exp(Li(d_j, l))
This optimization objective ensures the speaker generates discriminative utterances in the emergent language and the listener correctly identifies target stimuli from those utterances.
%forming the foundation for goal-relabelling and predicate functions in the ETHER approach.

It is critical for ETHER's design that we further require the RG to be \textbf{descriptive}~\citep{Choi2018,DenamganaiAndWalker2020a}, as detailed in Definition~\ref{def:descriptive_rg}.

\begin{definition}[Descriptive RG and no-target logit]
\label{def:descriptive_rg}
In a \textbf{descriptive RG}, the target stimulus may be absent from the candidates presented to the listener agent, and being replaced by a \textit{descriptive distractor}. 
RG rounds therefore come in two kinds: \textit{target-present} and \textit{target-absent}, and the listener's decision module outputs a $K+2$-logit distribution where the extra class has the meaning that none of the $K+1$ candidates is the target. 
Following \citet{Denamganai2023visual-COMPODIS}, that extra class is carried by a \textbf{learnable scalar} $\ell_\text{no-target}$, as a parameter of the model, such that writing $z(s,g) = Li(s,g)$ for the raw compatibility score:
\begin{equation}
\label{eq:no_target_softmax}
p\big( (d_i)_{i\in[0,K+1]} \mid (s_i)_{i\in[0,K]} ; g \big) = Softmax \Big( ( z(s_i, g) )_{i\in[0,K]} \cup \{ \ell_\text{no-target} \} \Big)
\end{equation}
\end{definition}

Descriptiveness is critical for ETHER's design because it enables a \textit{single-stimulus} judgement, which allows a RG listener agent to serve as an HER predicate function. 
%Normalising over candidates alone would force the probability mass onto some candidate, so that a listener could only ever rank stimuli against one another, never assess one on its own merit. 
Indeed, an HER predicate function must answer whether a state $s$ satisfies a goal $g$ for that pair alone, independently of any candidate set of distractors, since the learning reward function of Equation~\ref{eq:learning_reward} is evaluated at every timestep of a relabelled trajectory. 
%Without the no-target extra class, the verdict on $s$ would depend on which distractors happened to be sampled alongside it, making the rewards along a relabelled trajectory a function of the distractor draw.
Conversely, because $\ell_\text{no-target}$ is a learned parameter rather than a property of a candidate set, the module trained with $K$ distractors can be read out on a single stimulus ($K=0$), where Equation~\ref{eq:no_target_softmax} reduces to a likelihood:
\begin{equation}
\label{eq:descriptive_readout}
p\big( d_0 \mid s_0 ; g \big) = \frac{\exp z(s_0,g)}{\exp z(s_0,g) + \exp \ell_\text{no-target}} = \sigma \big( z(s_0,g) - \ell_\text{no-target} \big)
\end{equation}
with $\sigma$ the logistic function. 
Section~\ref{sec:learning_predicate_function} builds the predicate function of ETHER on this readout. 
For further details on RGs and their parameterisation, see Appendices~\ref{app:emecom-background} and ~\ref{sec:ether-rg}.

\subsection{Goal-Conditioned RL}

In this section, we formalize the goal-conditioned RL problem.
%In this section, we formalize the goal-conditioned RL problem and introduce the optimization objective for simultaneously learning the policy and auxiliary functions required for effective learning.
In goal-conditioned RL, a goal-conditioned agent makes use of a policy $\pi : \mathcal{S} \times \mathcal{G} \rightarrow \mathcal{A}$ to interact at each time step $t$ with an environment to maximize its cumulative discounted reward over each episode $\sum_t \gamma^t r(s_t, a_t, s_{t+1}, g_t)$, where $\gamma \in [0, 1]$ is the discount factor, $r : \mathcal{S} \times \mathcal{A} \times \mathcal{S} \times \mathcal{G} \rightarrow \mathbb{R}$ is the environment-defined goal-conditioned reward function over the state space $\mathcal{S}$, action space $\mathcal{A}$, and goal space $\mathcal{G}$. 
It interacts by choosing an action $a_t \in \mathcal{A}$ based on an observed state $s_t \in \mathcal{S}$ and a predefined goal $g\in\mathcal{G}$ sampled at the beginning of the episode. 
Along with the output of the reward function, the agent observes at each step the next state $s_{t+1}$ sampled from the transition distribution $T(s_{t+1}|s_t, a_t )$.  

The objective of goal-conditioned RL is to find the optimal policy $\pi^\star$ that maximizes the expected return across all possible goals:
% \begin{align}
% \pi^\star = &\underset{\pi}{\arg\max} \, J_{RL}(\pi) \, \, \\
% %\text{s.t.}\, \, \\ 
% J_{RL}(\pi) &= \mathop{\mathbb{E}}_{g \sim \mathcal{G}} \left[ \mathop{\mathbb{E}}_{\tau \sim \pi} \left[ \sum_{t=0}^{T} \gamma^t r(s_t, a_t, s_{t+1}, g) \right] \right]
% \end{align}
\begin{align}
\pi^\star = &\underset{\pi}{\arg\max} \, J_{RL}(\pi) \, \, \\
\text{s.t.}\, \, %\\ 
J_{RL}(\pi) &= \mathop{\mathbb{E}}_{\substack{g \sim \mathcal{G} \\ \tau \sim \pi(\cdot|g)}} 
\Bigg[ \sum_{t=0}^{T} \gamma^t r(s_t, a_t, s_{t+1}, g) \Bigg]
\end{align}
with $\tau = (s_0, a_0, s_1, a_1, \ldots, s_T) \sim \pi$ represents a trajectory generated by following policy $\pi$ conditioned on goal $g$. 
%This formulation captures the essence of learning a goal-directed behavior policy that generalizes across the goal space.
Value-based methods that utilize a goal-conditioned Q-function, i.e. Universal Value Function Approximator \citep{schaul2015universal} can be used to optimize this objective.
The optimal Q-function, denoted as $Q^\star(s, a, g)$, satisfies the Bellman optimality equation $Q^\star(s, a, g) = \mathbb{E}_{s' \sim T(s'|s,a)} \left[ r(s, a, s', g) + \gamma \max_{a'} Q^\star(s', a', g) \right]$.
%The policy can then be derived from the Q-function as $\pi^\star(s, g) = \underset{a \in \mathcal{A}}{\arg\max} \, Q^\star(s, a, g)$.
The policy can then be derived from the Q-function as $\pi^\star(s, g) = \arg\max_{a \in \mathcal{A}} \, Q^\star(s, a, g)$.
In practice, we approximate the Q-function using neural networks and optimize it through temporal difference learning. 
% This involves minimizing the following loss function:
% \begin{equation}
% \mathcal{L}(\theta) = \mathbb{E}_{(s, a, r, s', g) \sim \mathcal{D}} \left[ \left( r + \gamma \max_{a'} Q_{\theta^-}(s', a', g) - Q_{\theta}(s, a, g) \right)^2 \right]
% \end{equation}
% where $\mathcal{D}$ represents the replay buffer containing transitions from past experiences, $\theta$ are the parameters of the Q-network being optimized, and $\theta^-$ are the parameters of a target network that is periodically updated to stabilize learning.
Optimising $J_{RL}$ involves significant practical challenges, starting with the reward function $r(s, a, s', g)$ often producing sparse signals, providing non-zero rewards only when the goal is fulfilled. 
This sparsity makes learning sample-inefficient, especially in complex environments where the probability of randomly achieving a goal is extremely low. 
To address this issue, candidates like HER introduce components to enhance the training process by more effectively utilizing the agent's experience.

%While previous works makes use of Deep Q-learning (DQN)\citep{DBLP:journals/corr/MnihKSGAWR13} to evaluate the Q-function with neural networks and perform off-policy updates by sampling transitions $(s_t, a_t, r_t, s_{t+1}, g)$ from a replay buffer, we employ Recurrent Replay with Distributed DQN (R2D2) \citep{kapturowski2018recurrent}.

\subsection{HER and its Limitations}
\label{subsec:her-limitations}
%In the context of goal-conditioned RL, rewards are inherently sparse for any given goal and this is exacerbated the larger the goal space $\mathcal{G}$ is. In order to alleviate these issues, \citet{andrychowicz2017hindsight} proposed Hindsight Experience Replay (HER) which involves relabelling unsuccessful (null-reward) trajectories where the agent failed to reach the sampled goal $g \in \mathcal{G}$, with a new goal $g^\prime \in \mathcal{G}$ that is actually found to be fulfilled in the final state of the relabelled episode. This approach improves the sample efficiency of off-policy RL algorithms %reduces the sparse reward problem by taking advantage of failed trajectories, repurposing them by reassigning them to the goals that were actually achieved. %In doing so, it improves the sample efficiency of off-policy RL algorithms, such as DQN and R2D2. In effect, for each unsuccessful trajectory, the agent's memory/replay buffer is updated with one negative trajectory and an additional positive (relabelled) trajectory. In order to do so, HER assumes the existence of a goal-relabelling function $m:S\rightarrow G$, which is an oracle (i.e. externally providing expert knowledge of the environment to the algorithm). It maps a state $s$ onto a goal $g$ that is achieved in this state. As the original experiments of \citet{andrychowicz2017hindsight} deal with spatial goals, vanilla HER can extract an alternative goal from the achieved state (because $\mathcal{G}=\mathcal{S}$), but in the more general case
Section \ref{sec:introduction} covered a high-level overview of HER. 
We look formally at it here.
At the start of an episode, a goal is sampled $g \sim \mathcal{G}$. 
At each time step $t$, a transition $(s_t, a_t, r_t, s_{t+1}, g)$ is stored in the agent's memory/replay buffer with reward $r_t$ coming from what we refer to as the \textbf{behavioural reward function}, i.e. the reward function instantiated by the environment. 
Goal-relabelling involves using another reward function: at the end of an unsuccessful episode of length $T$, relabelling and reward prediction occurs in order to store a seemingly-successful (relabelled) trajectory: an \textit{alternative} goal $g^\prime$ and corresponding reward sequence $(r^\prime_t)_{t\in[0,T]}$ are inferred using a constructed reward function, that we refer to as the \textbf{learning reward function} (detailed below). 
This is in the absence of access to the environment-privy behavioural reward function.
These relabelled transitions are added to the replay buffer $(s_t, a_t, r^\prime_t, s_{t+1}, g^\prime)_{t\in[0,T]}$ for each time step $t$. 
Thus, HER improves upon reward sparsity inherent in goal-conditioned RL by relabelling unsuccessful trajectories where the agent failed to reach goal $g \in \mathcal{G}$ with a new goal $g^\prime \in \mathcal{G}$ that has been achieved in the final state of the relabelled trajectory. 

Yet, HER implicitly assumes the existence of two functions, which we explicitly define below.

\begin{definition}[Goal-relabelling and predicate functions]
\label{def:m_f}
A \textbf{goal-relabelling function} $m: \mathcal{S} \rightarrow \mathcal{G}$, which maps a state $s$ onto a goal $g$ that is achieved in this state. 
A \textbf{predicate function} $f: \mathcal{S} \times \mathcal{G} \rightarrow \{0, 1\}$, which encodes whether the agent in a state $s$ satisfies a given goal $g$.
\end{definition}
Together, they induce the \textbf{learning reward function} that is used to infer the reward at each timestep of the relabelled trajectories.
\begin{definition}[Learning reward function]
\label{def:learning_reward}
Given a goal-relabelling function $m$ and a predicate function $f$, a learning reward function $r_\text{learn}: \mathcal{S} \times \mathcal{A} \times \mathcal{S} \times \mathcal{G} \rightarrow \mathbb{R}$ can be defined as:
\begin{equation}
\label{eq:learning_reward}
r_\text{learn}(s_t, a_t, s_{t+1}, g) = f(s_{t+1}, g)
\end{equation}
\end{definition}

To lighten notations, we extend the definition of $f$ to also take as input a goal $g\in\mathcal{G}$ and a whole RL trajectory $\tau = (s_0, \ldots, s_{|\tau|})$ such that it evaluates to the value of the last state of the trajectory, i.e. $f(\tau, g) = f(s_{|\tau|}, g)$.

HER implicitly assumes that both functions are provided as oracles, accessible black-box functions that inject expert knowledge to the algorithm. 
Originally, HER was presented in a special case where the goal and state spaces were the same, $\mathcal{G}=\mathcal{S}$~\cite{andrychowicz2017hindsight}.
However, HER is not readily applicable to the more general case of $\mathcal{G}\neq\mathcal{S}$ unless those external oracles are provided, reducing HER's applicability.

HER offers two strategies to infer an alternative goal to an unsuccessful trajectory. 
Firstly, the \textbf{final} strategy infers an alternative goal using the goal-relabelling function on the \textbf{final} state of the unsuccessful trajectory of length $T$, $\hat{g}^0 = m(s_T)$, and the corresponding rewards are computed via the learning reward function using the predicate oracle $f$,  $\forall t \in [0,T-1], r^0_t = r_\text{learn}(s_t, a_t, s_{t+1}, \hat{g}^0) = f(s_{t+1}, \hat{g}^0)$. 
%DQN update rule remains identical to [29], transitions are sampled from the replay buffer, and the network is updated using one-step td-error minimization. \\
Or, any of the \textbf{future-k} strategies can be used, with $k\in\mathbb{N}$ being an hyperparameter. 
They consist of applying the \textbf{final} strategy to $k$ different, contiguous sub-parts of the main trajectory. 

Recognising the limitations of HER pertaining to the relabelling function, HIGhER~\citep{cideron2020higher} proposes to supervisedly learn it as an instruction/goal generator from early/random successful trajectories.
While it guarantees agreement between the \textbf{learning} and \text{behavioural} reward functions, it relies on the prohibitive assumption that random RL policies are able to reach some success by acting randomly on a (sparse-reward) task. 
We provide in Appendix~\ref{app:higher} further discussion of HIGhER~\citep{cideron2020higher}'s limitations and possible extensions. 
%\todo[]{explain HIGHER and ACTRCE, and provide in appendix discussion of ACTRCE ? }

\section{Hindsight Reinforcement Learning Problem Formulation}
\label{sec:hrl-problem-formulation}

We define a unified hindsight relabelling strategy and integrate it with the search for goal-relabelling and predicate functions within the goal-conditioned RL optimisation problem. This derives a novel problem formulation, which we refer to as the \textbf{Hindsight RL (HRL) optimisation problem}. \\

\textbf{Unified Hindsight Relabelling Strategy.} We can observe that the \textbf{final} strategy is a special case of the \textbf{future-k} strategy where $k=1$ and the sub-trajectory considered is the entire trajectory itself. 
Moreover, we note that the contiguous-ness of the subparts is not necessary, and removing it would make way to more sophisticated sub-trajectories sampling. A general \textbf{unified strategy} is defined as:

For a given trajectory $\tau = (s_0, \ldots, s_T)$ of length $T$, let $\mathcal{T} = \{t_1, t_2, \ldots, t_k\}$ be a set of $k$ timestep indices sampled from $\mathcal{P}(\tau)$ a given distribution over $\{1, 2, \ldots, |\tau|\}$, e.g. uniform distribution. 
For each index $t_i \in \mathcal{T}$, we generate an alternative goal $\hat{g}^i = m(s_{t_i})$ and compute the corresponding rewards $\forall t \in [0,t_i-1], r^i_t = r_\text{learn}(s_t, a_t, s_{t+1}, \hat{g}^i) = f(s_{t+1}, \hat{g}^i)$.
In this unified framework, the \textbf{final} strategy corresponds to $\mathcal{T} = \{T\}$, while the \textbf{future-k} strategy uses $\mathcal{T} = \{t_1, t_2, \ldots, t_k\}$ where each $t_i$ is the endpoint of a sub-trajectory. 
%This formulation not only provides theoretical clarity but also allows for more flexible implementations where the indices in $\mathcal{T}$ need not correspond to contiguous sub-trajectories, potentially enabling more sophisticated relabelling schemes.

%\textbf{Naïve HRL Problem Formulation.}
%The unified hindsight relabelling strategy can be integrated into the goal-conditioned RL optimization framework naïvely as the following joint optimization problem:
\textbf{HRL Problem Formulation.}
The unified hindsight relabelling strategy can be integrated into the goal-conditioned RL optimization framework as the following joint optimization problem:
%\vspace{-10pt}
\begin{align}
\pi^*, m^*, f^* &= \underset{\pi, m, f}{\arg\max} \: J_{HRL}(\pi, m, f) \\
\text{s.t. }\, \, 
J_{HRL}(\pi, m, f) &= J_{RL}(\pi) + J_{HER}(\pi, m, f) %\, ,
%\vspace{-10pt}
\end{align}
where the HER component is framed using the unified hindsight relabelling strategy as:
% \begin{align}
% J_{HER}(\pi, m, f) &= \mathbb{E}_{g\sim\mathcal{G}} \left[ \mathbb{E}_{\substack{\tau \sim \pi \\ f(\tau, g) = 0}} 
% \left[ \mathbb{E}_{\mathcal{T} \sim \mathcal{P}(\tau)} \left[ \sum_{t_i \in \mathcal{T}} \sum_{t=0}^{t_i} \gamma^t f(s_{t+1}, m(s_{t_i})) \right] \right] \right]
% \end{align}
\begin{align}
\vspace{-10pt}
J_{HER}(\pi, m, f) = &\mathop{\mathbb{E}}_{g\sim\mathcal{G}} \Bigg[ 
\mathop{\mathbb{E}}_{\substack{\tau \sim \pi(\cdot | g) \\ f(\tau, g) = 0}} \Bigg[ 
\mathop{\mathbb{E}}_{\mathcal{T} \sim \mathcal{P}(\tau)} \Bigg[  %\nonumber \\ 
\sum_{t_i \in \mathcal{T}} \sum_{t=0}^{t_i} \gamma^t f(s_{t+1}, m(s_{t_i})) \Bigg] \Bigg] \Bigg]
\end{align}

Implying maximising the sum of discounted \textbf{learned rewards} over the $k$ $\mathcal{T}$-defined \textbf{goal-relabelled} sub-trajectories of \textbf{unsuccessful trajectories} $\tau$ generated by policy $\pi$.

While this formulation does create a joint optimization of policy $\pi$, goal-relabelling function $m$, and predicate function $f$, it is susceptible to trivial solutions and degenerative cases, which we gather below in a set of \textbf{degenerate} pairs $(m,f)$.

\begin{definition}[Degenerate relabelling-predicate pairs]
\label{def:d_deg}
Let $X \subseteq \mathcal{S}$ be a finite set of states with $|X| \geq 2$, over which a pair $(m,f)$ is evaluated, and let $c \in \mathbb{N}, c \geq 1$, be a collapse tolerance. 
The set of degenerate pairs is $\mathcal{D}_\text{deg}(X,c) = \mathcal{D}_\text{collapse}(X,c) \cup \mathcal{D}_\text{trivial}(X)$, with:
\begin{align}
\label{eq:d_collapse}
\mathcal{D}_\text{collapse}(X,c) &= \Big\{ (m,f) : \big| \{ m(s) : s \in X \} \big| \leq c \Big\} \\
\label{eq:d_trivial}
\mathcal{D}_\text{trivial}(X) &= \Big\{ (m,f) : \forall (s,g) \in X \times \mathcal{G}, f(s,g) = 1 \Big\}
\end{align}
\end{definition}

In Equation~\ref{eq:d_collapse}, $\{ m(s) : s \in X \}$ denotes the set of distinct goals that $m$ emits over $X$, so that $\mathcal{D}_\text{collapse}$ gathers the relabelling functions emitting at most $c$ distinct goals. 
Equivalently, this clause can be expressed with a continuous tolerance $\delta \in [0,1)$ instead of the discrete tolerance $c$, as a measure of the expressivity of $m$, by overloading the definition of $\mathcal{D}_\text{collapse}$ by the type of its second argument: 
\begin{equation}
\label{eq:d_collapse_continuous}
\mathcal{D}_\text{collapse}(X,\delta) = \Big\{ (m,f) : \mathop{\mathbb{E}}_{s_1, s_2 \sim \mathcal{U}(X)} \big[ \mathbb{I}(m(s_1) \neq m(s_2)) \big] \leq \delta \Big\}
\end{equation}
where $\mathcal{U}(X)$ is the uniform distribution over $X$ and $\mathbb{I}$ the indicator function. 
The counting tolerance $c$ reads more directly, while $\delta$ is independent of $|X|$ and therefore comparable across tasks; our results hold in both forms.
Note that $\mathcal{D}_\text{deg}$ is parameterised by $X$ rather than by the state distribution $\rho^\pi$ induced by the policy, so that it does not have to vary as the policy is trained.

\textit{Examples.} Consider BabyAI~\citep{Chevalier-Boisvert2018}'s PickupDist-v0 task, whose instruction space contains all the `pick up the $\langle$colour$\rangle$ $\langle$object$\rangle$' combinations. 
A pair in $\mathcal{D}_\text{collapse}$ is $m(s) = $ `pick up the grey box' for every state $s$, together with the environment's own predicate as $f$: every unsuccessful trajectory is relabelled with the same instruction, so that $J_{HER}$ is maximised over the states in which that single object is held, while the policy is only ever taught one instruction and generalises to none. 
Here $|\{m(s) : s \in X\}| = 1$ and the diversity of Equation~\ref{eq:d_collapse_continuous} is $0$. 
A pair in $\mathcal{D}_\text{trivial}$ is $f(s,g) = 1$ everywhere, together with an oracle $m$: the learning reward of Equation~\ref{eq:learning_reward} is then constant, the relabelled return is the constant $\sum_t \gamma^t$ regardless of behaviour, maximizing $J_{HER}$ but leaving all gradient with zero information. 
The tolerances matter because partial collapse is just as harmful: a relabelling function emitting only three of the task's instructions escapes a strict $|\{m(s) : s\in X\}| = 1$ criterion while remaining useless. 
Collapse of this kind is extensively studied in the context of Generative Adversarial Networks~\citep{goodfellow2014generative,salimans2016improved-techniques-for-training-gans,kossale2022mode} and self-supervised learning (see dimensional collapse - \cite{jing2022understanding,he2024preventing}). 

Interestingly, we could constraint the predicate function to yield a \textit{learning} reward function that agrees with the \textit{behavioural} one, but we assume that this is not necessary in order to create some feedback for the RL policy to start learning features, as the main objective of the HRL problem.
We acknowledge though that it may impact the convergence of the RL policy towards the optimal of $J_\text{RL}$, in ways that are yet to be uncovered.

We leave it to future works to thoroughly investigate these matters, but we leave them out of scope here because we do not plan to optimise this objective directly.

\section{Method}
\label{sec:method}

\begin{figure*}[ht]
    \centering
    \includegraphics[width=0.9\linewidth]{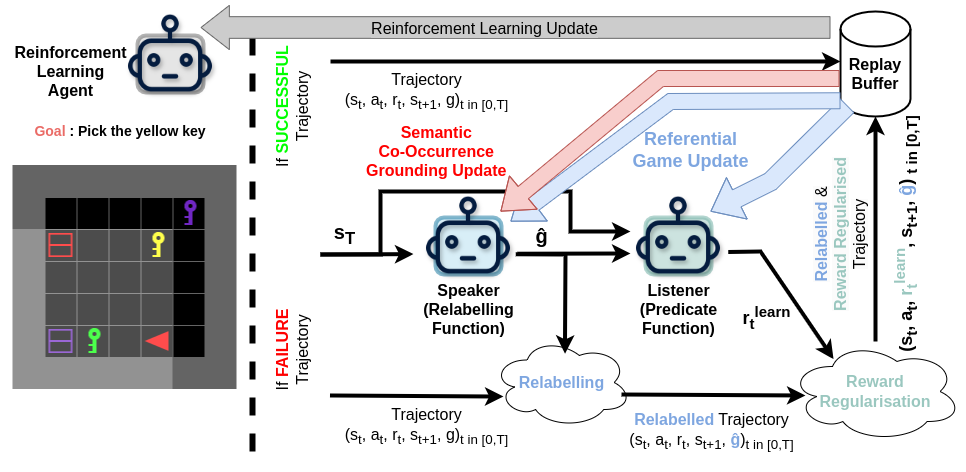}
 %   \caption{ETHER features three agents, two from a Referential Game, a \textit{speaker} and a \textit{listener}, and the Reinforcement Learning agent. On trajectories where the RL agent successfully follows instructions $g$, the trajectory is added to the replay buffer and the final state $s_T$ and episode's instruction $g$ are added to the state-instruction dataset. Sampling from this dataset allows training of the \textit{speaker} agent in a supervised learning manner, mimicking how HIGhER trains its \textbf{instruction generator}. On trajectories where the RL agent fails to follow $g$ , the \textit{speaker} agent is used as an \textbf{instruction generator} to relabel the trajectory, replacing the failed goal $g$ with a linguistic description $g^0$ of the episode's final state $s_T$. Because $s_T$ may be repeated throughout the episode, it is important to regularise the rewards throughout the trajectory. This is performed by using the \textit{listener} agent as a \textbf{predicate function}. Finally, the relabelled and reward regularised trajectory can be added to the replay buffer. Samples from the replay buffer can be used  RL updates on the RL agent as well as RG updates on both the \textit{speaker} and \textit{listener} agents. \label{fig:ether-algo} }
 %to updates all three agents.
%    }
    \caption{
    ETHER uses three agents, an RL agent and a \textit{speaker} and a \textit{listener} from a RG. 
    %Trajectories where the RL agent successfully follows instructions $g$ are added to the replay buffer and the final state $s_T$ and episode's instruction $g$ are added to the state-instruction dataset. 
    %Sampling from this dataset allows training of the \textit{speaker} agent in a supervised learning manner, mimicking how HIGhER trains its \textbf{instruction generator}. 
    Failed trajectories, where the RL agent did not follow instructions $g$, are relabelled by the RG \textit{speaker} with an alternative goal $\hat{g}=m_{RG}(s_T)$.
    %, which is in effect a description of final state $s_T$ in the RG's emergent language. 
    The RG \textit{listener}, within the \textit{predicate function} $f_{RG}$, adjusts the rewards throughout the trajectory to reflect the extent to which each state match the alternative goal $\hat{g}$. 
    The relabelled trajectory is then added to the replay buffer, from which samples can be used to update all agents with losses $J_{HRL}$ and $J_{RG}$.
    }
    \label{fig:ether-arch}
    \vspace{-15pt}
\end{figure*}

%Section~\ref{sec:extending-higher}  details our extension of HIGhER and introduces ETHER, a novel architecture designed to overcome HIGhER's limitations.

%\subsection{Leveraging Unsuccessful Trajectories with Emergent Communication}
\label{sec:rg-as-unsupervised-rl-task}

% \begin{figure}[t]
%     \centering 
%     \begin{subfigure}{0.35\textwidth}
%         \centering
%         \includegraphics[width=\textwidth]{figures/babyai_green_key.png}
%         \caption{Expert agent}
%         \label{fig:my_label}
%     \end{subfigure}
%     \begin{subfigure}{0.35\textwidth}
%         \centering
%         \includegraphics[width=\textwidth]{figures/random_green_key.png}
%         \caption{Random agent}
%         \label{fig:my_label}
%     \end{subfigure}
%     \caption{Episode goal semantics (left columns) and count of observation semantics from trajectories conditioned on a given goal (right histogram). \textbf{Left:} trajectories from BabyAI's built-in expert agent which always reaches the goal. \textbf{Right:} random agent trajectories. In both cases the semantics of the goal are among the most observed semantic features for any given trajectory. This effect is less pronounced in the random agent.}
    
%     \label{fig:co_occurrence}
% \end{figure}
We address the HRL problem using an alternating optimization framework, alternating between learning an RL policy that maximises $J_{HRL}$ and learning relabelling and predicate functions that makes $J_{HRL}$ denser in feedback signals.
We propose ETHER to do the latter by following the Emergent Communication paradigm and learn a relabelling function as a speaker agent and a predicate function as a listener agent, as detailed in Section~\ref{sec:learning_predicate_function}. 
The algorithm is shown in Algorithm~\ref{alg:ETHER}, with shared components between RG and RL agents highlighted in Figure~\ref{fig:ether-arch}, framing the RG as an unsupervised auxiliary task for RL~\citep{jaderberg2016UNREAL}. 
We acknowledged in Section~\ref{sec:hrl-problem-formulation} that learning a predicate function that yields a \textbf{learning} reward function that agrees with the \textbf{behavioural} one may impact the convergence of the RL policy.
Therefore, in order to enforce this constraint, ETHER must align the RG's emergent language with the environment's natural language, a challenge that we address in Section~\ref{subsec:co-occurrence} with our alignment method relying on semantic co-occurrences between observations and goals.
%Section\ref{sec:learning_predicate_function} details the listener agent as a predicate function, 

\begin{comment}
In the following, we rebase the architecture around the Emergent Communication paradigm in order to both learn a relabelling function, in the form of the speaker agent, and a predicate function, in the form of a listener agent.
The resulting algorithm is illustrated in Figure~\ref{fig:ether-algo}.
Figure~\ref{fig:ether-arch} also highlights shared components between the RG agents and the RL agent in ETHER, thus allowing the RG to be acknowledged as an unsupervised auxiliary task for RL, following the work of ~\citet{jaderberg2016UNREAL}.
This change of paradigm brings about a new challenge in the alignment of the emergent language used by the RG agents to the natural-like language of the benchmark/environment.
In the following Section~\ref{sec:learning_predicate_function}, we highlight how to use an RG's listener agent as a predicate function, and then we detail in Section~\ref{subsec:co-occurrence} our proposal to align the emergent language with the environment's language.
\end{comment}

\subsection{Learning Relabelling \& Predicate Functions via Referential Games}
\label{sec:learning_predicate_function}

Assimilating the set $\Sigma^\star$ of all emergent language utterances over vocabulary $\Sigma$ as an instruction/goal space, the RG speaker agent actually maps from the state of observations/states $\mathcal{S}$ to an instruction/goal space, as needed of a relabelling function (or Instruction Generator, as in HIGhER~\citep{cideron2020higher}).
Thus, with ETHER, we propose to learn both functions of Definition~\ref{def:m_f} in unsupervised fashion from the agents of a descriptive RG, assimilating the set $\Sigma^\star$ of all emergent language utterances as an instruction/goal space. 
The relabelling function is the speaker agent, and the predicate function is obtained by reading out the listener agent on the single state being evaluated, following Equation~\ref{eq:descriptive_readout}. 

\begin{definition}[RG-induced relabelling and predicate functions]
\label{def:m_rg_f_rg}
Given a descriptive RG (Definition~\ref{def:descriptive_rg}) with speaker $Sp$ and listener $Li$, the \textbf{RG-induced pair} $(m_{RG}, f_{RG})$ is defined, for all state $s\in\mathcal{S}$ and goal $g\in\mathcal{G}\subset\Sigma^\star$, as:
\begin{align}
\label{eq:m_rg}
m_{RG}(s) &= Sp(s) \\
\label{eq:f_rg}
f_{RG}(s,g) &= \mathbb{I}\big( \sigma( z(s,g) - \ell_\text{no-target} ) \geq \beta \big)
\end{align}
with the indicator function $\mathbb{I}(\cdot)$ and a threshold $\beta$.
\end{definition}

Note that $\beta$ is not a free hyperparameter: since $\sigma$ is increasing and $\sigma(0)=\nicefrac{1}{2}$, the canonical choice $\beta=\nicefrac{1}{2}$ makes Equation~\ref{eq:f_rg} equivalent to $\mathbb{I}( z(s,g) \geq \ell_\text{no-target} )$, i.e. the predicate holds exactly when the listener agent finds goal $g$ a better account of state $s$ than it finds the \textit{no-target} alternative. 
The threshold is therefore the decision boundary that the RG itself has learned, rather than one we impose. 
Thus, the listener agent of a descriptive RG can readily act as a predicate function (as illustrated in Figure~\ref{fig:ether-arch}), provided that the RG's speaker agent is used as a relabelling function.
%This extension already incorporates contrastive learning, as a discriminative RG is literally asking the listener agent to contrast positive (the target) and negative (the distractors) stimuli.
%We refer to this augmentation as the Emergent Textual Hindsight Experience Replay (ETHER) agent.

Crucially, this construction is not susceptible to the degenerate cases of Definition~\ref{def:d_deg}: solutions of the RG objective are non-degenerate by construction, as stated below.

\begin{theorem}[Non-degeneracy of RG-induced solutions]
\label{thm:non_degeneracy}
Let $(Sp^\star, Li^\star)$ minimise $J_{RG}$ (Equation~\ref{eq:j_rg}) with the Hinge loss of Equation~\ref{eq:hinge_loss} of margin $\nu>0$, over a descriptive RG (Definition~\ref{def:descriptive_rg}) with $K\geq1$ distractors drawn independently from $\mathcal{D}_{RG}$, and assume $J_{RG}(Sp^\star, Li^\star)=0$ over both target-present and target-absent rounds.
Then the induced pair $(m_{RG}, f_{RG})$ of Definition~\ref{def:m_rg_f_rg}, with $\beta=\nicefrac{1}{2}$, is non-degenerate over $\mathcal{D}_{RG}$:
\begin{equation}
\label{eq:non_degeneracy}
(m_{RG}, f_{RG}) \notin \mathcal{D}_\text{deg}(\mathcal{D}_{RG}, c) \quad \forall c < |\mathcal{D}_{RG}|
\end{equation}
and equivalently $(m_{RG}, f_{RG}) \notin \mathcal{D}_\text{deg}(\mathcal{D}_{RG}, \delta)$ for all $\delta < 1 - \nicefrac{1}{|\mathcal{D}_{RG}|}$.
\end{theorem}

In other words, the contrastive nature of the RG objective forces the speaker agent to emit a \textit{distinct} utterance for every stimulus, ruling out the collapse of $m_{RG}$, while descriptiveness forces the learned $\ell_\text{no-target}$ to separate targets from non-targets, ruling out a vacuous $f_{RG}$.
The proof is deferred to Appendix~\ref{app:non_degeneracy}.
See Appendix~\ref{sec:ether-rg} for details on RG instantiation.

\subsection{Language Alignment: Semantic Co-occurrence Grounding}
\label{subsec:co-occurrence}

Goals/instructions take the form of semantic descriptions of the necessary interactions between the agent and objects that are present in the environment (e.g., "pick up the green key", "open the door"). 
We hypothesise that, upon specifying a goal, agent observations will be biased to contain semantic components present in said goal. 
We test this hypothesis in the BabyAI environment.
Figures~\ref{fig:co_occurrence:expert} and ~\ref{fig:co_occurrence:random} (and in Appendix~\ref{sec:co-occurrence}) show, indeed, the semantics of the goal are some of the most salient observed semantics in both expert trajectories and random walks. 
Given the similarity between semantics of observations and goals in both successful and unsuccessful trajectories, which we refer to henceforth as \textbf{semantic co-occurrence}, we ask ourselves: ``How can this underlying environmental structure be leveraged to learn a semantic understanding of the goals?''
Using EmeCom principles, this translates as: ``How can this underlying environmental structure be leveraged to constraint the RG's emergent language to be aligned with the (natural) language used to describe goals in any instruction-following task?'' 

To answer this, we introduce the \textbf{semantic co-occurrence grounding loss}, 
%which aims to enhance an agent's language grounding ability during RG training by biasing the speaker agent's token embeddings towards parts of the agent's field of view containing semantic components that are emphasised by the current goal. 
%Only the words/tokens present in the linguistic goal description provided are used, there is not need for any environment-private information. 
%An example of this would be to bring the embedding of the token 'blue' closer to the visual embeddings of the part of the observation that contains a blue-coloured object.
%We refer to this architecture as ETHER+.
 which aims to enhance an agent's language grounding ability during RG training. 
This loss does not rely on private information from the environment, but solely on making use of information observable by an instruction-following agent. 
Only the words /tokens present in the instruction/goal provided are used as labels. 
Formally, we define an instruction/goal as a series of tokens, $g=(g_i)_{i\in [1,L]} \in \mathcal{G}\subset\Sigma^\star$, where $L$ is the maximum sentence length hyperparameter, as defined in the RG setup (see Appendix~\ref{sec:ether-rg}).

Thus, for each of those token present in the goal $g$ of a given episode (out of all the tokens available in the vocabulary $\Sigma$, as defined in the RG setup), the semantic co-occurrence grounding loss aims to bring a prior semantic-only embedding of the tokens closer to the visual embeddings of all the observations during a given episode.
We denote by $(\lambda_w)_{w\in \Sigma}$ all the prior semantic-only embeddings for the vocabulary $\Sigma$. 
It will also further separate from the visual embeddings of all the observations during the episode the prior semantic-only embeddings of \textbf{all the vocabulary tokens that are not present in the current goal $g$}.

The semantic co-occurrence grounding loss is contrastive and inspired by \citet{radford2021CLIP}. 
More formally, as we define $f(\cdot)$ as the visual module in the RG speaker agent architecture (cf. Section~\ref{app:model-architecture}), we write the semantic co-occurrence grounding loss as follows:
\begin{align}
\label{eq:semantic-co-occurrence-grounding-loss}
\mathcal{L}^{sem.}_{co-occ.} ( g | (\lambda_w)_{w\in \Sigma} ) &= \mathop{\mathbb{E}}_{s \sim \rho^\pi} \Biggl[ \sum_{w\in V} \mathcal{H}(w) \cdot %\nonumber \\ 
\sum_{g_i \in g} \biggl( \mathbf{1}_{w}(g_i) - \frac{\lambda_w \cdot f(s)^T}{||\lambda_w ||_2 \cdot ||f(s)||_2} \biggr)^2 \Biggr]
\end{align}

where $||\cdot||_2$ corresponds to the $L2$ norm, $\rho^\pi$ is the distribution over states in $s\in\mathcal{S}$ that is induced by using the RL policy $\pi$ to harvest the observations/stimuli, and ${1}_{w}(\cdot)$ is a noisy indicator function:
\begin{equation}
\vspace{-5pt}
\label{eq:indicator-function}
    \mathbf {1}_{w}(w^\prime)= (1-\epsilon_{noise}) \times \begin{cases}1 &{\text{ if }}~w^\prime = w~,\\-1 &{\text{ if }}~w^\prime \neq w~.\end{cases}
\end{equation}
where $\epsilon_{noise}$ is a noise uniformly sampled from $[0,0.2]$, using the noisy labels trick from \citep{salimans2016improved-techniques-for-training-gans}.

As the loss is implemented over mini-batches of sampled stimuli, we also perform masking to reject tokens with null entropy over the mini-batch. 
For instance, in the proposed experiments performed on BabyAI~\citep{Chevalier-Boisvert2018}'s PickupDist-v0 task, the instruction/goal always contains the prefix `pick up', therefore, when considering a mini-batch of stimuli (however they may come from different episodes), the likelihoods of the tokens `pick' and `up' are maximal over the mini-batch and therefore their associated appearance distribution over the sampled stimuli will have null entropy. 
In Equation~\ref{eq:semantic-co-occurrence-grounding-loss}, $\mathcal{H}(w)$ denote the entropy of the appearance distribution of token $w\in \Sigma$, and its presence as a multiplicative term is for masking purposes.
%We refer to the architecture incorporating the loss of Equation~\ref{eq:semantic-co-occurrence-grounding-loss} as ETHER.

%We acknowledge that there are other potential applications for our semantic co-occurrence auxiliary task, such as using this loss on a critic.
%\todo[inline]{Explain succintly (1 phrase) how this could be used in a critic}

%\input{parts2/metrics}

\section{Experiments}
\label{sec:experiments}

%\subsection{Experimental setup}
All experiments are run in an altered version of the BabyAI environment 'BabyAI-PickUpDist-v0' \cite{Chevalier-Boisvert2018}. 
The agent is rewarded at each episode for picking up a specifically coloured and shaped object among other distractor objects given a natural(-like) language instruction, e.g. ``Pick up the blue ball''. 
The episode ends when any object is picked up unless the episode times out after $40$ available timesteps (as in \cite{cideron2020higher}). 
An episode is successful if the agent picks up the instruction-defined object, or unsuccessful if any other object is picked or a time-out occurs. 
Implementation details, such as agent architecture, are described in Appendix~\ref{sec:implementation-details}. %Agent architectures are described in Appendix \ref{app:model-architecture}.

We train agents for a budget of $1$M observations.
We compare to a baseline R2D2~\citep{kapturowski2018recurrent} agent, and an upper-bound HER agent that has access to oracle relabelling and predicate functions, referred to as \textit{PickUp HER}.
The oracle relabelling function only highlights features relevant to the instructions obtained from the environment, i.e. picked up objects, if any.
Thus, it maps all states where no object is picked up to the `EoS' instruction.
To show the impact of a relabelling function that highlights richer features, albeit not necessarily relevant, we propose a control condition, referred to as \textit{PickUp+Descr HER}, whose relabelling oracle maps all states where no object is picked up to an instruction describing surrounding objects, e.g. `wall in front, blue box on the right'.

\begin{comment}
\begin{wraptable}{R}{0.4\textwidth}
\vspace{-30pt}
\caption{\textbf{Top:} Success ratios (percentage of mean and standard deviation) for agents with burn-in feature of R2D2 after 200k observations.
\textbf{Bottom:} Accuracies (percentage of mean and standard deviation) for semantic grounding of the speaker agent after 200k observations. 
}
\label{tab:performance}
%\centering
%    \renewcommand{\arraystretch}{1.5}
\begin{tabular}{@{}lccc@{}} 
%\begin{tabular}{c|c|c}
\toprule
    \textbf{Agent} & \textbf{Success Ratio} & \\
    R2D2 & 16.54 $\pm$ 1.37 & \\ %1.372 
    \midrule
    HIGhER+ & 14.84 $\pm$ 1.40 & \\ %1.408 
    HIGhER++ (n=1) & 15.89 $\pm$ 1.19 & \\ %1.193
    HIGhER++ (n=2) & 16.80 $\pm$ 2.07 & \\ %2.067
    HIGhER++ (n=4) & 18.10 $\pm$ 2.54 & \\ %2.542
    \midrule 
    ETHER & 27.63 $\pm$ 1.20 & \\
    ETHER+ & 27.16 $\pm$ 2.57 & \\
    \bottomrule
%\caption{Accuracies (percentage of mean and standard deviation) for semantic grounding of the speaker agent after 200k observations.}
%\label{tab:alignement-accuracy}
    \toprule
     & \textbf{Any-Colour} & \\ %\textbf{Any-Shape} \\
    ETHER & 9.117 $\pm$ 1.676 & \\ %0.0 $\pm$ 0.0 \\
    ETHER+ & 32.75 $\pm$ 6.29 & \\ %0.0 $\pm$ 0.0 \\
    \bottomrule
\end{tabular}
\end{wraptable}

\end{comment}

\begin{table*}[th]
% \vspace{-12pt}
\caption{
%Success ratios for agents with burn-in feature of R2D2 after 200k observations.
Performances (mean $\pm$ std.err.) of different agents after $1$M observations at the instruction-following task, and language alignment's accuracies in terms of the \textit{Any-Shape} and \textit{Any-Colour} metrics.
}
\label{tab:performance}
\centering
\begin{tabular}{@{}llll@{}} 
%\begin{tabular}{c|c|c}
\toprule
    \textbf{Agent} & \textbf{Success \%} & \textbf{Any-Colour} & \textbf{Any-Shape} \\
    \midrule 
%    &               &                   &   Acc.        & Prec.         & Rec.          & Acc.          & Prec.         & Rec.          \\
%    R2D2 & 16.54 $\pm$ 1.37 \\ %1.372 
    R2D2            & $58.01\pm4.07$    & --            & --            \\ %1.372 
    \midrule
%    HIGhER+ & 14.84 $\pm$ 1.40 \\ %1.408 
%    HIGhER++ (n=1) & 15.89 $\pm$ 1.19 \\ %1.193
%    HIGhER++ (n=2) & 16.80 $\pm$ 2.07 \\ %2.067
%    HIGhER++ (n=4) & 18.10 $\pm$ 2.54 \\ %2.542
    Pickup+Descr HER & $61.72\pm3.91$    & $100.00\pm0.0$& $100.0\pm0.0$ \\
    Pickup HER   &  $74.41\pm6.83$   & $6.24\pm1.48$ & $6.24\pm1.48$ \\
    \midrule 
    ETHER           & $67.68 \pm 4.56$  & $24.76\pm5.30$& $15.95\pm22.56$ \\
%    ETHER & 27.63 $\pm$ 1.20 \\
%    ETHER+ & 27.16 $\pm$ 2.57 \\
    \bottomrule
\end{tabular}
%\vspace{-10pt}
\end{table*}

\textbf{Hypotheses.} 
%We test 4 hypotheses: \textbf{(H1)} Replacing the oracle predicate in HIGhER with a derived deterministic predicate improves over the R2D2 baseline; \textbf{(H2)} ETHER’s hindsight learning and RG based unsupervised auxiliary tasks improve sample efficiency and final performance over R2D2; \textbf{(H3)} ETHER’s theoretically grounded predicate learning outperforms heuristic-based HIGhER variants; \textbf{(H4)} ETHER+’s semantic grounding constraint affects RL performance.
We test $3$ hypotheses:  \textbf{(H1)} ETHER’s hindsight learning and RG-based unsupervised auxiliary tasks improve sample efficiency and final performance over baseline; 
\textbf{(H2)} ETHER's semantic co-occurrence grounding aligns the RG's emergent language with the RL environment's natural(-like) language; 
\textbf{(H3)} Hindsight learning with instructions/goals irrelevant to the benchmark environment's instructions, like in \textit{Pickup+Descr HER}, hurts sample efficiency and final performance.
% \textbf{Hypothesis.} Firstly, our extensions to HIGhER replace the oracle predicate function with a derived, deterministic predicate function which is not theoretically sound, thus we investigate here whether this can still be beneficial to the RL agent's learning by comparison to our R2D2 baseline \textbf{(H1)}. Secondly, as ETHER instantiates hindsight learning and an unsupservised auxiliary task for RL in the form of the RG, we expect it to improve the sample-efficiency and the asymptotic performance compared to our R2D2 baseline \textbf{(H2)}. Thirdly, since ETHER learns a principled predicate function (in the form of the listener agent of a RG) which is theoretically sound, as opposed to the predicate functions derived in our various HIGhER extensions, we hypothesise that ETHER should also improve the sample-efficiency and the asymptotic performance compared to HIGhER extensions \textbf{(H3)}. Finally, as ETHER+ constraints the EL via semantic co-occurrence grounding, we ponder whether this constraint has any impact on the RL agent's performance \textbf{(H4)}. 

\textbf{Evaluation}. For \textbf{(H1)} and \textbf{(H3)}, to evaluate the sample efficiency and performance, we report the success percentage of the RL agents after a fixed budget of $1$M observations. 
For \textbf{(H2)}, to evaluate alignment of the emergent language towards the environment's natural(-like) language, we propose two metrics, the \textbf{Any-Colour} and \textbf{Any-Shape} metrics, which measure whether \textbf{any} of the visual semantics visible in a given RL observation are mentioned in the emergent language utterances spoken by the RG speaker, in context of semantics pertaining to, respectively, colour and shape.
We report the accuracy of the RG speaker at using \textbf{any} of the relevant semantics' word, computed over a test set of $1024$ stimuli.  

\textbf{Results.} Table~\ref{tab:performance}'s second column shows the success percentages for the different agents and conditions.
\begin{comment} 
Our HIGhER extensions requires a great amount of negative examples in the contrastive training (n=4) to validate \textbf{(H1)}, providing marginal improvements over the baseline. This shows that our derived predicate function is practically feasible but limited. Conversely, both ETHER approaches outperform all others, almost doubling in final performance, validating \textbf{(H2)} and \textbf{(H3)}.
These results are cementing the usage of visual discriminative referential games as viable unsupervised auxiliary task for RL and they are showing that our principled, RG-learned predicate function is both theoretically sound and practical.
Finally, regarding \textbf{(H4)}, we observe similar mean asymptotic performance between ETHER and ETHER+, but ETHER+'s distribution has a greater standard deviation which leads us to think that the semantic co-occurrence grounding may exert some detrimental constraints onto the RL agent.
We bring the reader's attention onto the fact that the noise parameter $\epsilon_\text{noise}$ of the semantic co-occurrence grounding loss (cf. equation~\ref{eq:indicator-function}) may still be too strong and thus explain this detrimental effect on the RL performance.
%This detriment might be due to the noise parameter of the semantic co-occurrance grounding loss $\epsilon_\text{noise}$ being too large (equation~\ref{eq:indicator-function}).

\subsection{Semantic Co-Occurrence Grounding Improves Emergent Language Alignment}
\label{sec:semantic-co-occurrence-grounding-improves-emergent}
\end{comment}
As the performance of ETHER is more than $15\%$ higher relative to baseline R2D2's, we validate hypothesis \textbf{(H1)}.
Nevertheless, performances of ETHER are below that of upper-bound condition \textit{Pickup HER}.
We suspect that the emergence of the emergent language in parallel of the RL policy learning may have an impact on the policy learning due to the initial transient phase where RG agents adapt the emergent language to suit their needs.
Indeed, this transient phase is not present in the context of the upper-bound control condition.
Performances are even higher than control condition \textit{Pickup+Descr HER}, which validates \textbf{(H3)} and highlights a need to further our understanding at a distributional level of the interplay between relevant instructions/goals and irrelevant ones created by learnt relabelling functions.
Finally, ETHER's language alignment measures, well above the null values of the ablated ETHER, show that the semantic co-occurrence grounding loss does encourage some alignment between the emergent language and the benchmark's natural-like language regarding the colour semantic, primarily, and the shape semantic, secondly, validating \textbf{(H2)}, with values even above that of the (performance) upper-bound control condition.
ETHER measures are well below that of the \textit{Pick+Descr HER} condition, which highlights limitations of the semantic co-occurrence grounding loss.
Coupled with performance results, it highlights a non-trivial relationship between language alignment and RL performance.
Both matters ought to be studied more thoroughly in subsequent works.

%The latter possibly explains the current performance of the agents not exceeding $33\%$ of success ratio, which corresponds to the random success ratio \textbf{if there was only the shape semantic to discriminate against}. 
%Indeed, the task proposes only 3 different shape semantics: the ball, the key, and the box.

% \begin{wraptable}[R]{0.5\textwidth}
% \centering
%     \renewcommand{\arraystretch}{1.5}
%     \begin{tabular}{c|c|c}
%     \textbf{Agent} & \textbf{Success Ratio} & \\ \hline
%     ETHER & 27.63 $\pm$ 1.20 & \\ \hline
%     ETHER+ & 27.16 $\pm$ 2.57 & \\ \hline
%     \end{tabular}
% \caption{Success ratios ( percentage mean and standard deviation) for agents with burnin feature of R2D2 after 200k steps in a modified version of the BabyAI PickUpDist task.}
% \label{tab:Higher-PickUp}
% \end{wraptable}

%\input{parts/discussion}

%\section{Discussion}
\section{Conclusion}
\label{sec:conclusion}

We highlighted HER~\citep{andrychowicz2017hindsight}'s limitations in terms of goal relabelling and predicate functions in instruction-following tasks and formalized a novel problem, the Hindsight Reinforcement Learning (HRL) problem, which emphasise the need to jointly learn relabelling and predicate functions alongside the RL policy.
We propose the Emergent Textual Hindsight Experience Replay (ETHER) agent to tackle the HRL problem. 
%and HIGhER~\cite{cideron2020higher}. 
ETHER uses a RG and a semantic grounding scheme as unsupervised auxiliary tasks~\citep{jaderberg2016UNREAL} to learn relabelling and predicate functions while aligning the RG emergent language with the the instruction-following task's language. 
%The speaker and listener agents make an artificial language emerge that can be aligned with the natural-like language used to describe goals in the BabyAI benchmark, which is expressive enough to also describe unsuccessful RL trajectories, providing feedback to the RL agent to leverage the linguistic, structured information in all trajectories. Thus, we show that  is a viable unsupervised auxiliary task for goal-conditioned RL in sparse reward settings, making HER more widely applicable.
Our experiments showed that ETHER’s learned goal relabelling and predicate functions enable increased sample efficiency despite imperfect emergent language alignment. 
Our work bridges EmeCom and goal-conditioned RL, and further investigations of HER in wider contexts.
\section{Limitations}
\label{limitations}

\textbf{On Language Alignment Metrics.}
For the computation of the \textbf{Any-Colour/Shape} metrics, we use private information from the BabyAI benchmark that corresponds to the symbolic representation of the agent's field of view (as opposed to the pixel observation that the agents have as input state from the environment), here after referred to as symbolic image. 
The symbolic image describes the colour and shapes of the objects that are in the field of view of the agent using indices\footnote{cf. \url{https://minigrid.farama.org/api/wrappers/\#symbolic-obs}}. 
For each observation, we convert the indices of the corresponding symbolic image into colour and shape word tokens and check whether the RG's speaker agent use any of those word token in its emergent language description of the current observation.
%Thus, the 'Any-Colour' accuracy metric registers high accuracy for the current observation iff \textbf{any} of the visible object's colour-related word tokens is used, and similarly with shape-related tokens for 'Any-Shape'.
These metrics only provide a lower bound to language alignment and do not allow for verification of whether each word token of a given colour or shape are related to their expected semantic with a one-to-one relationship.
Yet, these metrics still allow for verification of whether the colour and shape information are being consistently used by the RG agents in ways that are necessary towards full language alignment. 
%allows for interpretation of the emergent language utterances.
We highlight that we did also experiment with the conjunctions counterparts' metrics, the \textbf{All-Colour/Shape} metrics, as well as the \textbf{Any/All-Object} metrics (checking whether any/all objects are mentioned, in terms of both their colour \textbf{and} their shape ; in order to disambiguate from the case where the emergent language description would make use of the colour-related token for a first visible object and the shape-related token to refer to another visible object rather than the same as the colour-related token's one), but ETHER is not able to perform better than $0\%$ on these metrics. 
We plan to investigate in further details in future works how to make progress on these more difficult alignment metrics, as well as include metrics related to the structure of the language, such as compositionality as it has the potential to improve learning speed~\citep{Jiang2019-language-abstraction-hierarchical-rl,li2019ease}.

\textbf{On Augmentations of ETHER.}
We proposed in this paper the minimally viable architecture to show initial results with ETHER, and left out many features, such as possible sharing of components between RL agent and RG agents (e.g. observation encoder), or the application of a supervised learning loss similar to HIGhER~\citep{cideron2020higher}'s to further constrain the learned relabelling function. 
We leave these matters to be thoroughly investigated in subsequent works.

\textbf{On External Validation of ETHER.}
The BabyAI's PickupDist task is not the most challenging instruction-following task ETHER could be deployed on, and we study it here as a toy application with low computational complexity, in order to check internal validity and to motivate subsequent works at the interface of EmeCom and goal-conditioned RL. 
We leave it to subsequent works to perform external validity check of the design choices of ETHER and its components on more challenging tasks. 

\textbf{On the Constraining of the HRL Problem.}
As highlighted Section~\ref{sec:hrl-problem-formulation}, our formulation of the HRL problem is minimal, focusing solely on the need to learn jointly the RL policy and the relabelling and predicate functions, while leaving out any constraints that could be applied in order to prevent from trivial solutions and the collapsing problem~\citep{salimans2016improved-techniques-for-training-gans,kossale2022mode,jing2022understanding,he2024preventing}, for instance.
We did provide some experimental evidence of the impact of the lack of constraining of the predicate function to yield a \textit{learning} reward function that agrees with the \textit{behavioural} one, following \textbf{(H3)} in Section~\ref{sec:experiments}.
Nevertheless, we re-iterate that our formulation's main goal is to create some feedback for the RL policy to start learning features, putting an emphasis on the initial phase of the RL policy learning, as opposed to the final/convergence phase.
We leave it to subsequent work to investigate constraints on the HRL problem, and more specifically of the kind that enhance the convergence properties.

\begin{comment}
\section*{Broader Impact}

This work consists solely of simulations, thus alleviating some of the ethical concerns, as well as concerns regarding any consequences emerging due to the failure of the system presented. With regards to the ethical aspects related to its inclusion in the field of Artificial Intelligence, we argue that our work aims to have positive outcomes on the development of human-machine interfaces, albeit being not yet mature enough to aim for this goal. The current state of our work does not allow us to extrapolate towards negative outcomes.

This work should benefit the research community of language emergence and grounding, in its current state.
\end{comment}

%\begin{comment}
\section*{Acknowledgements}
%\section*{Acknowledgements}
This work was supported by the EPSRC Centre for Doctoral Training in Intelligent Games \& Games Intelligence (IGGI) [EP/L015846/1]. 

We gratefully acknowledge the use of Python\cite{python-2009}, IPython\cite{ipython-perez-2007}, SciPy\cite{SciPy-NMeth2020}, Scikit-learn\cite{Scikit-learn:JMLR:v12:pedregosa11a}, Scikit-image\cite{scikit-image-van2014}, NumPy\cite{NumPy-Array2020}, Pandas\cite{pandas1-mckinney-proc-scipy-2010,pandas2-reback2020}, OpenCV\cite{opencv_library}, PyTorch\cite{pytorch-paszke-NEURIPS2019_9015}, TensorboardX\cite{huang2018tensorboardx}, Tensorboard from the Tensorflow ecosystem\cite{tensorflow2015-whitepaper}, without which this work would not be possible.
%\end{comment}

% Bibliography:
% biblatex:
%\printbibliography
% natbib:
%\bibliography{bibli}
\bibliography{main}

\newpage
\appendix
\onecolumn

\section{Broader impact}

No technology is safe from being used for malicious purposes, which equally applies to our research. However, we view many of the ethical concerns surrounding research to be mitigated in the present case. These include data-related concerns such as fair use or issues surrounding use of human subjects, given that our data consists solely of simulations.

With regards to the ethical aspects related to its inclusion in the field of Artificial Intelligence, we argue that our work aims to have positive outcomes on the development of human-machine interfaces since we investigate, among other things, alignment of emergent languages with natural-like languages.

The current state of our work does not allow extrapolation towards negative outcomes.
We believe that this work is of benefit to the research community of reinforcement learning, language emergence and grounding, in their current state.

%However, aiming to develop artificial agents that relies on the same symbolic behaviours and the same social assumptions (e.g. using CLBs) than human beings is aiming to reduce misunderstanding between human and machines.
%Thus, the current work is targeting benevolent applications.
%Subsequent works around the benchmark that we propose are prompted to focus on emerging protocols in general (not just posdis-compositional languages), while still aiming to provide a better understanding of artificial agent's symbolic behaviour biases and differences, especially when compared to human beings, thus aiming to guard against possible misunderstandings and misaligned behaviours. 

\section{Implementation Details}
\label{sec:implementation-details}

\begin{table}[t]
\centering
\vspace{-50pt}
\caption{Hyper-parameter values relevant to R2D2 in the different architectures presented. Missing parameters follow Ape-X~\citep{horgan2018apex}.}
\label{table:hyperparams}
\begin{tabular}{@{} ll @{}} 
\toprule
\multicolumn{2}{c}{R2D2} \\ 
\midrule
\multicolumn{1}{l}{Number of actors} & \multicolumn{1}{c}{32} \\
\multicolumn{1}{l}{Actor update interval} & \multicolumn{1}{c}{1 env. step} \\
\multicolumn{1}{l}{Sequence unroll length} & \multicolumn{1}{c}{20} \\
\multicolumn{1}{l}{Sequence length overlap} & \multicolumn{1}{c}{10} \\
\multicolumn{1}{l}{Sequence burn-in length} & \multicolumn{1}{c}{10} \\
\multicolumn{1}{l}{N-steps return} & \multicolumn{1}{c}{3} \\
\multicolumn{1}{l}{Replay buffer size} & \multicolumn{1}{c}{$1\times10^4$ obs.} \\
\multicolumn{1}{l}{Priority exponent} & \multicolumn{1}{c}{0.9} \\
\multirow{2}{0.45\linewidth}{Importance sampling exponent} & \multicolumn{1}{c}{0.6} \\
& \\
\multicolumn{1}{l}{Discount $\gamma$} & \multicolumn{1}{c}{$0.98$} \\
\multicolumn{1}{l}{Minibatch size} & \multicolumn{1}{c}{64} \\
\multicolumn{1}{l}{Optimizer} & \multicolumn{1}{c}{Adam~\citep{kingma2014adam}} \\
\multicolumn{1}{l}{Learning rate} & \multicolumn{1}{c}{$6.25\times10^{-5}$} \\
\multicolumn{1}{l}{Adam $\epsilon$} & \multicolumn{1}{c}{$10^{-12}$} \\
\multirow{2}{0.35\linewidth}{Target network update interval} & \multicolumn{1}{c}{2500} \\
&  \multicolumn{1}{c}{updates} \\
%\multicolumn{2}{l}{Target network update interval} & \multicolumn{2}{c}{2500 updates} \\
\multicolumn{1}{l}{Value function rescaling} & \multicolumn{1}{c}{None} \\
%h(x) = sign(x)(√|x|+ 1−1) +x, = 10−3
\bottomrule
\end{tabular}
\end{table}

While previous works makes use of Deep Q-learning (DQN)\citep{DBLP:journals/corr/MnihKSGAWR13} to evaluate the Q-function with neural networks and perform off-policy updates by sampling transitions $(s_t, a_t, r_t, s_{t+1}, g)$ from a replay buffer, we employ Recurrent Replay with Distributed DQN (R2D2) \citep{kapturowski2018recurrent}.
Table~\ref{table:hyperparams} highlights the hyperparameters used for our implementation of R2D2.
More details can be found, for reproducibility purposes, in our open-source implementation at %HIDDEN\_FOR\_REVIEW\_PURPOSES. 
\url{https://github.com/Near32/Regym/tree/develop-ETHER/benchmark/ETHER}. 

%Training was performed for each run on 1 NVIDIA GTX1080 Ti, and the average amount of training time for a run is 18 hours, on an observation budget of 5 million samples, with 32 actors.
Training was performed for each run on 1 NVIDIA GTX1080 Ti. Experiment's training times ranged from 8 to 24 hours depending on the architecture.
%, on an observation budget of 5 million samples, with 32 actors.
Hyperparameters for R2D2 and ETHER were found using W\&B~\citep{wandb}'s Hyperparameter Sweep feature.
\subsection{Agent Architecture}
\label{app:model-architecture}
\begin{comment}
\begin{figure}[t]
    \centering
    \begin{subfigure}[t]{1\linewidth}
        \includegraphics[width=1.0\linewidth]{figures/ETHER-arch.png}
        \caption{}
        \label{fig:ether-arch}
    \end{subfigure}
    \caption{ETHER agent architecture describing the \textit{Shared Observation Encoder} that feeds its output \textbf{convolutional feature maps} to both RG agents and the RL agent. Note that, prior to being fed to any agent, some form of agent-specific adaptations are applied to the feature maps, with the RL agent having the most sophisticated ones relying on some \textit{FiLM-ed layers} \citep{perez2018film} that are conditioned on the output of a \textit{GRU layer} that embeds the \textit{linguistic goal description}.
    }
\end{figure}
\end{comment}

The ETHER architecture is made up of three differentiable agents, the language-conditioned RL agent and the two RG agents (speaker and listener).
%Similarly, the HIGhER architecture is built around a language-conditioned RL agent and an Instruction Generator, which plays the same role as the speaker agent in a RG thus we equate them in the following architectural details. 
Each agent consists of at least a language module and a visual module. 
The \textit{listener} agent additionally incorporates a third decision module that combines the outputs of the other two modules.
The RL agent similarly incorporates a third decision module with the addition that this third module contains a recurrent network, acting as core memory module for the agent. 
Using the Straight-Through Gumbel-Softmax (STGS) approach in the communication channel of the RG, the \textit{speaker} agent is prompted to produce the output string of symbols with a \textit{Start-of-Sentence} symbol and the visual module's output as an initial hidden state while the \textit{listener} agent consumes the string of symbols with the null vector as the initial hidden state. 
In the following subsections, we detail each module architecture in depth.

\textbf{Visual Module.} 
The visual module $f(\cdot)$ consists of an observation encoder (typically a convolutional neural network), optionally shared between agents, followed by some (agent-specific) adaptation layers in the form of a multi-layer perceptron (MLP). %, as shown in Figure~\ref{fig:ether-arch}.
The former consists of four blocks of a convolutional layers with $32$, $64$, $128$, and $256$ filters, strides of $8$, $4$, $4$, and $2$, each followed by a $2$D batch normalization layer and a ReLU non-linear activation function. 
%The two first convolutional layers have $32$ filters, whilst the last one has $64$. 
The bias parameters of the convolutional layers are not used, as it is common when using batch normalisation layers. 
Inputs are stimuli consisting of frames of the environment resized to $56\times56\times3$. 
%The RL agent's adaptation layers consist of $2$ FiLM layers \citep{perez2018film} conditioned on the \textit{linguistic goal description} $g\in\mathcal{G}$ after it is processed by the RL agent's language module.
The adaptation layers consists of $3$ fully-connected layers of $256$, $128$, and $64$ neurons, each followed by batch normalization layer and ReLU non-linear activation function.

\textbf{Language Module.} 
The language module $g(\cdot)$ consists of some learned Embedding followed by a one-layer GRU network~\citep{cho2014learning} of $64$ neurons in the case of the RL agent, $128$ neurons in the case of the RG agents. 
In the context of the \textit{listener} agent, the input message $m=(m_i)_{i\in[1,L]}$ (produced by the \textit{speaker} agent) is represented as a string of one-hot encoded vectors of dimension $|V|$ and embedded in an embedding space of dimension $64$ via a learned Embedding. The output of the \textit{listener} agent's language module, $g^l(\cdot)$, is the last hidden state of the RNN layer, $h^l_L = g^L(m_L, h^l_{L-1})$.
In the context of the \textit{speaker} agent's language module $g^S(\cdot)$, the output is the message $m=(m_i)_{i\in[1,L]}$ consisting of one-hot encoded vectors of dimension $|V|$, which are sampled using the STGS approach from a categorical distribution $Cat(p_i)$ where $p_i = Softmax(\nu(h^s_i))$, provided $\nu$ is an affine transformation and $h^s_i=g^s(m_{i-1}, h^s_{i-1})$. 
$h^s_0=f(s_t)$ is the output of the visual module, given the target stimulus $s_t$.

\textbf{Decision Module.} From the RL agent to the RG's listener agent, the decision module are very different since their outputs are either, respectively, in the action space $\mathcal{A}$ or the space of distributions over $K+1$ stimuli. 
For the RL agent, the decision module takes as input a concatenated vector comprising the output of the adaptation layers and some other information relevant to the RL context (e.g. previous reward and previous action selected, following the recipe in \citet{kapturowski2018recurrent}). 
The resulting concatenated vector is then fed to the core memory module, a one-layer LSTM network~\citep{hochreiter1997long} with $1024$ hidden units, which feeds into the advantage and value heads of a $1$-layer dueling network \citep{wang2016dueling}.

In the case of the RG's listener agent, similarly to \citet{Havrylov2017}, the decision module builds a probability distribution over a set of $K+1$ stimuli/images $(s_0, ..., s_K)$, consisting of $K$ distractor stimuli and the target stimulus, provided in a random order (see Figure~\ref{fig:ether-rg}), given a message $m$ using the scalar product:
\begin{align}
\label{eq:discr-stgs}
    p((d_i)_{i\in[0,K]} &| (s_i)_{i\in[0,K]} ; m) = \nonumber \\ 
    Softmax &\Bigr( ( h^l_L \cdot f(s_i)^T )_{i\in[0,K]} \Bigl).
\end{align}

\section{Further background on Emergent Communication}
\label{app:emecom-background}

\textbf{Compositionality \& Generalisation.}
As mentioned in the main text, alongside semantical grounding of emergent language, compositionality is an important property studied in EmeCom. Compositionality has been shown to further the learnability of said languages \citep{kirby2002learning, Smith2003, Brighton2002, Li&Bowling2019} and, on the other hand, natural languages' compositionality promises to increase the generalisation ability of the artificial agent that would be able to rely on them as a grounding signal, as it has been found to produce learned representations that generalise, when measured in terms of the data-efficiency of subsequent transfer and/or curriculum learning \citep{Higgins2017SCAN, Mordatch2017, MoritzHermann2017, Jiang2019-language-abstraction-hierarchical-rl}.
Nevertheless, some sufficient conditions can be found to further the emergence of compositional languages and generalising learned representations~\citep{Kottur2017, Lazaridou2018, Choi2018, Bogin2018, Guo2019, Korbak2019, Chaabouni2020, DenamganaiAndWalker2020b}.

The backbone of the field rests on games that emphasise the functionality of languages, namely, the ability to efficiently communicate and coordinate between agents. 
The first instance of such an environment is the \textit{Signaling Game} or \textit{Referential Game (RG)} by \citet{lewis1969convention}, where a speaker agent is asked to send a message to the listener agent, based on the \textit{state/stimulus} of the world that it observed. 
The listener agent then acts upon the observation of the message by choosing one of the \textit{actions} available to it in order to perform the `best' \textit{action} given the observed \textit{state} depending on the notion of `best' \textit{action} being defined by the interests common to both players.
In discriminative RGs, typically, the listener action is to discriminate between a target stimulus and some other distractor stimuli, conditioned solely on the speaker's message describing said target stimulus.
%The latter defined the discriminative variant, as opposed to the generative variant where the listener agent must reconstruct/generate the whole target stimulus (usually played with symbolic stimuli).
Visual (discriminative) RGs have been shown to be well-suited for unsupervised representation learning, either by competing with state-of-the-art self-supervised learning approaches on upstream classification tasks \citep{Dessi2021-emecom_as_ssl}, or because they have been found to further some forms of disentanglement~\citep{Higgins2018, Kim2018, Chen_2018-MIG, Locatello2020-cx} in learned representations \citep{Xu2022-COMPODIS,Denamganai2023visual-COMPODIS}. 
Such properties can enable ``better up-stream performance''\citep{Van_Steenkiste2019-xm}, greater sample-efficiency, and some form of (systematic) generalization~\citep{montero2021the, Higgins2017DARLA, Steenbrugge2018-sq}. 
Indeed, disentanglement is thought to reflect the compositional structure of the world, thus disentangled learned representations ought to enable an agent wielding them to generalize along those lines. 
%The work of \citet{Chaabouni2020} showed that, in the context of generative, symbolic (i.e. disentangled stimuli) referential games, the degree of compositionality of the emerging languages and the agents ability to generalize to zero-shot stimuli are not correlated, but (i) ``when a language is positionally disentangled (and, to a lesser extent, bag-of-symbols disentangled), it is very likely that the language will be able to generalize -- a guarantee we do not have from less informative topographic similarity'', and (ii) the data regime (e.g. low or high) is a better predictor for generalization (i.e. ``generalization emerges `naturally' if the input space if large'').
Thus, this paper aims to investigate visual discriminative RGs as auxiliary tasks for RL agents.

\section{Non-degeneracy of RG-induced relabelling and predicate functions}
\label{app:non_degeneracy}

This appendix proves Theorem~\ref{thm:non_degeneracy}, i.e. that the RG-induced pair $(m_{RG}, f_{RG})$ of Definition~\ref{def:m_rg_f_rg} escapes both failure modes gathered in $\mathcal{D}_\text{deg}$ (Definition~\ref{def:d_deg}). 
We first set the notation and the two consequences of optimality that the proof consumes below.

\textbf{Notation.}
For a stimulus $s$ and an utterance $u\in\Sigma^\star$, we write $z(s,u) = Li(s,u)$ for the raw compatibility score of the listener agent, and $\ell_\text{no-target}$ for the learnable scalar of Definition~\ref{def:descriptive_rg}.
In a descriptive RG, a round presents $K+1$ candidate stimuli and the decision module scores the $K+2$ classes
$( z(s_i, u) )_{i\in[0,K]} \cup \{ \ell_\text{no-target} \}$, as in Equation~\ref{eq:no_target_softmax}.
Writing $\mathcal{C}$ for the index set of those $K+2$ classes and $c^\star\in\mathcal{C}$ for the correct one, the Hinge loss of Equation~\ref{eq:hinge_loss} reads
\begin{equation}
\label{eq:hinge_descriptive}
\mathcal{L} = \max\Big(0, \; \nu - \text{score}(c^\star) + \underset{c\in\mathcal{C}\setminus\{c^\star\}}{\max} \, \text{score}(c) \Big)
\end{equation}
where $c^\star$ is the index of the target stimulus on \textit{target-present} rounds, and the no-target class on \textit{target-absent} rounds. 
Since $\mathcal{L}\geq0$ and $J_{RG}$ is its expectation over rounds (Equation~\ref{eq:j_rg}), the assumption $J_{RG}(Sp^\star, Li^\star)=0$ implies that $\mathcal{L}=0$ on every round of positive probability. 
Equation~\ref{eq:hinge_descriptive} with $\mathcal{L}=0$ is then equivalent to the margin condition: 
\begin{equation}
\label{eq:margin_condition}
\text{score}(c^\star) \geq \nu + \underset{c\in\mathcal{C}\setminus\{c^\star\}}{\max} \, \text{score}(c) \, .
\end{equation}

\begin{proof}[Proof of Theorem~\ref{thm:non_degeneracy}]
We show separately that $(m_{RG}, f_{RG}) \notin \mathcal{D}_\text{collapse}(\mathcal{D}_{RG}, c)$ for every $c<|\mathcal{D}_{RG}|$ (Part A) and that $(m_{RG}, f_{RG}) \notin \mathcal{D}_\text{trivial}(\mathcal{D}_{RG})$ (Part B). 
Since $\mathcal{D}_\text{deg}$ is the union of the two (Definition~\ref{def:d_deg}), this proves the claim. 

\textbf{Part A: $J_\text{RG}$-optimal $m_{RG}$ is not collapsed.} 
We operate via proof by contradiction: we assume that there exist $s\neq s'$ in $\mathcal{D}_{RG}$ with $m_{RG}(s) = m_{RG}(s')$, and write $u = Sp(s) = Sp(s')$ for their common utterance. 
Because the $K\geq1$ distractors of a round are drawn independently from $\mathcal{D}_{RG}$, the target-present round whose target is $s$ and whose distractors include $s'$ has positive probability; so does the round whose target is $s'$ and whose distractors include $s$. 
Applying Equation~\ref{eq:margin_condition} to the first round, with $c^\star$ the index of $s$ and $s'$ among the competing classes, gives $z(s,u) \geq \nu + z(s',u)$. 
Applying it to the round where the roles are exchanged and the speaker emits the same $u$, gives $z(s',u) \geq \nu + z(s,u)$. 
Summing the two inequalities yields $0 \geq 2\nu$, contradicting $\nu>0$. 

Hence $Sp^\star$ is injective on $\mathcal{D}_{RG}$, so $m_{RG}$ emits $|\{ m_{RG}(s) : s\in\mathcal{D}_{RG} \}| = |\mathcal{D}_{RG}|$ distinct goals, which is the maximum possible. 
By Equation~\ref{eq:d_collapse}, $(m_{RG}, f_{RG}) \notin \mathcal{D}_\text{collapse}(\mathcal{D}_{RG}, c)$ for every tolerance $c<|\mathcal{D}_{RG}|$. 
For the continuous form of Equation~\ref{eq:d_collapse_continuous}, injectivity gives $\mathbb{I}(m_{RG}(s_1)\neq m_{RG}(s_2)) = \mathbb{I}(s_1 \neq s_2)$, so that
\begin{equation}
\label{eq:diversity_value}
\mathop{\mathbb{E}}_{s_1,s_2\sim\mathcal{U}(\mathcal{D}_{RG})} \big[ \mathbb{I}(m_{RG}(s_1)\neq m_{RG}(s_2)) \big] = 1 - \frac{1}{|\mathcal{D}_{RG}|}
\end{equation}
again the maximum attainable, whence $(m_{RG}, f_{RG}) \notin \mathcal{D}_\text{collapse}(\mathcal{D}_{RG}, \delta)$ for every $\delta < 1 - \nicefrac{1}{|\mathcal{D}_{RG}|}$. 

Note that Part A uses only $\nu>0$, $K\geq1$ and the independence of the distractor draw: it holds whether or not the no-target class is present, since including $\ell_\text{no-target}$ among the competing classes of Equation~\ref{eq:margin_condition} only strengthens the two inequalities. 

\textbf{Part B: $J_\text{RG}$-optimal $f_{RG}$ is not trivial.} 
With $\beta=\nicefrac{1}{2}$, Definition~\ref{def:m_rg_f_rg} gives $f_{RG}(s,g) = \mathbb{I}( z(s,g) \geq \ell_\text{no-target} )$, as $\sigma$ is increasing and $\sigma(0)=\nicefrac{1}{2}$. 
It therefore suffices to exhibit a pair on which the score falls below $\ell_\text{no-target}$, and descriptiveness supplies one because it makes both kinds of round occur with positive probability. 

Consider a \textit{target-absent} round, in which the target is replaced by a descriptive distractor, so that the correct class $c^\star$ is the no-target extra class and all $K+1$ presented stimuli $(s_i)_{i\in[0,K]}$ are non-targets for the emitted utterance $u$. 
Equation~\ref{eq:margin_condition} then reads $\ell_\text{no-target} \geq \nu + \max_{i\in[0,K]} z(s_i, u)$, hence $z(s_i,u) < \ell_\text{no-target}$ (since $\nu>0$) and we have: 
\begin{equation}
\label{eq:f_rg_rejects}
f_{RG}(s_i, u) = 0 \qquad \forall i \in [0,K] \, .
\end{equation}
As $(s_i)_{i\in[0,K]} \in \mathcal{D}_{RG}$ and $u = Sp^\star(\cdot) \in \mathcal{G}\subset\Sigma^\star$, Equation~\ref{eq:f_rg_rejects} contradicts the condition of Equation~\ref{eq:d_trivial}, which requires $f(s,g)=1$ for \textit{all} pairs in $\mathcal{D}_{RG}\times\mathcal{G}$. 
Therefore $(m_{RG}, f_{RG}) \notin \mathcal{D}_\text{trivial}(\mathcal{D}_{RG})$. 

%Conversely, a \textit{target-present} round with target $s_0$ and utterance $u=Sp^\star(s_0)$ has the target as correct class and $\ell_\text{no-target}$ among its competitors, so Equation~\ref{eq:margin_condition} gives $z(s_0,u) \geq \nu + \ell_\text{no-target} > \ell_\text{no-target}$ and thus $f_{RG}(s_0, m_{RG}(s_0)) = 1$.
%The predicate function is consequently not merely non-vacuous but also \textit{consistent} with the relabelling function, in the sense that every state satisfies the goal it is relabelled with.
Finally, combining the two parts with the union operator over set yields $(m_{RG}, f_{RG})\notin \mathcal{D}_\text{deg}(\mathcal{D}_{RG}, c)$, which completes the proof.
\end{proof}

\textit{Remark (approximate optimality).}
The assumption $J_{RG}=0$ is an idealisation, so we record the corresponding approximate statement. 
Suppose $J_{RG}(Sp, Li) = \varepsilon > 0$, which renders collapsed pair possible: let $s\neq s'$ be such a collapsed pair, i.e. $Sp(s)=Sp(s')=u$. 
By Equation~\ref{eq:hinge_descriptive}, the round having $s$ as target and $s'$ as a distractor incurs $\mathcal{L} \geq \nu - z(s,u) + z(s',u)$, while the round with the target and distractor roles exchanged incurs $\mathcal{L} \geq \nu - z(s',u) + z(s,u)$. 
The two lower bounds sum to $2\nu$, so at least one of the two rounds incurs $\mathcal{L}\geq\nu$. 
Markov's inequality applied to $\mathcal{L}\geq0$ and $\nu>0$ gives $\mathbb{P}(\mathcal{L}\geq\nu) \leq \nicefrac{\varepsilon}{\nu}$, whence the probability that a round exhibits a collapsed pair in either role is at most $\nicefrac{2\varepsilon}{\nu}$. 
Collapse is therefore not merely excluded at the optimum but rare in proportion to the sub-optimality, degrading linearly in $\mathcal{O}(\varepsilon)$.

\section{Further details on HIGhER}
%\todo[inline]{provide appendix explanation of ACTRCE}
\label{app:higher}

\subsection{HIGhER and its Limitations.}
\label{subsec:higher-limitations}

HIGhER\citep{cideron2020higher} aims to expand the applicability of HER, and to do so it explores how to learn the re-labelling/mapping function (hereafter referred to as $m_{HIGhER}$ and Instruction Generator), rather than assuming it is provided or by using some form of external expert knowledge. 
Nevertheless, it still relies on a predicate function being provided and queried as an oracle.
%$f^\prime$ being \textit{partially} provided by the environment's reward signal upon termination of each episode, as we will see further below.
HIGhER investigates using hindsight experience replay in the instruction following setting from pixel-based observations, which brings some particularities as it differs from the robotic setting of HER. 
Firstly, the goal space and state space are no longer the same, hence the motivation towards learning a re-labelling/mapping function.
Secondly, there is no obvious mapping from stimuli (e.g. visual/pixel states) to the instructions that define the goals using a natural-like language. 
For instance, for a given state, due to the expressivity of natural languages, multiple goals may be defined as being fulfilled in this state. 

Despite the non-obvious mapping from stimuli to fulfilled goals, HIGhER still succeeds in learning a deterministic re-labelling/mapping function.
HIGhER learns an instruction generator by supervised learning on a dataset $\mathcal{D}_{sup}=\{(s,g) / f(s,g)=1\}$ consisting of state-goal pairs where the predicate value is know to be $1$. 
These pairs are harvested from successful trajectories of the RL agent which occur throughout the learning process (they could be provided as demonstrations, but this is not explored by the original work of \citet{cideron2020higher}) and correspond to final states $s$ of successful/positive-reward trajectories along with relevant linguistic instructions defining the fulfilled goals $g$. 
The ability to harvest successful trajectories from an RL agent in the process of being trained is capped by how likely is it that this RL agent will fulfill goals albeit while randomly/cluelessly exploring the environment.
Thus, one major limitation of HIGhER is that in the absence of initial (and therefore random - when harvested from the learning RL agent) successful trajectories, the dataset $\mathcal{D}_{sup}$ cannot be built, and it ensues that the hindsight experience replay scheme cannot be leveraged since the instruction generator/mapping function, $m_{HIGhER}$, cannot begin to learn.

Finally, it is important to note that HIGhER is constrained to using only the \textbf{final} re-labelling strategy.
Recall that the Instruction Generator is solely trained on episode's final state, and it is likely that the distribution of final states over the whole state space $\mathcal{S}$ is far from being uniform.
Thus, applying the re-labelling/mapping function of HIGhER on states encountered in the middle of an episode is tantamount to out-of-distribution application and would likely result in unpredictable re-labelling mistakes.
In the original work of \citet{cideron2020higher}, this particularity is not addressed, and only the \textbf{final} re-labelling strategy is experimented with.

%which only consist of achieved goal descriptions, i.e. the ones defined by the instruction following tasks.
%In other words, the expressivity of the re-labelling/mapping function is artificially capped by the design choices made by the creators of the instruction following tasks. 

\subsection{Extending HIGhER's Applicability}
\label{sec:extending-higher}

%\todo[inline]{ETHER builds on HIGHER by adding three main novelties: (1) a semantic co-occurrence auxiliary task (2) an improvement on sample efficiency at learning a relabelling function based on contrastive learning and (3) the addition of a referential game pre-training to obtain a predicate function. Novelty (1) allows ETHER to extract communication-grounding learning signals from unsuccessful trajectories whereas (3) drops the assumption of a ground-truth predicate function which tells us if a state satisfies a given goal.}

As explained in Section~\ref{subsec:higher-limitations}, HIGhER, like HER, still relies on a predicate function being provided and queried as an oracle.
In the following, we propose to address this limitation in two ways.
We first propose to derive a partial predicate function from the re-labelling function in the Section \ref{subsubsec:deriving_predicate_function}.
Then, we show how to enhance the quality of the derived predicate function with a contrastive learning approach in Section \ref{subsubsec:enhancing-predicate-function-with-contrastive-learning}.

\subsubsection{Deriving a Predicate Function}
\label{subsubsec:deriving_predicate_function}

Because HIGhER only implements the \textbf{final} relabelling strategy, we remark that the predicate function is only necessary in order to compute the output of the \textit{learning reward function} when it is fed a state $s_t$ from an unsuccessful episode and the relabelling goal $g^\prime\in\mathcal{G}$, such that $g^\prime = m(s_{t_{final}})$. 
Notably, this new goal $g^\prime$ could have also been reached in previous steps of the same trajectory, and if so, those transitions should also feature a positive reward like the final state of the episode. 
This can be achieved by applying $m$ to all states in the trajectory and giving a positive reward \textit{if and only if} the new goal for a given state matches that of the relabelled goal for the last state $s_{last}$.
%, as per Equation~\ref{eq:deriving_predicate_function}:
%
%\begin{equation}
%r_t =
%    \begin{cases}
%    1 & \text{if } m(s_t) \equiv m(s_{t_{final}}) \\
%    0 & \text{otherwise} \\
%    \end{cases}
%    \label{eq:deriving_predicate_function}
%\end{equation}
%
This procedure derives a predicate function from a re-labelling function.
In the remainder of the paper, we will denote this extension as HIGhER$+$.

It is important to note that this procedure is not as sound as it could be because it makes the implicit and erroneous assumption that fulfilled goals are deterministic and unique for each state, whereas, firstly, the expressivity of natural language allows many different ways of expressing a similar semantic for a goal that would have been fulfilled in a given state (resulting in different theoretically-valid values for $m(s_t)$ and $m(s_{t_{final}})$) and, secondly, for any given state a distribution of fulfilled goals (with different semantics, not just synonymous expressions) could be defined.
For instance, whenever a ``pick up the blue ball'' can be identified as being fulfilled in a given state, then the goal ``pick up a blue object'', or ``pick up a ball'' are all as valid as the former. 

\subsubsection{Enhancing the Predicate Function with Contrastive Learning}
\label{subsubsec:enhancing-predicate-function-with-contrastive-learning}

Let us assume that we have access to a relabelling function, either learnt or given. 
Successful trajectories, those that yield a positive environment reward, are ground-truth indicators of a goal being satisfied on the last visited state. 
In contrast, the states in the trajectory leading up the goal-fulfilling state do not satisfy the same condition as the last state, as otherwise those transitions would receive a positive reward according to the environment's goal-conditioned reward signal. 
We exploit this structure to use contrastive learning methods. 

HIGhER learns an instruction relabelling function by making use of a dataset of (state, goal) pairs, as defined in Section~\ref{sec:background}. 
We further increase the accuracy of the learnt re-labelling function via contrastive learning where the positive examples are the same (state, goal) pairs in dataset $\mathcal{D}$ and the negative examples are defined as follows.
Let $T_{final}$ be the timestep of the final transition. 
Negative examples consist of pairs of states $S_{(T_{final} - i)}$ for $i \in [1, n]$ and their associated negative goal.
This negative goal is built contrastively to the true re-labelling goals as $G_{neg} = EoS$, i.e. using the End of Sentence (EoS) symbol. 
We use $EoS$ to trivially satisfy that the negative goal $G_{neg}$ differs from the goal of the positive example $g$. 
\subsection{On HIGhER's Ablation Study}
\label{sec:higher-ablation-study}

Prior to the architectures described in the main part of the paper, we iterated over many designs and induction biases.
%Firstly, we experimented with R2D2's features such as the burn-in feature.
Notably we experimented with R2D2's burn-in feature.

Table~\ref{tab:PickUp-higher-no-burnin} shows the success ratios of HIGhER agents without burn-in feature against baseline R2D2 without burn-in feature on the modified (one-pick-up) PickUpDist-v0 task from the BabyAI benchmark at the end of the $200k$ observation budget.
The results show that the contrastive learning scheme for the predicate function is rather hurting performance compared to HIGhER+, while still being above baseline.
The burn-in feature provides the RL agent better sample-efficiency by stabilising the training of the recurrent network in the architecture.
While the instruction generator/speaker agent is being trained, the resulting goal re-labelled experiences that enters the replay buffer are presumably non-stationary.
Thus, we attribute the lower performance of the above architectures to the fact that they struggle to deal with the non-stationarity of the goal re-labelled experiences in the absence of the stabilising burn-in feature. 

\begin{table}[b]%{l}{0.6\textwidth}
\caption{Success ratios (mean and standard deviation) for agents without the burn-in feature of R2D2 after 200k steps in a modified version of the BabyAI PickUpDist-v0 task. 3 random seeds for each agent.}
\label{tab:PickUp-higher-no-burnin}
\centering
\begin{tabular}{@{}lccc@{}} 
%\begin{tabular}{c|c|c}
\toprule
    \textbf{Agent} & \textbf{Mean} & \\ \hline
    R2D2 (no Burn-In) & 13.02 $\pm$ 1.26 & \\ \hline
    \midrule
    HIGhER+ (no Burn-In) & 16.02 $\pm$ 1.79 & \\ \hline
    HIGhER++ (n=1) (no Burn-In) & 14.97 $\pm$ 1.19 & \\ \hline
    HIGhER++ (n=2) (no Burn-In) & 15.89 $\pm$ 0.60 & \\ \hline
    HIGhER++ (n=4) (no Burn-In) & 13.93 $\pm$ 2.29 & \\
    \bottomrule
    \end{tabular}
\end{table}

\newpage
\section{On algorithmic details of ETHER}
\label{sec:ether-details}

In this section, we detail how ETHER is built, and contrast it with HIGhER, from an algorithmic point of view.
We start by presenting in Algorithm~\ref{alg:HIGhER} an extended version of the pseudo-code for the HIGhER algorithm from \citet{cideron2020higher} with the following additions:

\begin{enumerate}
    \item (i) contrasting further the \textbf{learning} vs \textbf{behavioural} reward function concerns that we highlighted in Section~\ref{subsec:her-limitations}, 
    \item (ii) flagging the reliance on the \textbf{learning} reward function that depends on the predicate function, which is provided as an oracle in both HER and HIGhER.
\end{enumerate}

Then, following the added nuances to the HIGhER algorithm, we can now show in greater and contrastive details the ETHER algorithm in Algorithm~\ref{alg:ETHER}, where we highlight the following:

\begin{enumerate}
    \item (i) the RG training can be done in parallel at any time, thus we present it in the most-inner loop of the algorithm,
    \item (ii) since ETHER trains its RG speaker and listener agents on the whole state space $\mathcal{S}$, the ability to perform either \textbf{final} or \textbf{future-k} re-labelling strategy is recovered. We present the case of the \textbf{future-k} re-labelling strategy below.
\end{enumerate}

{\footnotesize
\begin{algorithm}[h]
\caption{Hindsight Generation for Experience Replay (HIGhER)}
\label{alg:HIGhER}
\SetKwInOut{Given}{Given}
\SetKwInOut{Initialize}{Initialize}
\SetKwRepeat{Do}{repeat}{until}
\Given{
\begin{itemize}
    \item an off-policy RL algorithm (e.g., DQN, R2D2) and its replay buffer $R$, 
    \item a behavioural policy $\pi_{behaviour}: \mathcal{S} \times \mathcal{G} \rightarrow \mathcal{A}$
    \item a \textbf{learning} reward function $r: \mathcal{S} \times \mathcal{A} \times \mathcal{S} \times \mathcal{G} \rightarrow \mathbb{R}$ {\color{red}(oracle or learned - relying on the predicate function $f: \mathcal{S} \times \mathcal{G} \rightarrow \{0,1\}$)}, 
    \item a \textbf{behavioural} reward function $r: \mathcal{S} \times \mathcal{A} \times \mathcal{S} \times \mathcal{G} \rightarrow \mathbb{R}$ (provided by the environment), 
    \item a language scoring function (e.g., parser accuracy, BLEU, etc.).
\end{itemize}
}
\Initialize{ 
\begin{itemize}
    %\item the behavioural policy $\pi_{behaviour}$, 
    %\item the replay buffer $R$, 
    \item the dataset $\mathcal{D}_{sup}$ of $(state, goal)$ pairs and a train-test split strategy to yield $\mathcal{D}_{sup/train}$ and $\mathcal{D}_{sup/val}$, 
    \item the Instruction Generator $m_{HIGhER}$.
\end{itemize}
}
\For{$episode = 1, M$}{
    Sample a goal $g$ and an initial state $s_0$\ from the environment\;
    $t = -1$\;
    \Do{(episode ends)}{
        $t = t + 1$\;
        Execute an action $a_t$ chosen from the behavioural policy $\pi_{behavioural}$\;
        Observe a new state $s_{t+1}$ and a {\color{purple}\textbf{behavioural} reward $r_t = r_{behavioural}(s_t, a_t, g)$}\;
        Store the transition $(s_t, a_t, r_t, s_{t+1}, g)$ in $R$\;
        Update Q-network parameters using the policy $\pi_{behavioural}$ and sampled minibatches from $R$\;
    }
    \If{{\color{blue} \textbf{learning} reward $r_\text{learn}(s_t, a_t, s_{t+1}, g) = f(s_{t+1}, g) = = 1$}}{
        Store the pair $(s_{t+1}, g)$ in $\mathcal{D}_{sup/train}$ or $\mathcal{D}_{sup/val}$\;
        Update $m_{HIGhER}$ parameters by sampling minibatches from $\mathcal{D}_{sup/train}$\;
    }
    \ElseIf{$m_{HIGhER}$ validation score is high enough \& $\mathcal{D}_{sup/val}$ is big enough}{
        %Use the \textbf{final} re-labelling strategy as follows...\;
        Duplicate the previous episode's transitions in $R$\;
        Sample $\hat{g}^0 = m_{HIGhER}(s_{t+1})$\;
        %Replace $g$ by $\hat{g}^0$ in {\color{red} \textbf{ all the duplicated transitions} of the last episode}\;
        Compute the {\color{blue}\textbf{learning} rewards $\forall t, \hat{r}^0_t = r_\text{learn}(s_t, a_t, s_{t+1}, \hat{g}^0) = f(s_{t+1}, \hat{g}^0)$}\;
        Replace $g$ by $\hat{g}^0$ and $r_t$ by $\hat{r}^0_t$ in {\color{red}\textbf{all the duplicated transitions} of the last episode}\;
    }
}
\end{algorithm}

\begin{algorithm*}%[h]
\caption{Emergent Textual Hindsight Experience Replay (ETHER)}
\label{alg:ETHER}
\SetKwInOut{Given}{Given}
\SetKwInOut{Initialize}{Initialize}
\SetKwRepeat{Do}{repeat}{until}
\Given{
\begin{itemize}
    \item an off-policy RL algorithm (e.g., DQN, R2D2) and its replay buffer $R$, 
    \item a behavioural policy $\pi_{behaviour}: \mathcal{S} \times \mathcal{G} \rightarrow \mathcal{A}$
    \item a descriptive, discriminative RG algorithm, with its dataset buffer $\mathcal{D}_{RG}$ and its listener and speaker agents\;
    \item a \textbf{learning} reward function $r: \mathcal{S} \times \mathcal{A} \times \mathcal{S} \times \mathcal{G} \rightarrow \mathbb{R}$ {\color{green}( relying on the predicate function $f: \mathcal{S} \times \mathcal{G} \rightarrow \{0,1\}$ which is implemented via the RG's listener agent)}, 
    \item a \textbf{behavioural} reward function $r: \mathcal{S} \times \mathcal{A} \times \mathcal{S} \times \mathcal{G} \rightarrow \mathbb{R}$ (provided by the environment), 
    \item a language scoring function {\color{blue}(implemented via the RG's accuracy on the validation set)}.
\end{itemize}
}
\Initialize{ 
\begin{itemize}
    %\item the behavioural policy $\pi_{behaviour}$, 
    %\item the replay buffer $R$, 
    \item the dataset $\mathcal{D}_{sup}$ of $(state, goal)$ pairs and a train-test split strategy to yield $\mathcal{D}_{sup/train}$ and $\mathcal{D}_{sup/val}$, 
    \item the RG dataset $\mathcal{D}_{RG}$ of stimuli $state$ and a train-test split strategy to yield $\mathcal{D}_{RG/train}$ and $\mathcal{D}_{RG/val}$, 
    \item the Instruction Generator {\color{blue} $m_{ETHER}(\cdot)$, in the form of the RG's speaker agent}.
    \item {\color{blue} the learned predicate function $f_{ETHER}(\cdot)$, in the form of the RG's listener agent},
    \item {\color{blue} $K_{HER}\in\mathbb{N}$ specifying which re-labelling strategy to use (if $K_{HER}=0$ then \textbf{final}, otherwise \textbf{future-$K_{HER}$}).}
\end{itemize} 
}
\For{$episode = 1, M$}{
    Sample a goal $g$ and an initial state $s_0$\ from the environment\;
    $t = -1$\;
    \Do{episode ends}{
        $t = t + 1$\;
        Execute an action $a_t$ chosen from the behavioural policy $\pi_{behavioural}$\;
        Observe a new state $s_{t+1}$ and a {\color{purple}\textbf{behavioural} reward $r_t = r_{behavioural}(s_t, a_t, g)$}\;
        Store the transition $(s_t, a_t, r_t, s_{t+1}, g)$ in $R$\;
        Update Q-network parameters using the policy $\pi_{behavioural}$ and sampled minibatches from $R$\;
        {\color{blue} Store the stimulus $s_t$ in $\mathcal{D}_{RG/train}$ or $\mathcal{D}_{RG/val}$}\;
        {\color{blue} Update the RG's speaker and listener agents by playing $N_{RG}$ epochs of the RG, training on $\mathcal{D}_{RG/train}$ and performing evaluation on $\mathcal{D}_{RG/val}$}\;
    }
    \If{{\color{blue} \textbf{learning} reward $r_\text{learn}(s_t, a_t, s_{t+1}, g) = f_{ETHER}(s_{t+1}, g) = = 1$}}{
        Store the pair $(s_{t+1}, g)$ in $\mathcal{D}_{sup/train}$ or $\mathcal{D}_{sup/val}$\;
        Update {\color{blue}the RG's speaker agent parameters (ETHER) with supervised learning} by sampling minibatches from $\mathcal{D}_{sup/train}$\;
    }
    \ElseIf{\color{blue} RG validation accuracy on $\mathcal{D}_{RG/val}$ is high enough}{
        {\color{blue} Use the \textbf{future-$K_{HER}$} re-labelling strategy as follows...}\;
        $k = 0$, $T =$ last episode's length\;
        \Do{$k = = K_{HER}$}{
            Sample $T_{k}$ uniformly from $[1,T]$\; 
            Duplicate the previous episode's transitions in $R$, until sampled timestep $T_{k}$\;
            Sample $\hat{g}^0 = m_{ETHER}(s_{T_k})$\;
            %Replace $g$ by $\hat{g}^0$ in {\color{red} \textbf{ all the duplicated transitions} of the last episode}\;
            Compute the {\color{blue}\textbf{learning} rewards $\forall t, \hat{r}^0_t = r_\text{learn}(s_t, a_t, s_{t+1}, \hat{g}^0) = f_{ETHER}(s_{t+1}, \hat{g}^0)$}\;
            Replace $g$ by $\hat{g}^0$ and $r_t$ by $\hat{r}^0_t$ in {\color{red}\textbf{all the duplicated transitions} of the last episode}\;
            $k = k + 1$\;
        }
    }
}
\end{algorithm*}
}

\section{On the Referential Game in ETHER}
\label{sec:ether-rg}

In the following, we detail further the referential game (RG) used in the ETHER architectures.

As highlighted in Section~\ref{subsec:emecom}, we follow the nomenclature proposed in \citet{DenamganaiAndWalker2020a} and focus on a \textit{descriptive, discriminative, object-centric, (partially-observable) $2$-players/$L=10$-signal/$N=0$-round/$K=64$-distractor} RG variant, as illustrated in Figure~\ref{fig:ether-rg}.
%We focus primarily on a \textit{descriptive object-centric (partially-observable) $2$-players/$L=10$-signal/$N=0$-round/$K=31$-distractor} RG variant, as illustrated in Figure~\ref{fig:ether-rg}.

As an object-centric RG, as opposed to stimulus-centric, the listener and speaker agents are not being presented with the same exact target stimuli. 
Rather, they are being presented with different \textit{viewpoints} on the same target object shown in the target stimuli, where the word \textit{viewpoint} ought to be understood in a large sense. 
Indeed, object-centrism is implemented by applying data augmentation schemes such as gaussian blur, color jitter, and affine transformations to visual stimuli~\citep{Choi2018,Dessi2021-emecom_as_ssl}. 
Thus, the listener and speaker agents are presented with different stimuli that nevertheless keeps the conceptual object being presented constant.
Firstly introduced by \citet{Choi2018} (without it being of primary interest), where the pair of agents would literally be shown potentially the same 3D objects under different viewpoint, thus thinking of object-centrism as a \textit{viewpoint} shift is historically relevant.

We parameterise the communication channel with a Straight-Through Gumbel-Softmax (STGS) estimator~\citep{Havrylov2017,DenamganaiAndWalker2020b}. 
The vocabulary $\Sigma$ is fixed with $62$ ungrounded symbols, plus two grounded symbol accounting for the \textit{Start-of-Sentence} and \textit{End-of-Sentence} semantics, thus $|\Sigma|=64$. 
The maximum sentence length $L$ is always equal to $10$, thus placing our experiments in the context of an overcomplete communication channel whose capacity is far greater than the number of different meanings that the agents would encounter in our experiments~\citep{Kottur2018}.

While we focus here exclusively on the STGS parameterisation, many other communication channel could have been used (e.g. REINFORCE-based algorithms~\citep{williams1992simple}, quantization~\citep{carmeli2022quantizedEmeCom} or Obverter approaches~\citep{Choi2018,Bogin2018}).
Indeed, the STGS approach supposedly allows a richer signal towards solving the credit assignment problem that language emergence poses, since the gradient can be backpropagated from the listener agent to the speaker agent.
%, while, in comparison, it cannot be backpropagated when using more commonly adopted approaches based on. 
%This is due to the fact that the gradient can be backpropagated through the listener agent to the speaker agent, thus giving the speaker agent a gradient related to the listener agent's own inner 'reasoning'. 
%\todo[inline]{Having described the RG setup, the following section provides details on the architecture of the \textit{speaker} and \textit{listener} agents and the dataset used
%\footnote{For more details, please refer to our code released at: \url{https://github.com/Near32/ReferentialGym/tree/develop/zoo/referential-games\%2Bcompositionality\%2Bdisentanglement}.}.
%}
%\todo[inline]{Linguistic Functions. Discuss \citet{Wu2021-entropy-decomposition}'s entropy decomposition trick to align with \citet{jakobson1960linguistics}'s functions of language, and how \citet{Lowe2019}'s concepts of positive signalling and positive listening factors in, in order to emphasise how our ETHER builds over each of those.}
%\todo[inline]{Discuss that ETHER allows using the future strategy from HER, whereas THER does not ; thanks to meaningful rewards along the way of multi-objective instructions, e.g. opening doors before picking up an object.}
The descriptiveness implies that the target stimulus may not be passed to the listener agent, but instead replaced with a descriptive distractor.
In effect, the listener agent's decision module therefore outpus a $K+2$-logit distribution where the $K+2$-th logit represents the meaning/prediction that none of the $K+1$ stimuli is the target stimulus that the speaker agent was `talking' about. 
The addition is made following \citet{Denamganai2023visual-COMPODIS} as a learnable logit value, $logit_{no-target}$, it is an extra parameter of the model. 
In this case the decision module output is no longer as specified in Equation~\ref{eq:discr-stgs}, but rather as follows:
\begin{align}
\label{eq:descr-discr-stgs}
p((d_i)_{i\in[0,K+1]} | &(s_i)_{i\in[0,K]}; m) = Softmax \Bigl( \nonumber \\
( h^l_L \cdot f(s_i)^T &)_{i\in[0,K]} \cup \{\textcolor{red}{logit_\text{no-target}}\} 
\Bigr)
\end{align}
%The descriptiveneness is ideal but not necessary in order to employ the listener agent as a predicate function for the context of HER.
%Thus, in the main results of the paper, we present the version without descriptiveness.

In the remainder of this section, we detail the STGS-LazImpa loss that we employed in our referential game, as illustrated in Figure~\ref{fig:ether-rg}.

%\todo[inline]{Linguistic Functions. Discuss \citet{Wu2021-entropy-decomposition}'s entropy decomposition trick to align with \citet{jakobson1960linguistics}'s functions of language, and how \citet{Lowe2019}'s concepts of positive signalling and positive listening factors in, in order to emphasise how our ETHER builds over each of those.}

%\todo[inline]{Discuss that ETHER allows using the future strategy from HER, whereas THER does not ; thanks to meaningful rewards along the way of multi-objective instructions, e.g. opening doors before picking up an object.}

\subsection{STGS-LazImpa Loss}
\label{subsec:stgs-lazimpa}

Emergent languages rarely bears the core properties of natural languages \citep{Kottur2017,Bouchacourt2018, Lazaridou2018, Chaabouni2020}, such as Zipf’s law of Abbreviation (ZLA). 
In the context of natural languages, this is an empirical law which states that the more frequent a word is, the shorter it tends to be~\citep{zipf2016human-zla, strauss2007word-zla}.
\citet{rita2020lazimpa} proposed LazImpa in order to make emergent languages follow ZLA.

To do so, Lazimpa adds to the speaker and listener agents some constraints to make the speaker lazy and the listener impatient.
Thus, denoting those constraints as $\mathcal{L}_{STGS-lazy}$ and $\mathcal{L}_{impatient}$, we derive the STGS-LazImpa loss that adapts LazImpa from REINFORCE-based communication channels to STGS-based ones, as follows:
\begin{align}
\label{eq:stgs-lazimpa-loss}
\mathcal{L}_{STGS-LazImpa} (m, (s_i)_{i\in[0,K]}) &= \nonumber \\
\mathcal{L}_{STGS-lazy}(m) + \mathcal{L}_{impatient}&(m, (s_i)_{i\in[0,K]})
\end{align}

In the following, we detail those two constraints.

\textbf{Lazy Speaker.} The Lazy Speaker agent has the same architecture as common speakers. The ‘Laziness’ is originally implemented as a cost on the length of the message $m$ directly applied to the loss, of the following form:

\begin{equation}
\label{eq:lazy-loss}
\mathcal{L}_{lazy}( m ) = \alpha(acc)|m|
\end{equation}
where $acc$ represents the current accuracy estimates of the referential games being played, and $\alpha$ is a scheduling function, which is not differentiable. 
This is aimed to adaptively penalize depending on the message length.
Since the lazyness loss is not differentiable, they ought to employ a REINFORCE-based algorithm for the purpose of credit assignement of the speaker agent.

In this work, we use the STGS communication channel, which has been shown to be more sample-efficient than REINFORCE-based algorithms~\citep{Havrylov2017}, but it requires the loss functions to be differentiable.
Therefore, we modify the lazyness loss by taking inspiration from the variational autoencoders (VAE) literature~\citep{Kingma2013-VAE}.

The length of the speaker's message is controlled by the appearance of the EoS token, wherever it appears during the message generation process that is where the message is complete and its length is fixed.
Symbols of the message at each position are sampled from a distribution over all the tokens in the vocabulary that the listener agent outputs.
Let $(W_l)$ be this distribution over all tokens $w\in \Sigma$ at position $l\in [1,L]$, such that $\forall l\in[1,L],\, m_l \sim (W_l)$. 
We devise the lazyness loss as a Kullbach-Leibler divergence $D_{KL}(\cdot | \cdot)$ between these distribution and the distribution $(W_{EoS})$ which attributes all its weight on the EoS token.
Thus, we dissuade the listener agent from outputting distributions over tokens that deviate too much from the EoS-focused distribution $(W_{EoS})$, at each position $l$ with varying coefficients $\beta(l)$.
The coefficient function $\beta: [1,L] \rightarrow \mathbb{R}$ must be monotically increasing.
We obtain our STGS-lazyness loss as follows:
\begin{align}
\label{eq:stgs-lazyness-loss}
\mathcal{L}_{STGS-lazy}(m) &= \nonumber \\
\sum_{l\in[1,L]} 
\beta(l)
D_{KL} \Bigr( 
&(W_{EoS}) |
(W_l)
\Bigl)
\end{align}

\textbf{Impatient Listener.} Our implementation of the Impatient
Listener agent follows the original work of \citet{rita2020lazimpa}: it is designed to guess the target stimulus as soon as possible, rather than solely upon reading the EoS token at the end of the speaker's message $m$. 
Thus, following Equation~\ref{eq:discr-stgs}, the Impatient Listener agent outputs a probability distribution over a set of $K+1$ stimuli $(s_0, ..., s_K)$ for all sub-parts/prefixes of the message $m=(m_1,...,m_l)_{l\in[1,L]}=(m_{\leq l})_{l\in[1,L]}$,  for all $l \in [1,L]$:
\begin{align}
\label{eq:impatient-discr-stgs}
% \forall l \in [1,L], \;\;
p( \mathbf{(d^{\leq l}_i)_{i\in[0,K]}} | &(s_i)_{i\in[0,K]} ; \mathbf{m^{\leq l}}) = \nonumber \\ 
Softmax \Bigr( 
( \mathbf{h_{\leq l}} &\cdot f(s_i)^T )_{i\in[0,K]} 
\Bigl)
\end{align}
where $\mathbf{h_{\leq l}}$ is the hidden state/output of the recurrent network in the language module (cf. Section~\ref{app:model-architecture}) after consuming tokens of the message from position $1$ to position $l$ included.

Thus, we obtain a sequence of $L$ probability distributions, which can each be contrasted, using the loss of the user's choice, against the target distribution $(D_{target})$ attributing all its weights on the decision $d_{target}$  where the target stimulus was presented to the listener agent.
Here, we employ \citet{Havrylov2017}'s Hinge loss.
Denoting it as $\mathbb{L}(\cdot)$, we obtain the impatient loss as follows:
\begin{align}
\label{eq:impatient-loss}
\mathcal{L}_{impatient/\mathbb{L}}( m, &(s_i)_{i\in[0,K]}) = \nonumber \\
\frac{1}{L}
\sum_{l\in[1,L]}
\mathbb{L}( (&d^{\leq l}_{i\in[0,K]}, (D_{target}) ).
\end{align}

\newpage
\section{On the Semantic Co-Occurrence Hypothesis}
\label{sec:co-occurrence}

In Section~\ref{subsec:co-occurrence}, we hypothesised that, upon specifiying a goal, agent observations would be biased to contain semantic components present in said goal. 
We tested this hypothesis in the BabyAI environment and provided some examples in Figures~\ref{fig:co_occurrence:expert} and ~\ref{fig:co_occurrence:random}, when the linguistic goal description was ``Pick up the green key'', showing that the semantics of the goal (colour ``green'', shape ``key'') are some of the most salient observed semantics in the environment's observations of expert trajectories and random walks.

\begin{figure*}[htbp]
\centering
%\vspace{-20pt}
\begin{subfigure}{0.4\linewidth}
    \centering
    \includegraphics[width=\linewidth]{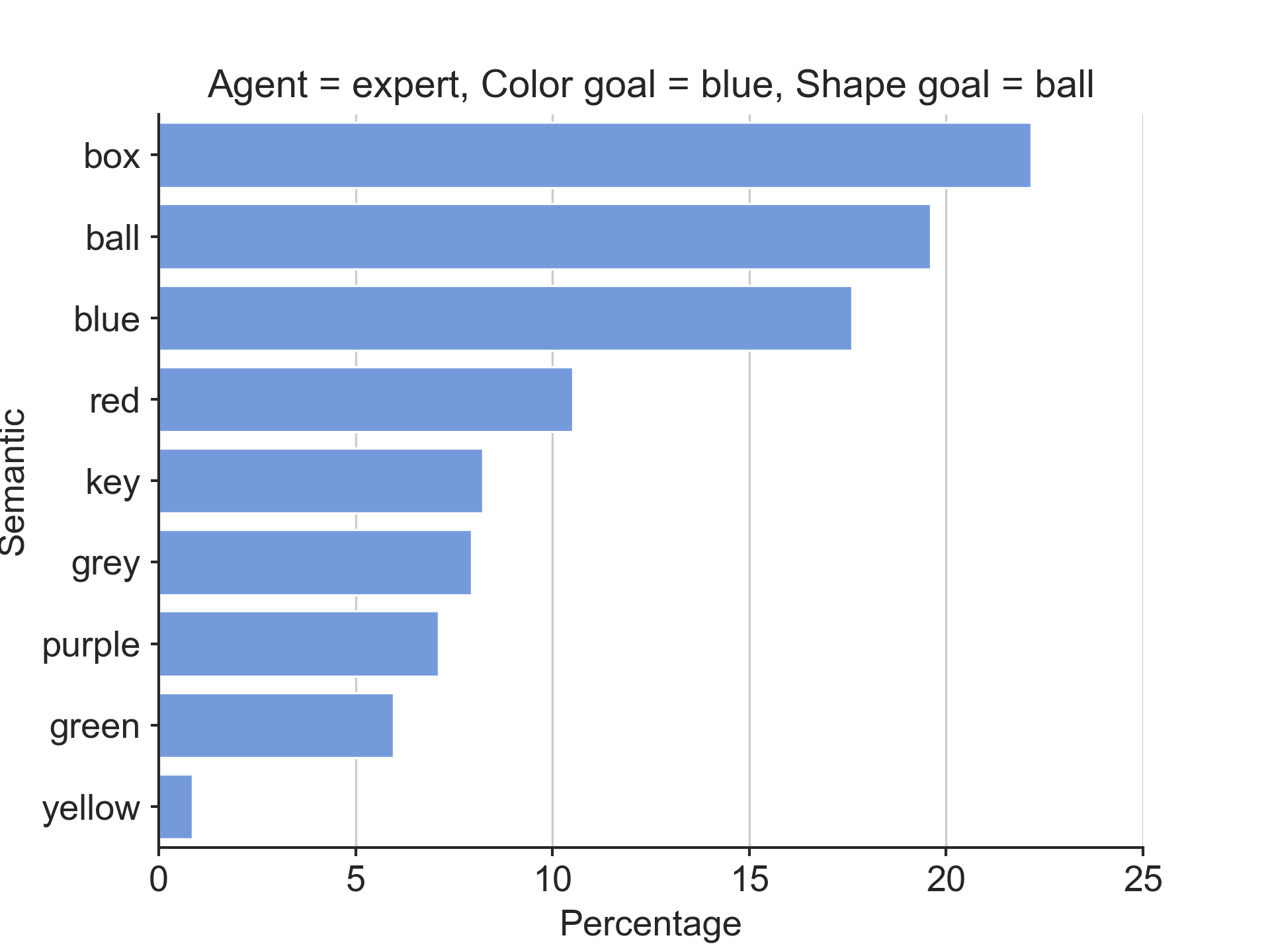}
    %\caption{} % Empty caption preserves layout counter structure
\end{subfigure}\hfill
\begin{subfigure}{0.4\linewidth}
    \centering
    \includegraphics[width=\linewidth]{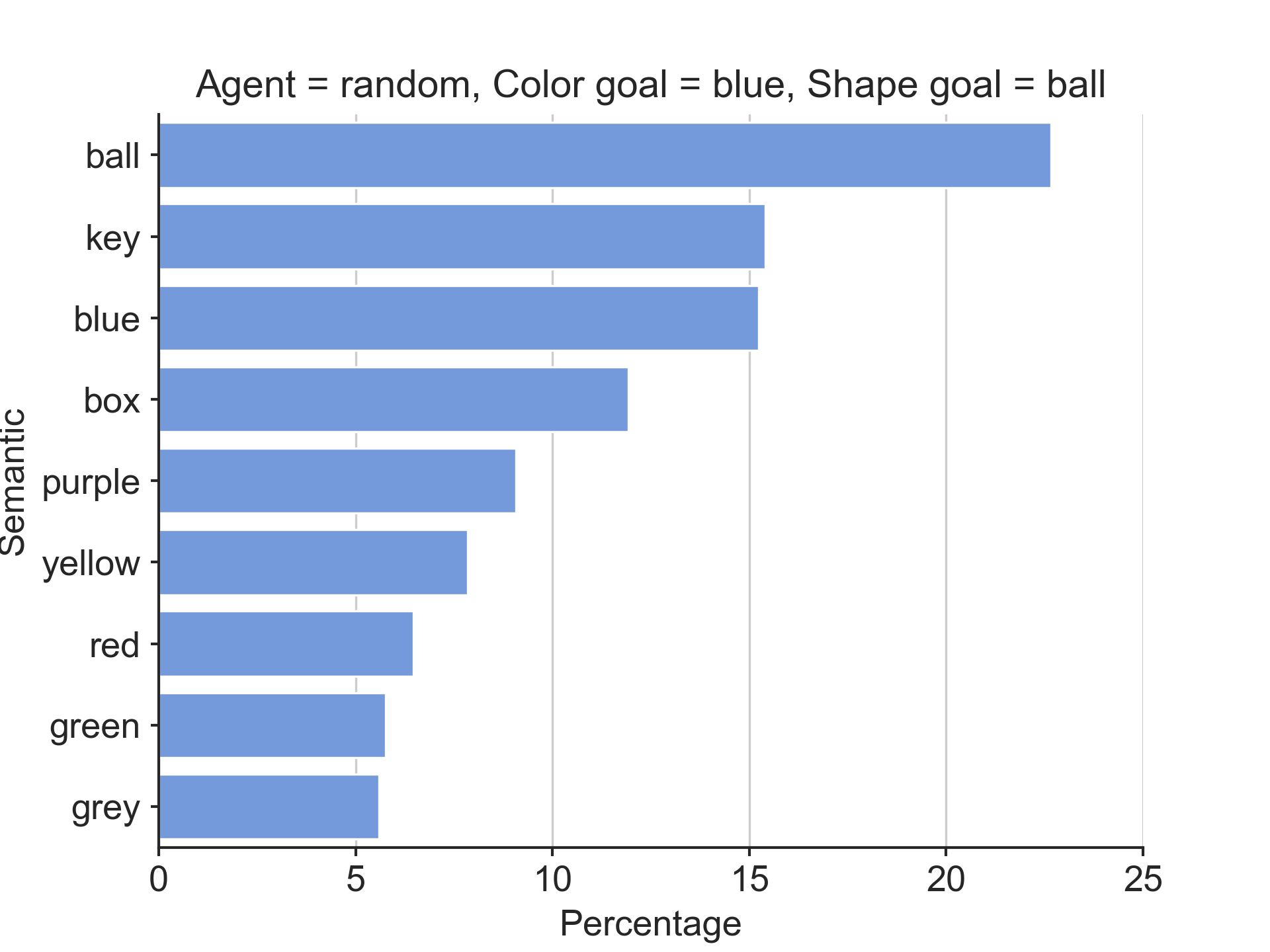}
    %\caption{} % Empty caption preserves layout counter structure
\end{subfigure}
\begin{subfigure}{0.4\linewidth}
    \centering
    \includegraphics[width=\linewidth]{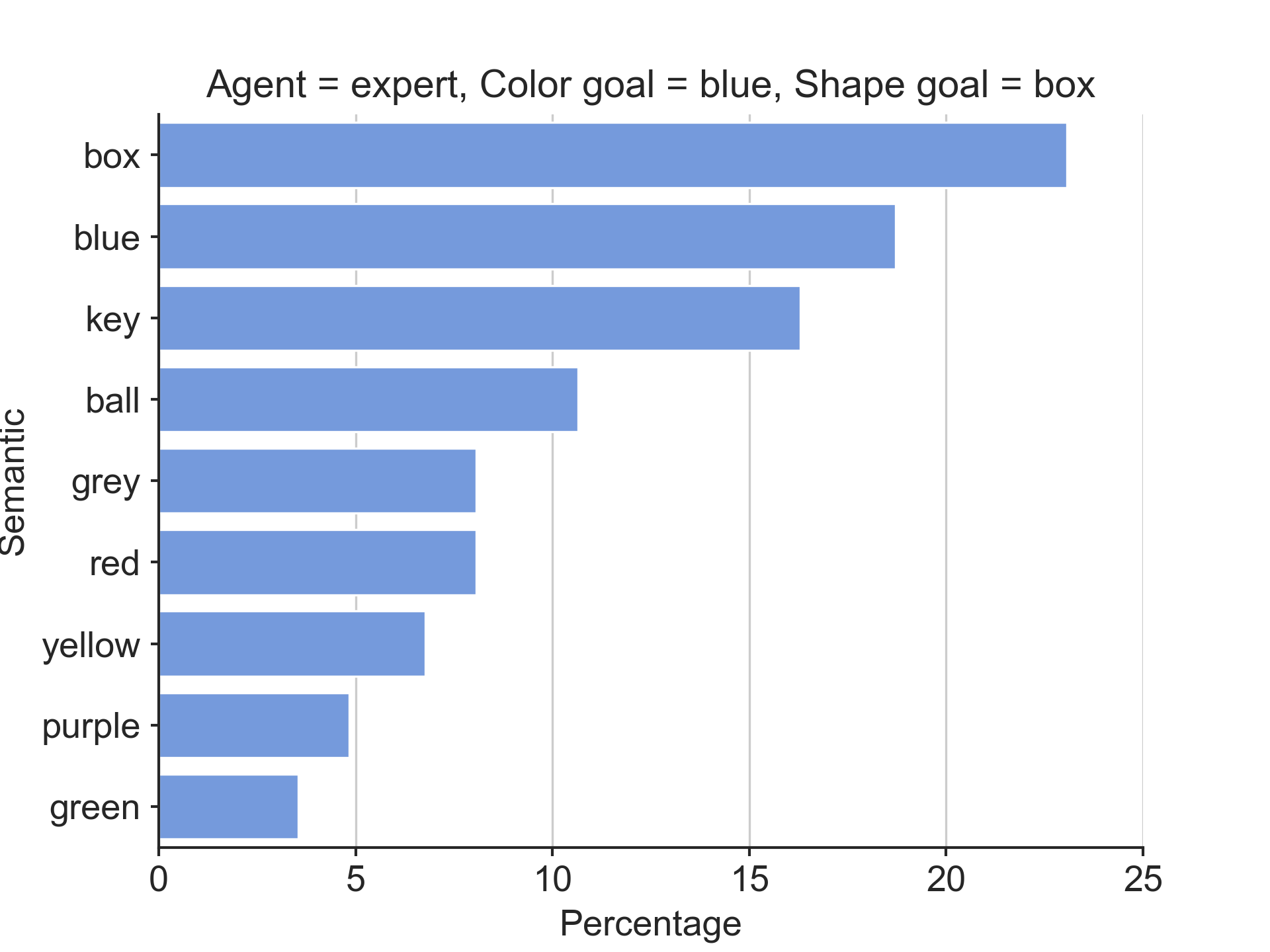}
    %\caption{} % Empty caption preserves layout counter structure
\end{subfigure}\hfill
\begin{subfigure}{0.4\linewidth}
    \centering
    \includegraphics[width=\linewidth]{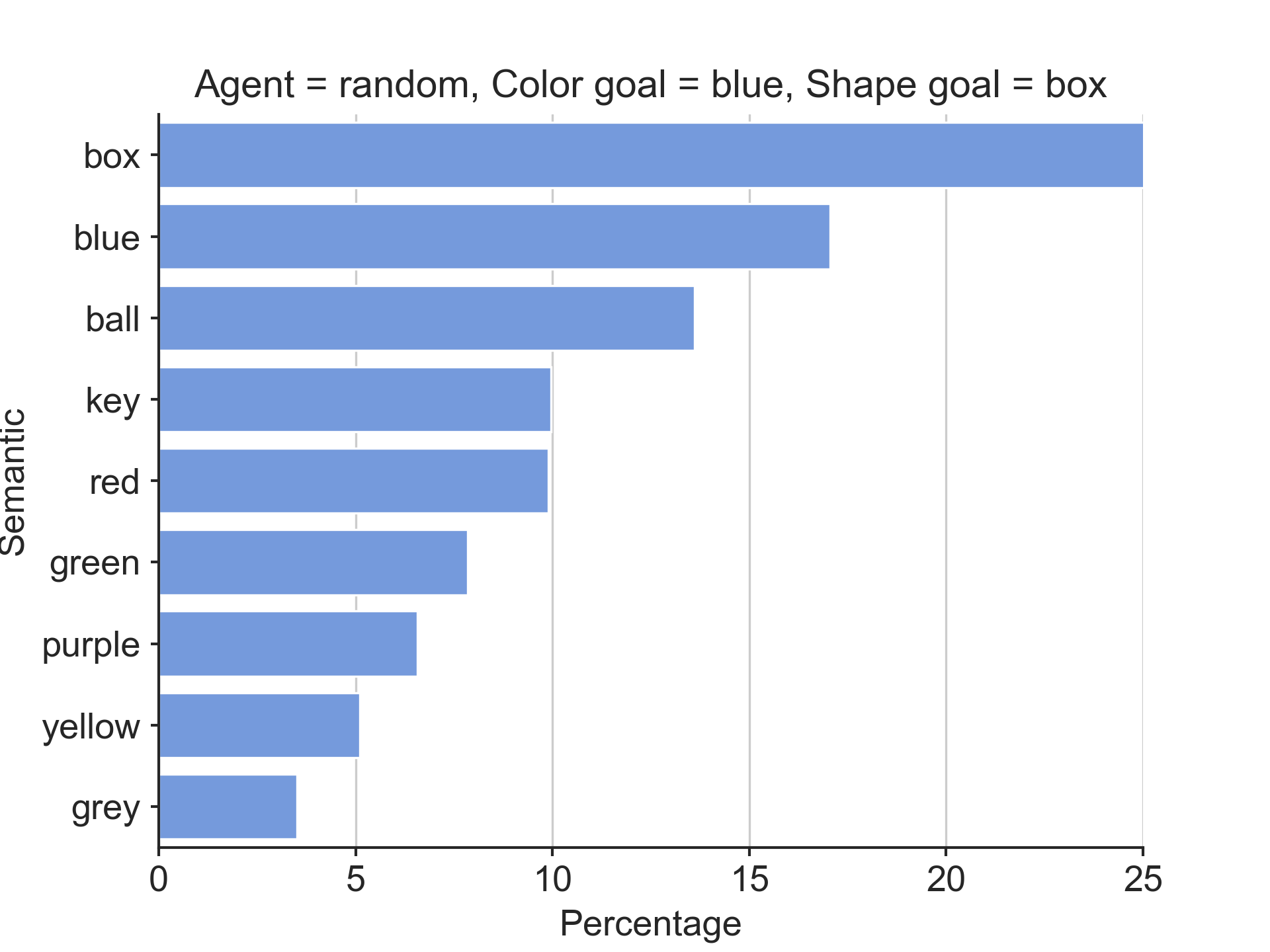}
    %\caption{} % Empty caption preserves layout counter structure
\end{subfigure}
\begin{subfigure}{0.4\linewidth}
    \centering
    \includegraphics[width=\linewidth]{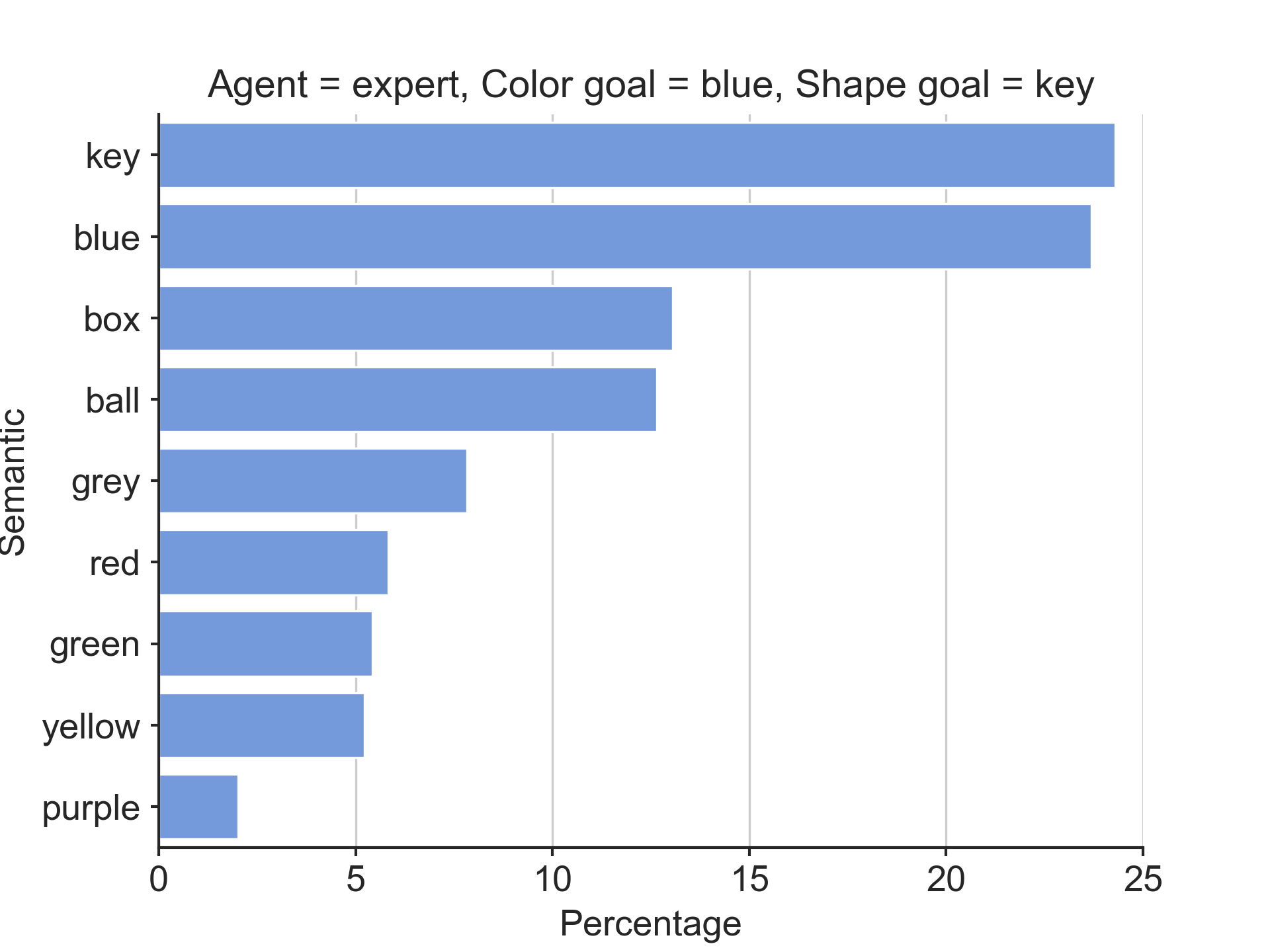}
    %\caption{} % Empty caption preserves layout counter structure
\end{subfigure}\hfill
\begin{subfigure}{0.4\linewidth}
    \centering
    \includegraphics[width=\linewidth]{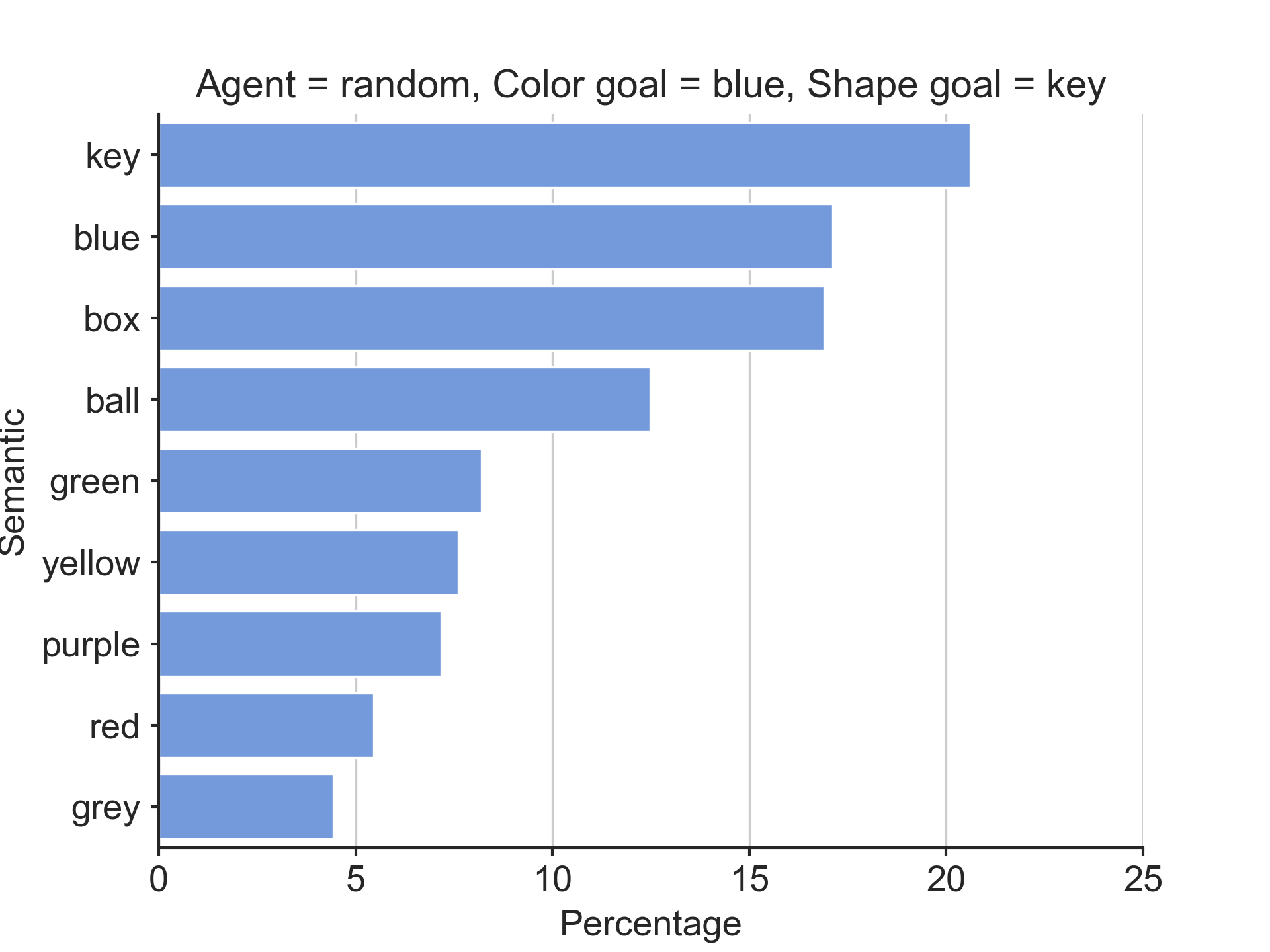}
    %\caption{} % Empty caption preserves layout counter structure
\end{subfigure}
\begin{subfigure}{0.4\linewidth}
    \centering
    \includegraphics[width=\linewidth]{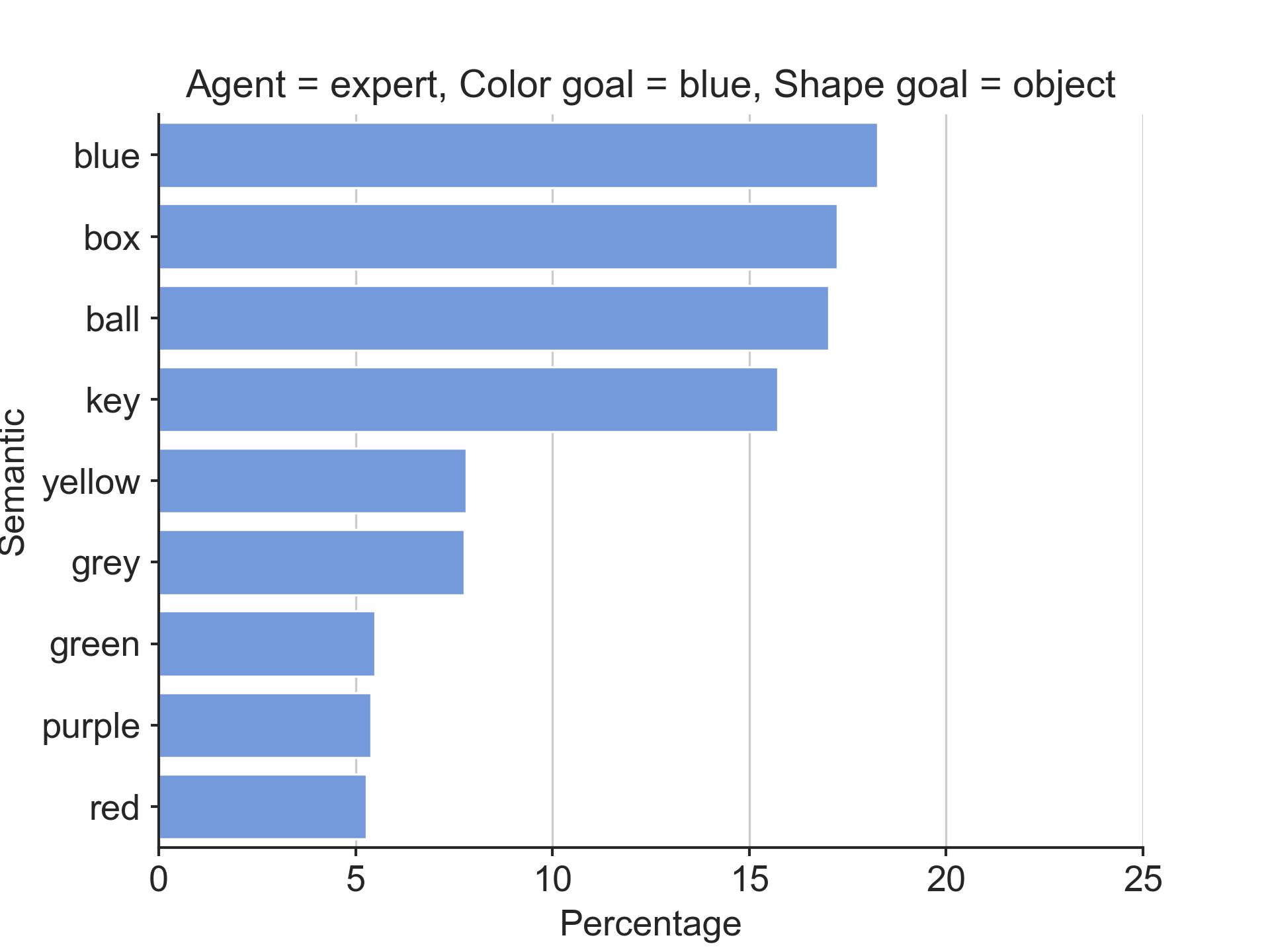}
    \caption{Expert agent}
\end{subfigure}\hfill
\begin{subfigure}{0.4\linewidth}
    \centering
    \includegraphics[width=\linewidth]{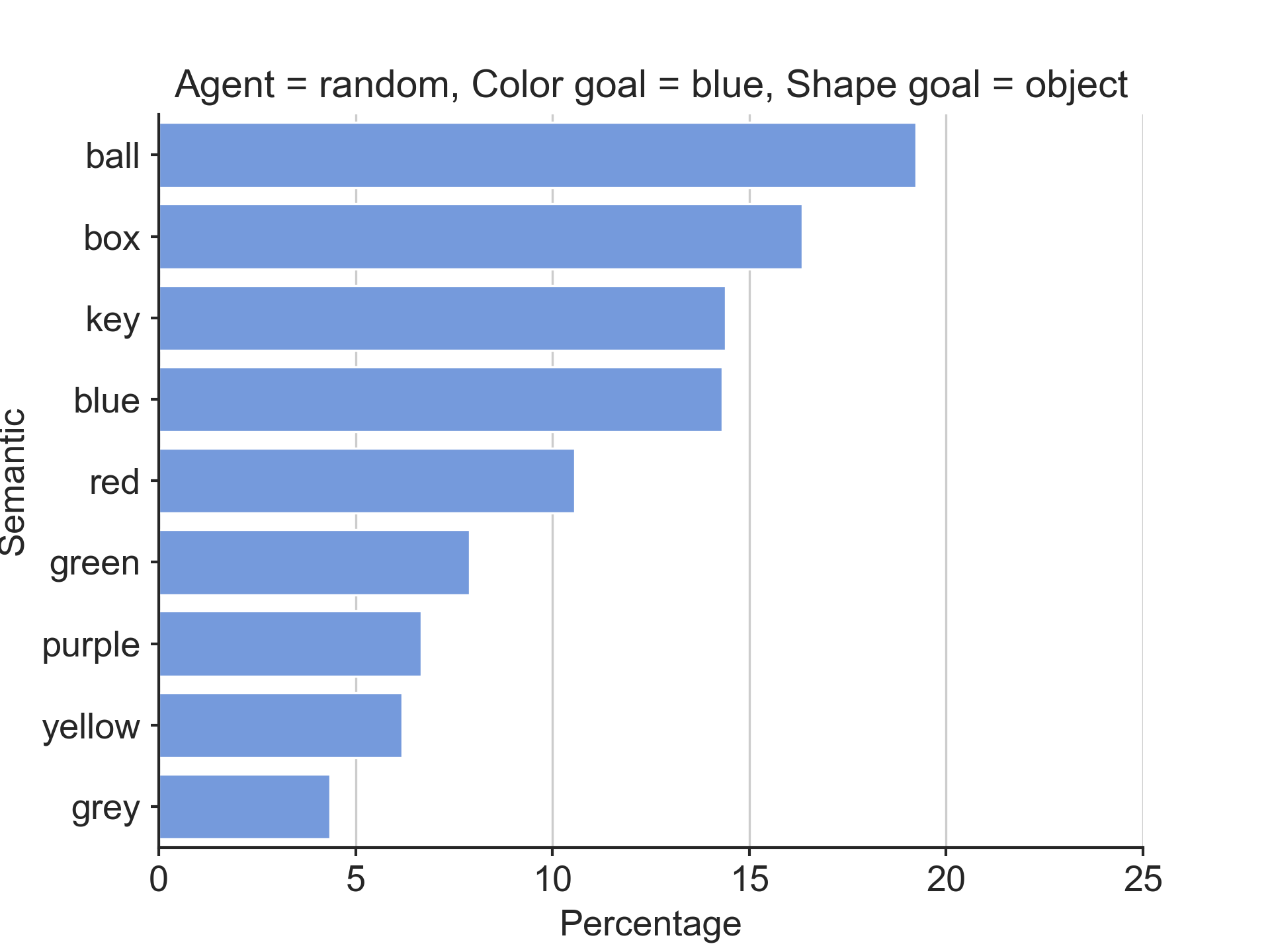}
    \caption{Random agent}
\end{subfigure}
\caption{\textbf{Left:} Trajectories for the blue colour goal from BabyAI's built-in expert agent which always reaches the goal. \textbf{Right:} Random agent trajectories. In both cases the semantics of the goal are among the most observed semantic features for any given trajectory. This effect is less pronounced in the random agent.}
\label{fig:appendix_co_occurrence_blue}
\end{figure*}

\begin{figure}[h]
    \centering
    \begin{subfigure}{0.45\textwidth}
        \centering
        \includegraphics[width=\textwidth]{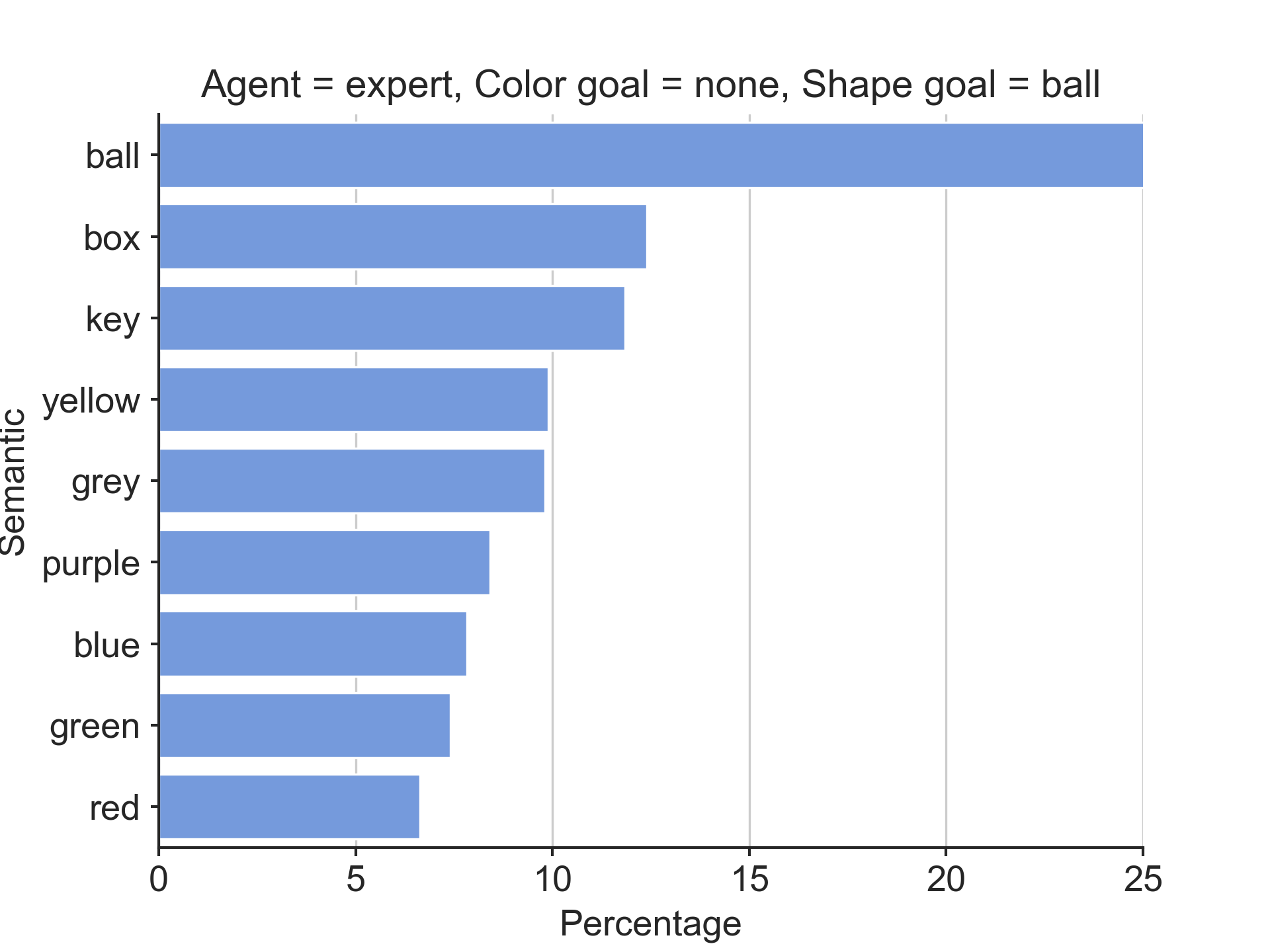}
    \end{subfigure}\hfill
    \begin{subfigure}{0.45\textwidth}
        \centering
        \includegraphics[width=\textwidth]{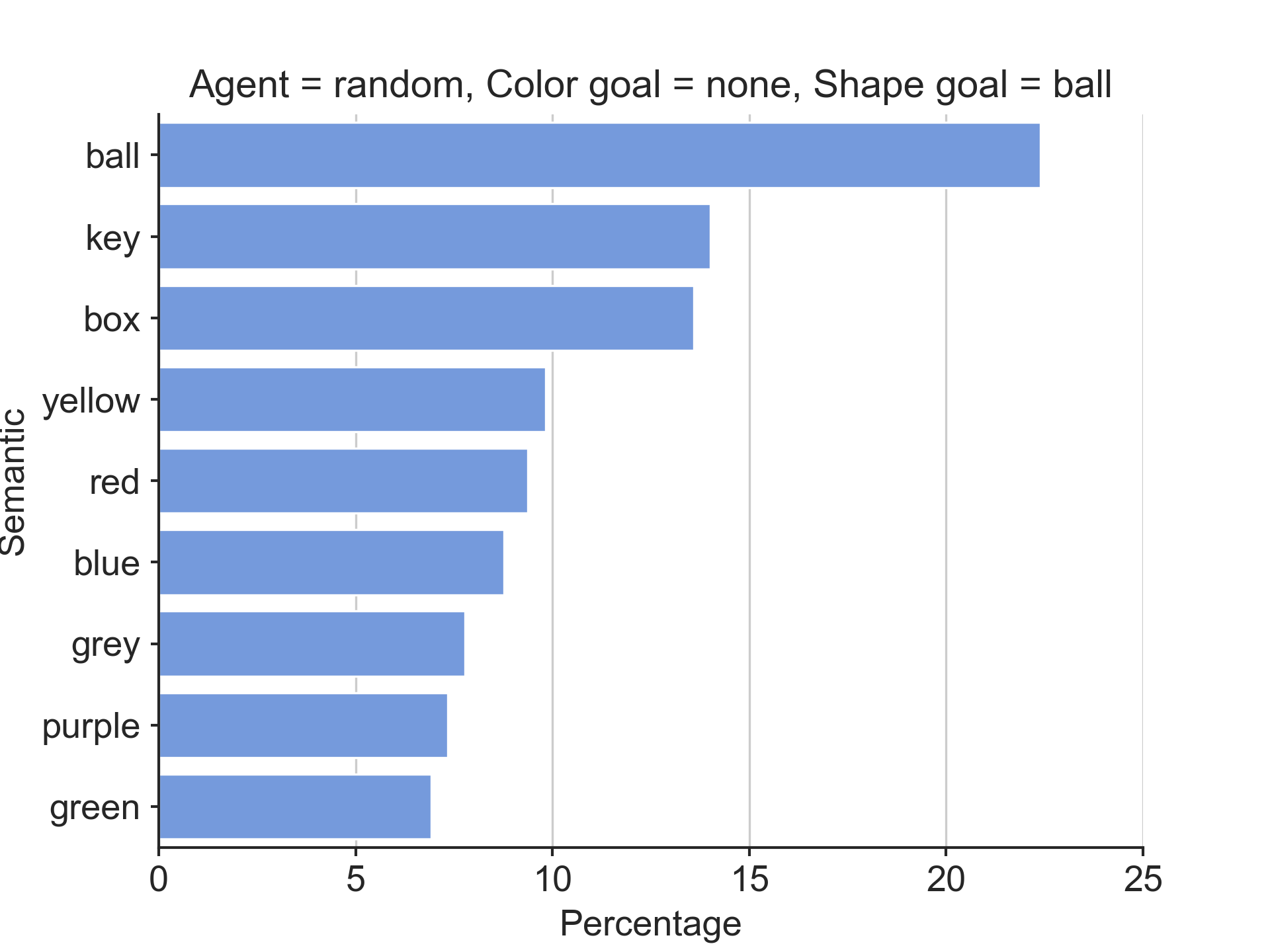}
    \end{subfigure} \\
    
    \begin{subfigure}{0.45\textwidth}
        \centering
        \includegraphics[width=\textwidth]{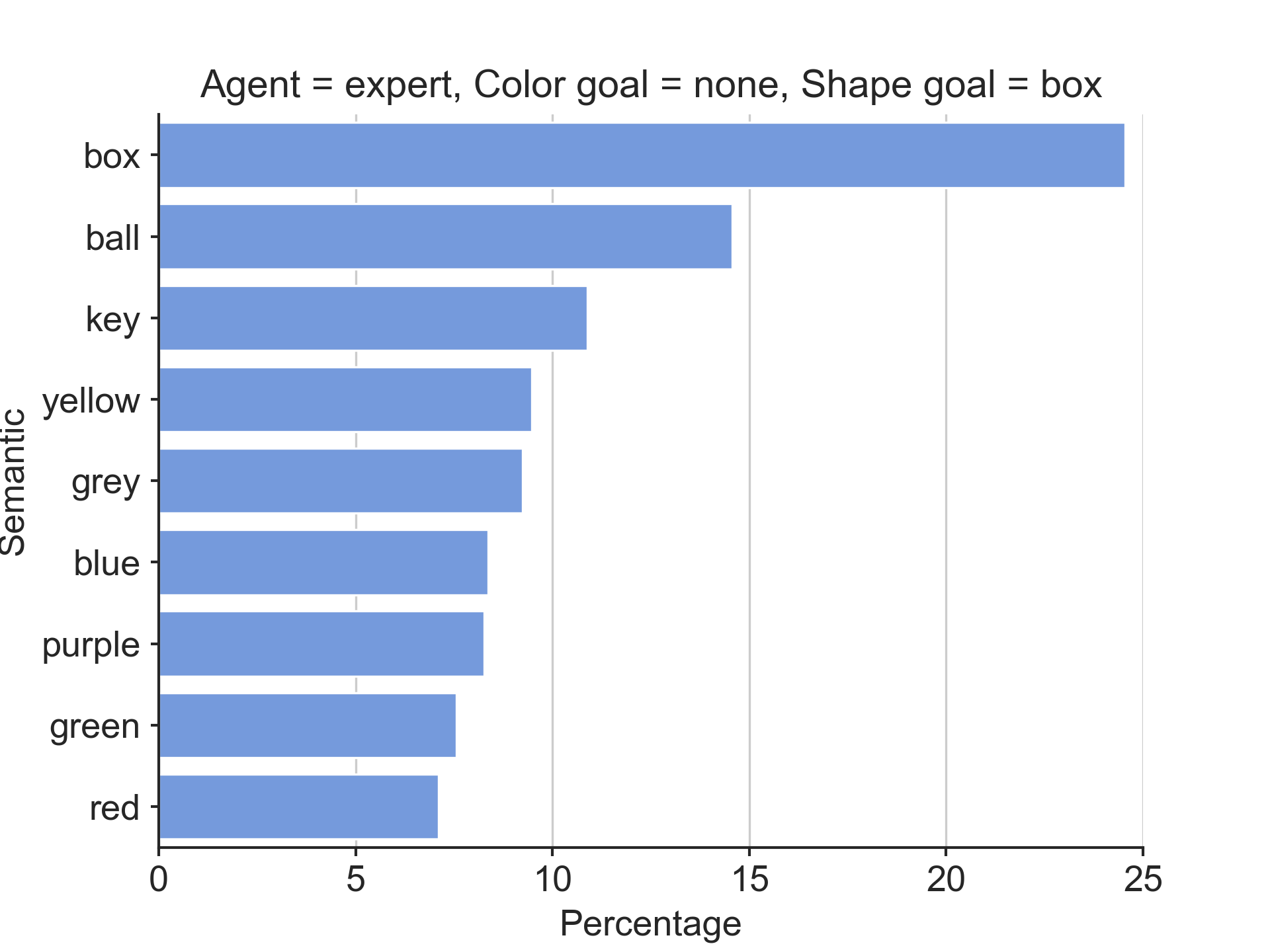}
    \end{subfigure}\hfill
    \begin{subfigure}{0.45\textwidth}
        \centering
        \includegraphics[width=\textwidth]{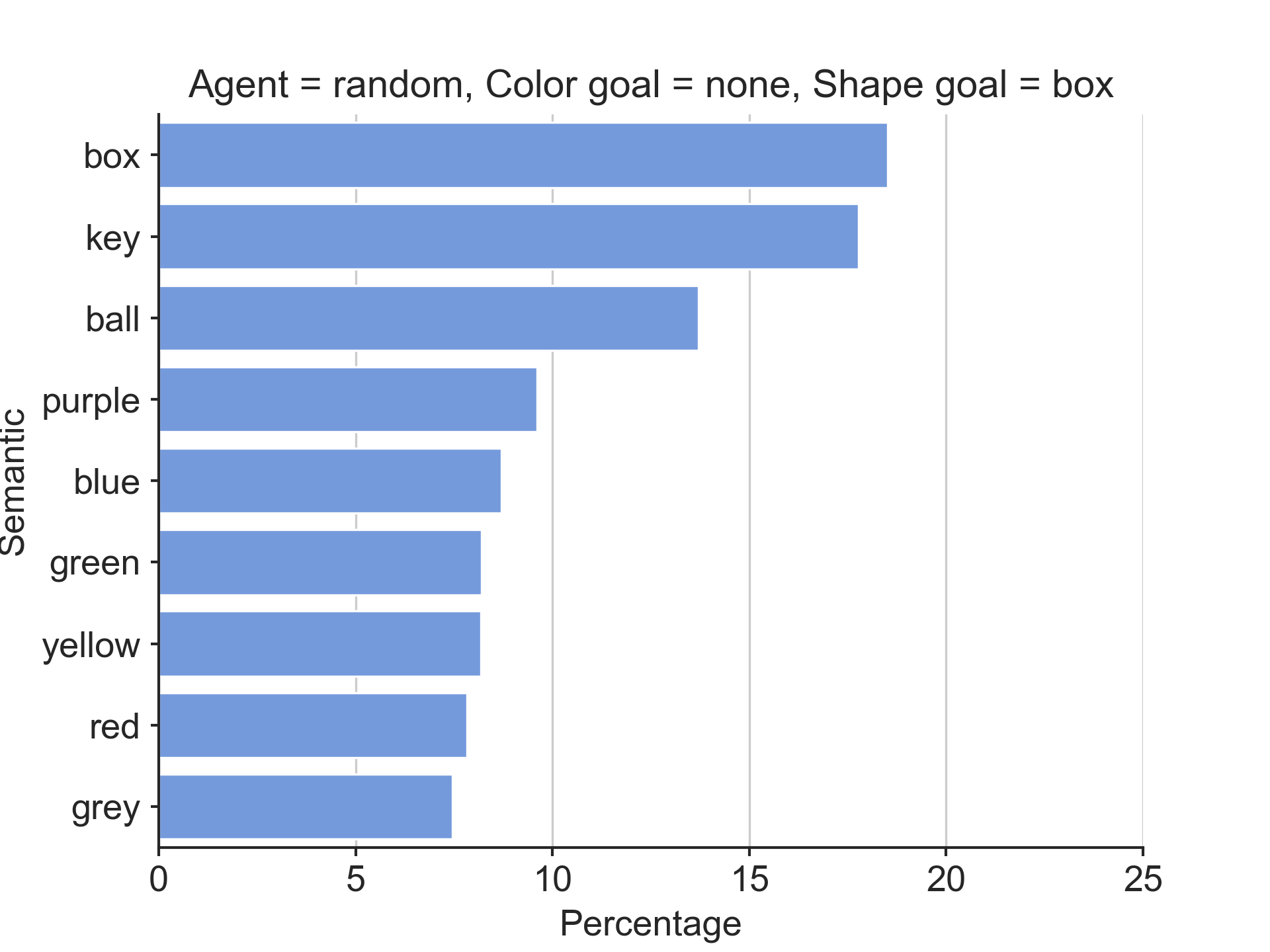}
    \end{subfigure} \\
    
    \begin{subfigure}{0.45\textwidth}
        \centering
        \includegraphics[width=\textwidth]{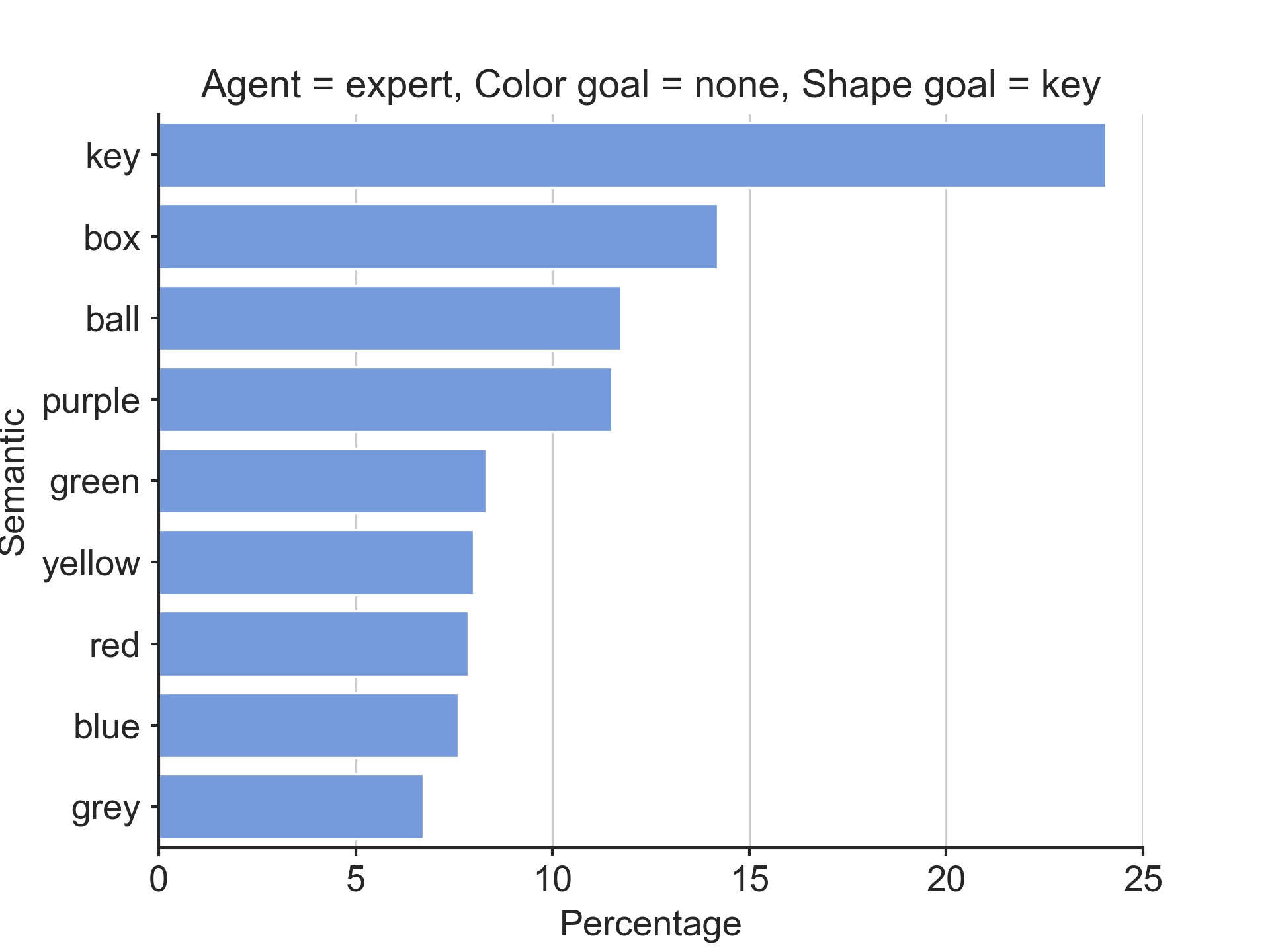}
        \caption{Expert agent}
    \end{subfigure}\hfill
    \begin{subfigure}{0.45\textwidth}
        \centering
        \includegraphics[width=\textwidth]{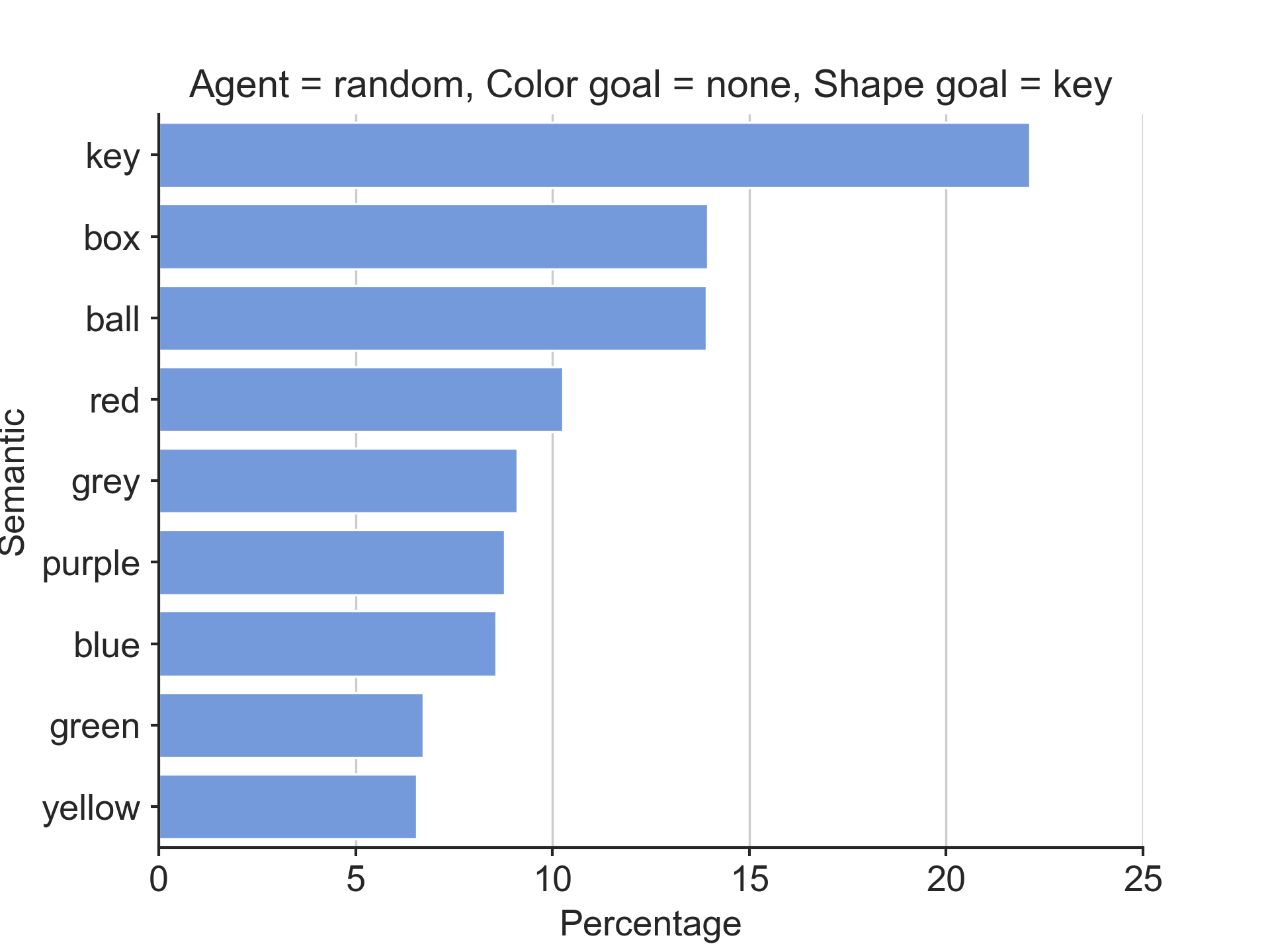}
        \caption{Random agent}
    \end{subfigure}
    
    \caption{\textbf{Left:} Trajectories for colorless objects from BabyAI's built-in expert agent which always reaches the goal. \textbf{Right:} Random agent trajectories. In both cases the semantics of the goal are among the most observed semantic features for any given trajectory. This effect is less pronounced in the random agent.}
    \label{fig:appendix_co_occurrence_none}
\end{figure}

\begin{figure}
    \centering
    \begin{subfigure}{0.45\linewidth}
        \centering
        \includegraphics[width=\textwidth]{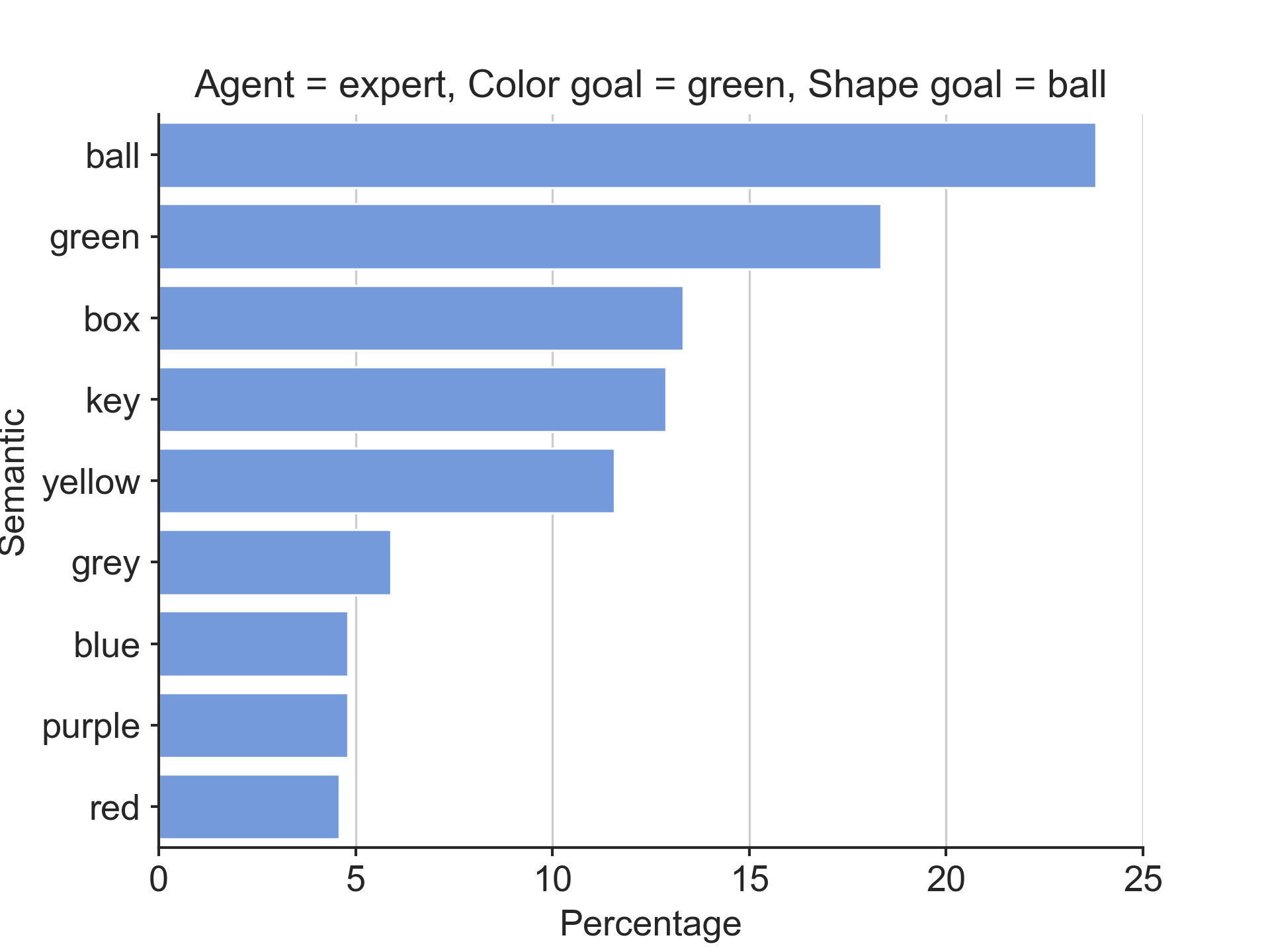}
    \end{subfigure}
    \begin{subfigure}{0.45\linewidth}
        \centering
        \includegraphics[width=\textwidth]{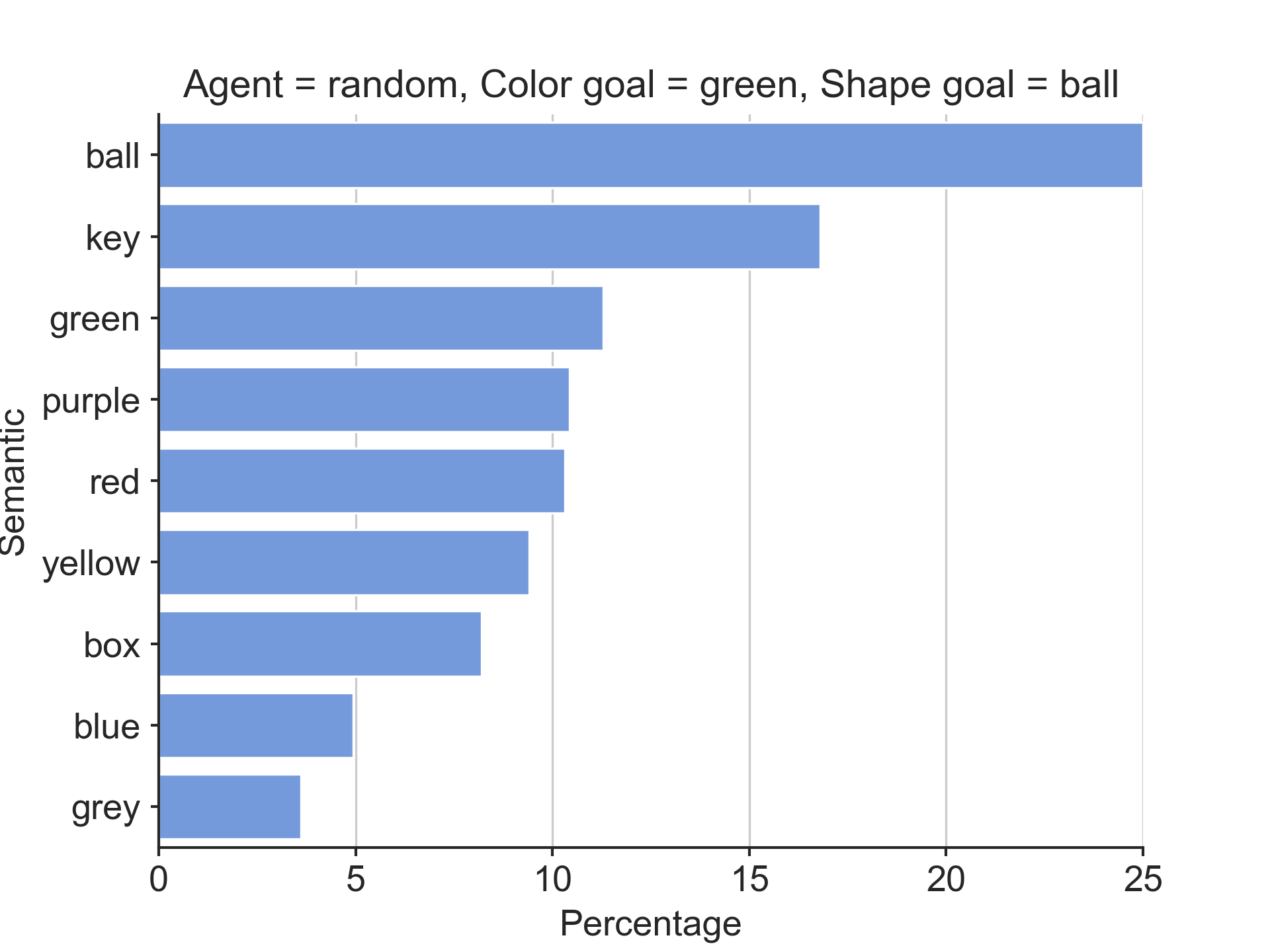}
    \end{subfigure}
    \begin{subfigure}{0.45\linewidth}
        \centering
        \includegraphics[width=\textwidth]{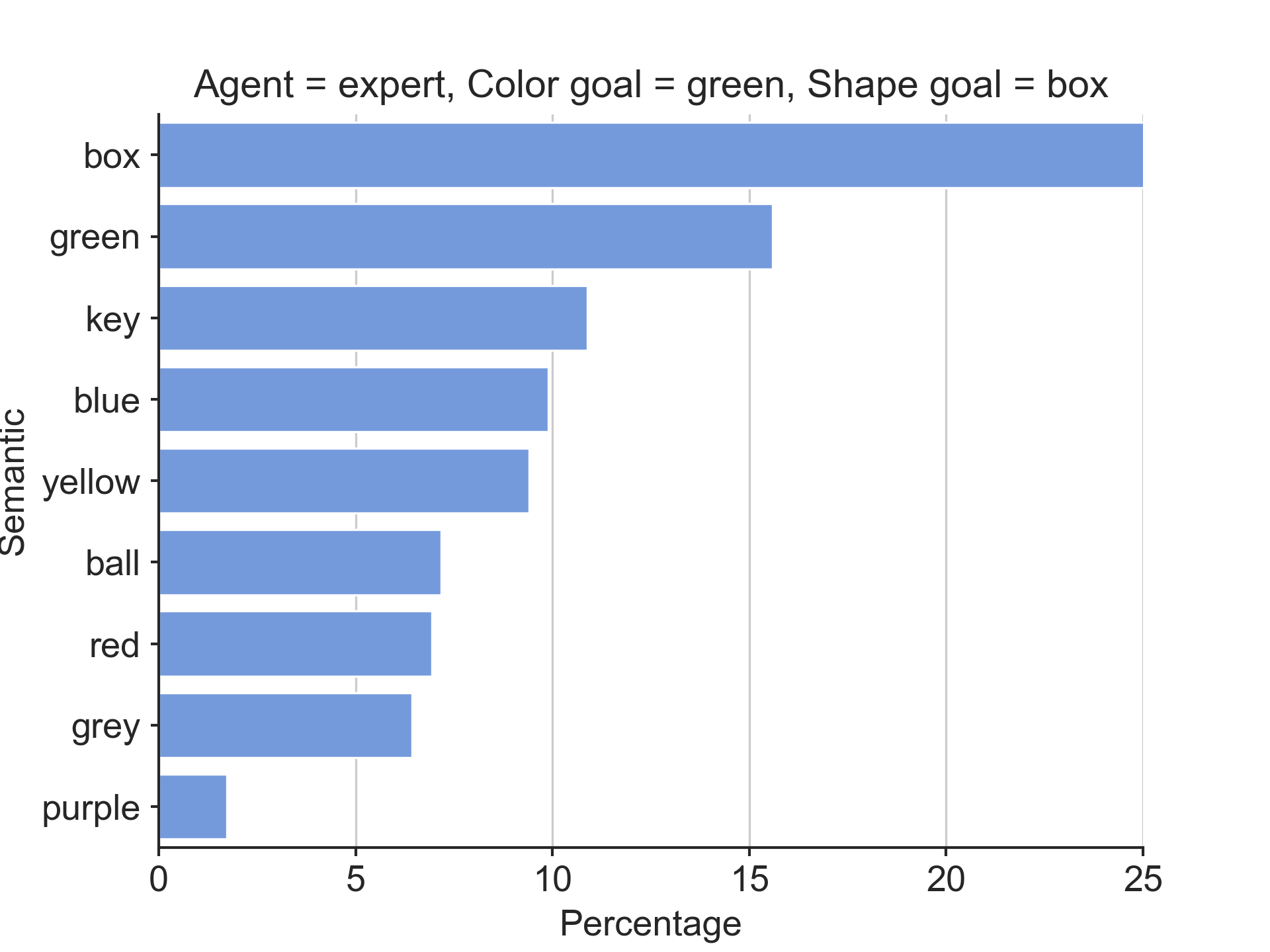}
    \end{subfigure}
    \begin{subfigure}{0.45\linewidth}
        \centering
        \includegraphics[width=\textwidth]{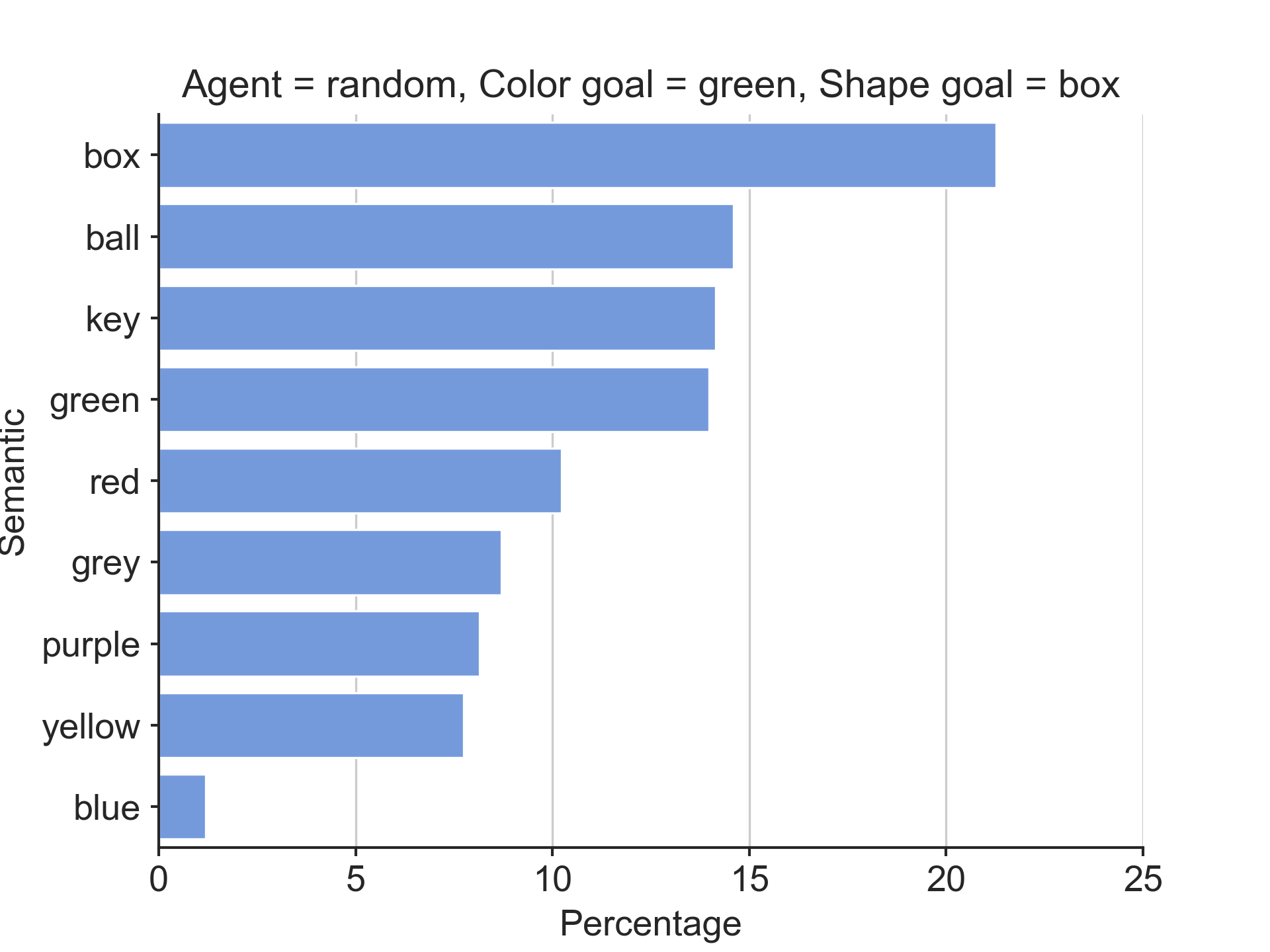}
    \end{subfigure}
    \begin{subfigure}{0.45\linewidth}
        \centering
        \includegraphics[width=\textwidth]{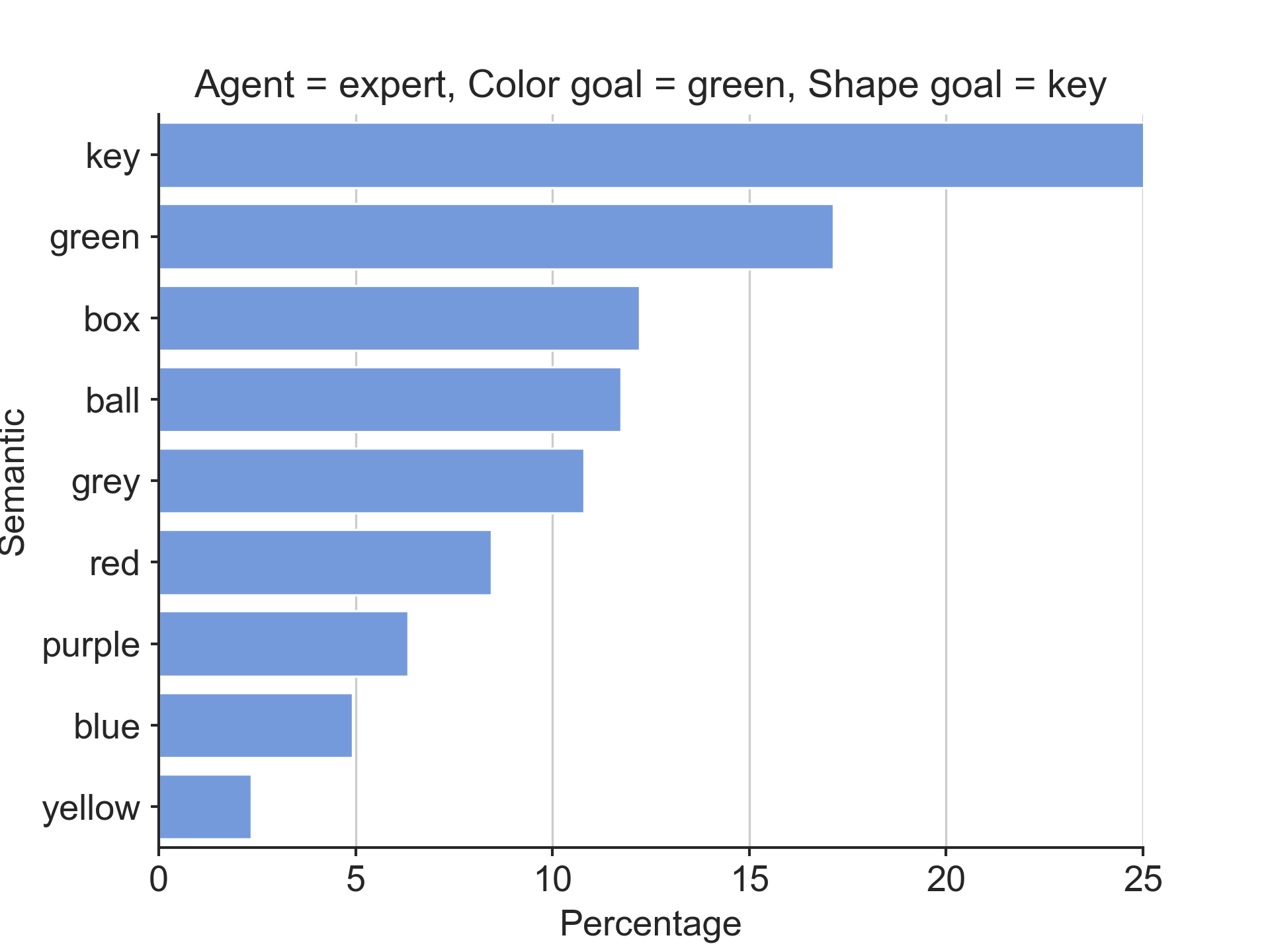}
    \end{subfigure}
    \begin{subfigure}{0.45\linewidth}
        \centering
        \includegraphics[width=\textwidth]{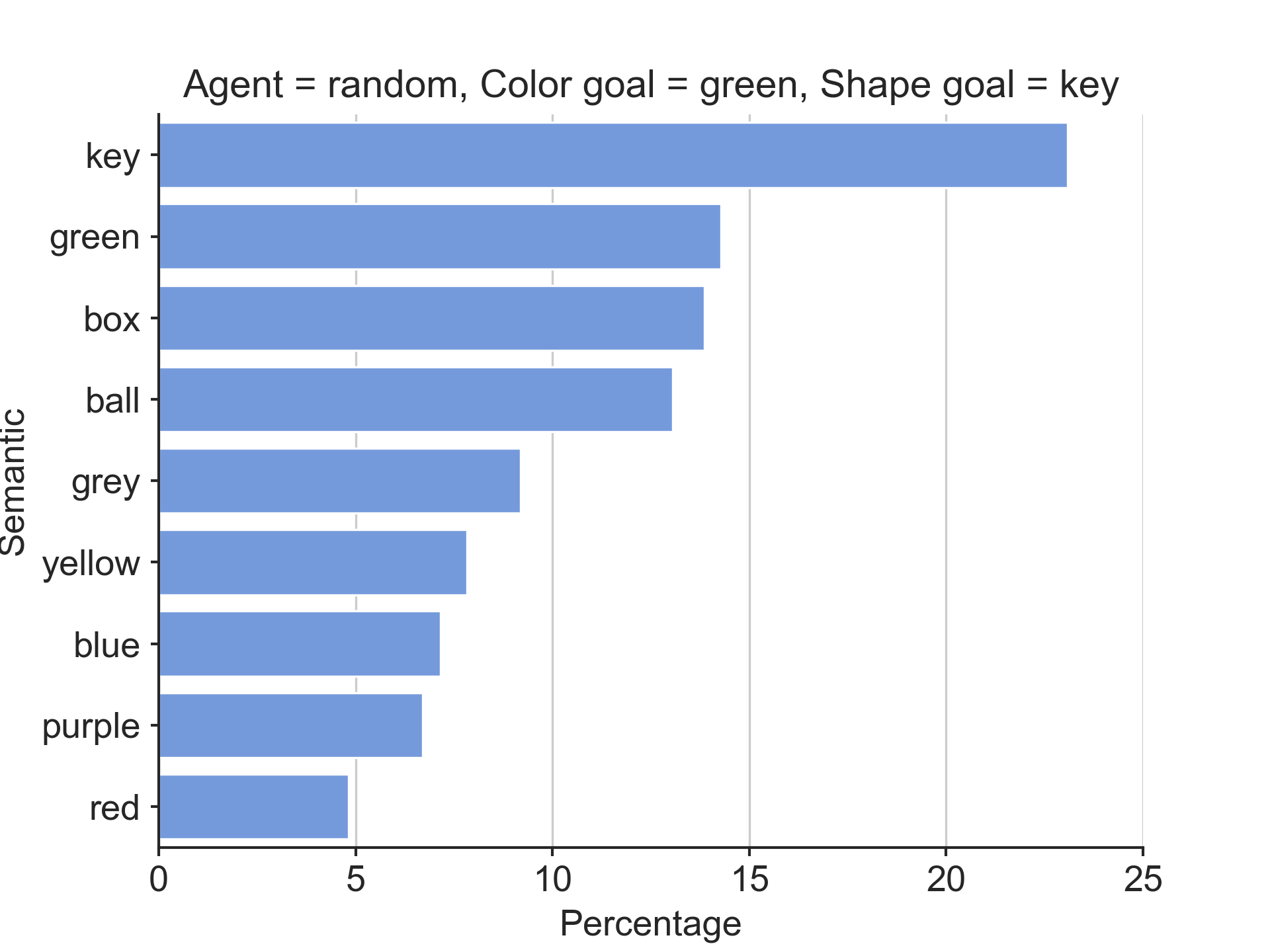}
    \end{subfigure}
    \begin{subfigure}{0.45\linewidth}
        \centering
        \includegraphics[width=\textwidth]{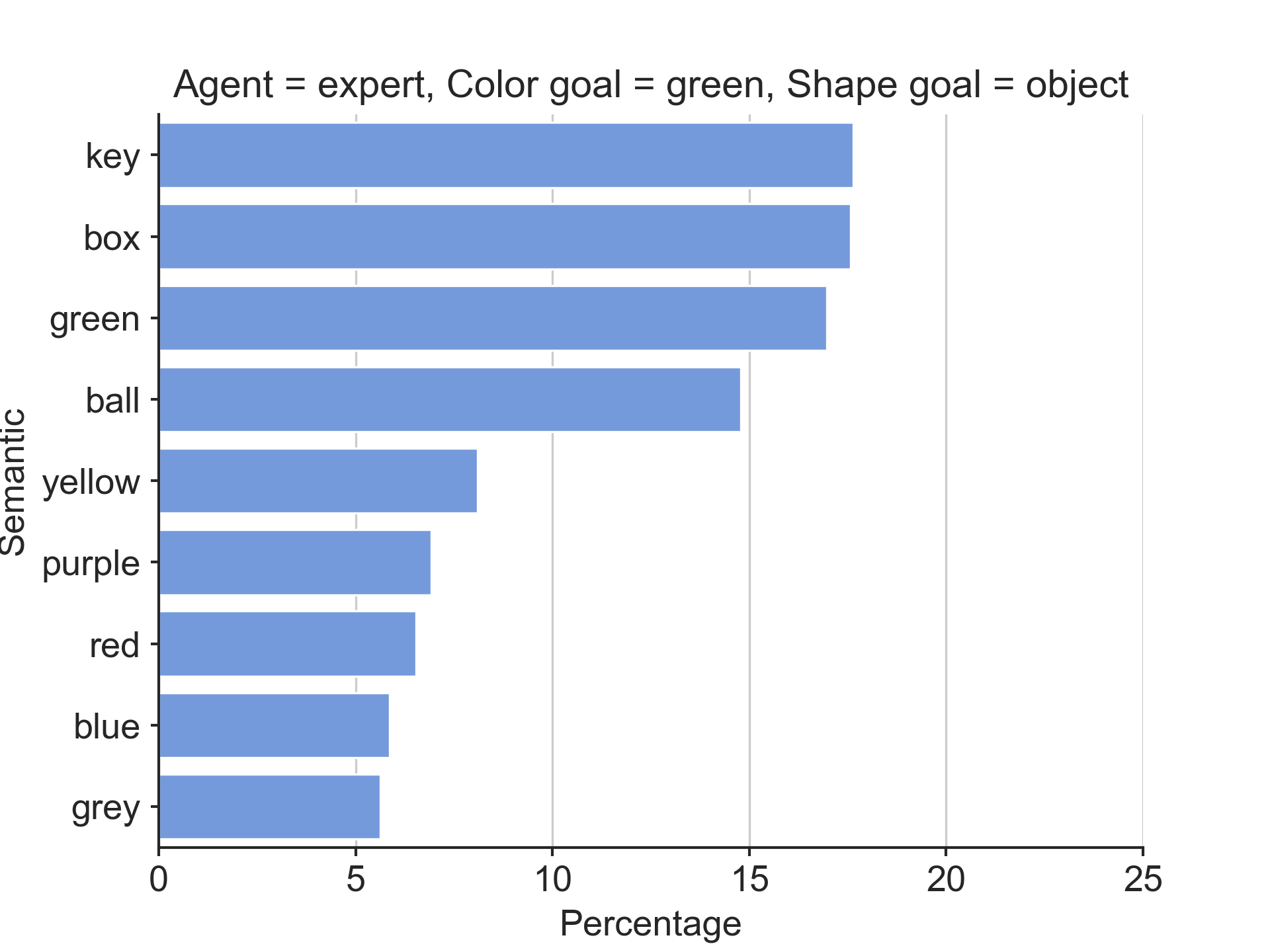}
        \caption{Expert agent}
    \end{subfigure}
    \begin{subfigure}{0.45\linewidth}
        \centering
        \includegraphics[width=\textwidth]{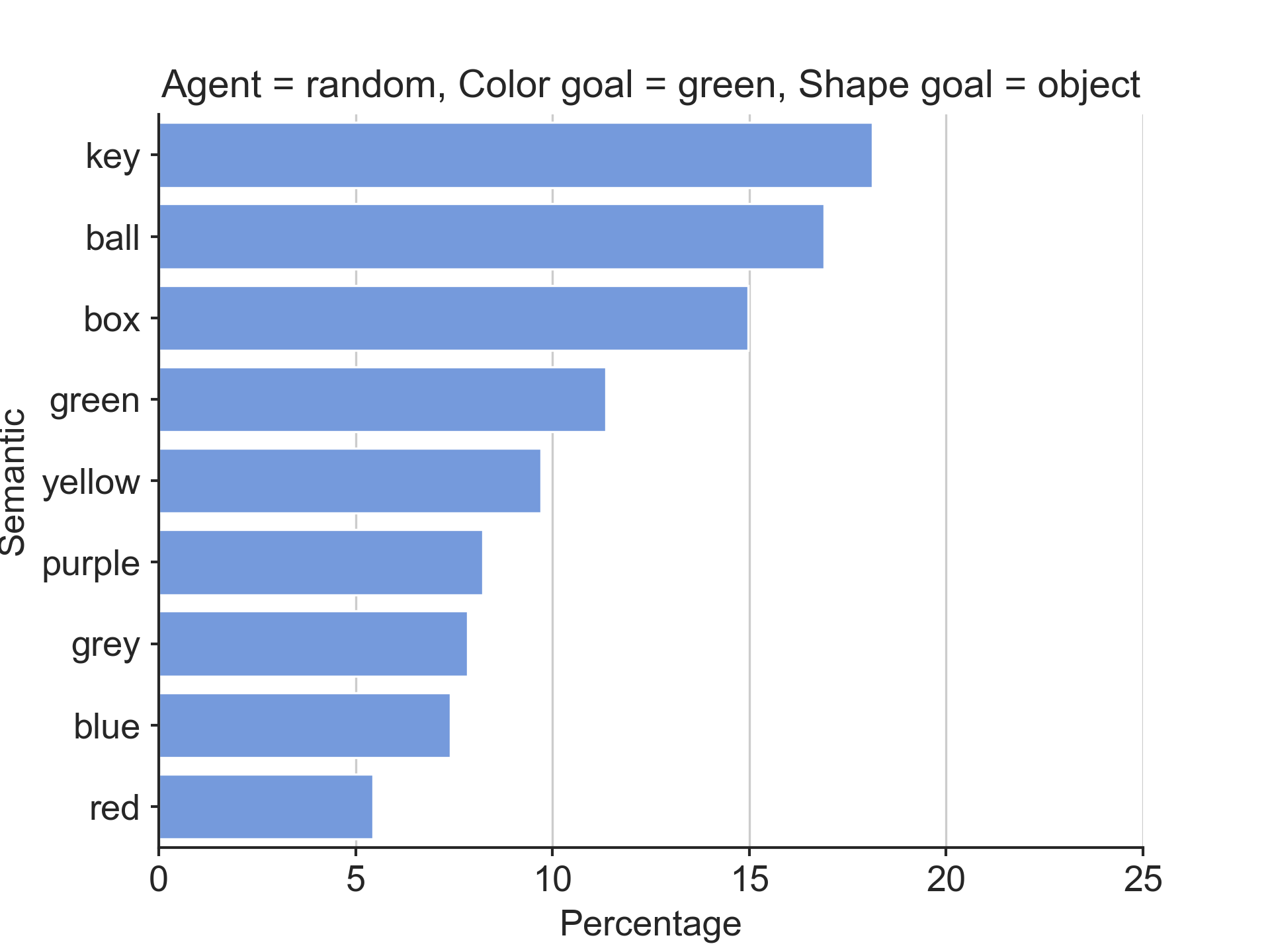}
        \caption{Random agent}
    \end{subfigure}
    \caption{\textbf{Left:} Trajectories for the green color goal from BabyAI's built-in expert agent which always reaches the goal. \textbf{Right:} Random agent trajectories. In both cases the semantics of the goal are among the most observed semantic features for any given trajectory. This effect is less pronounced in the random agent.}
    \label{fig:appendix_co_occurrence_green}
\end{figure}

\begin{figure}
    \centering
    \begin{subfigure}{0.45\linewidth}
        \centering
        \includegraphics[width=\textwidth]{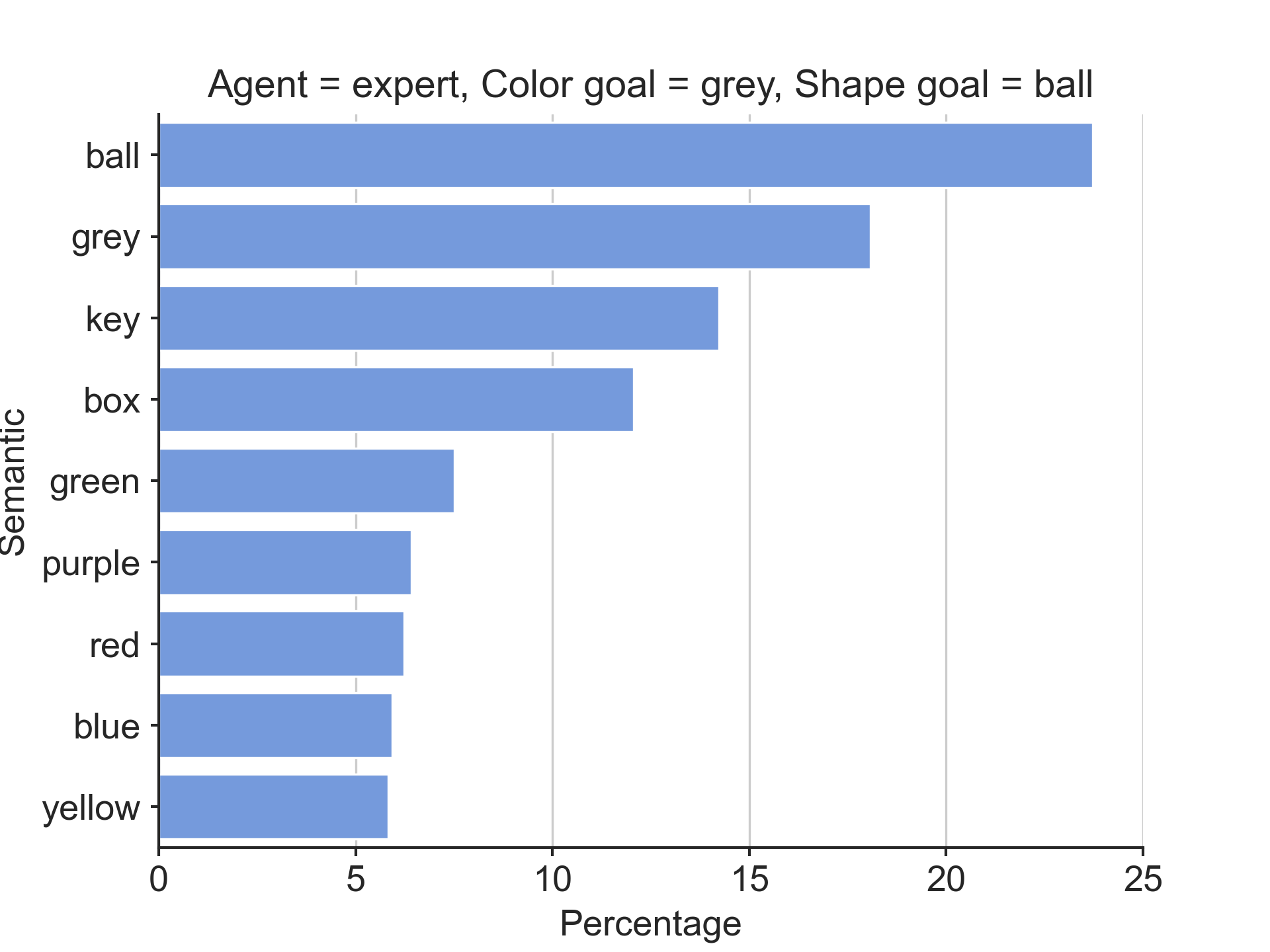}
    \end{subfigure}
    \begin{subfigure}{0.45\linewidth}
        \centering
        \includegraphics[width=\textwidth]{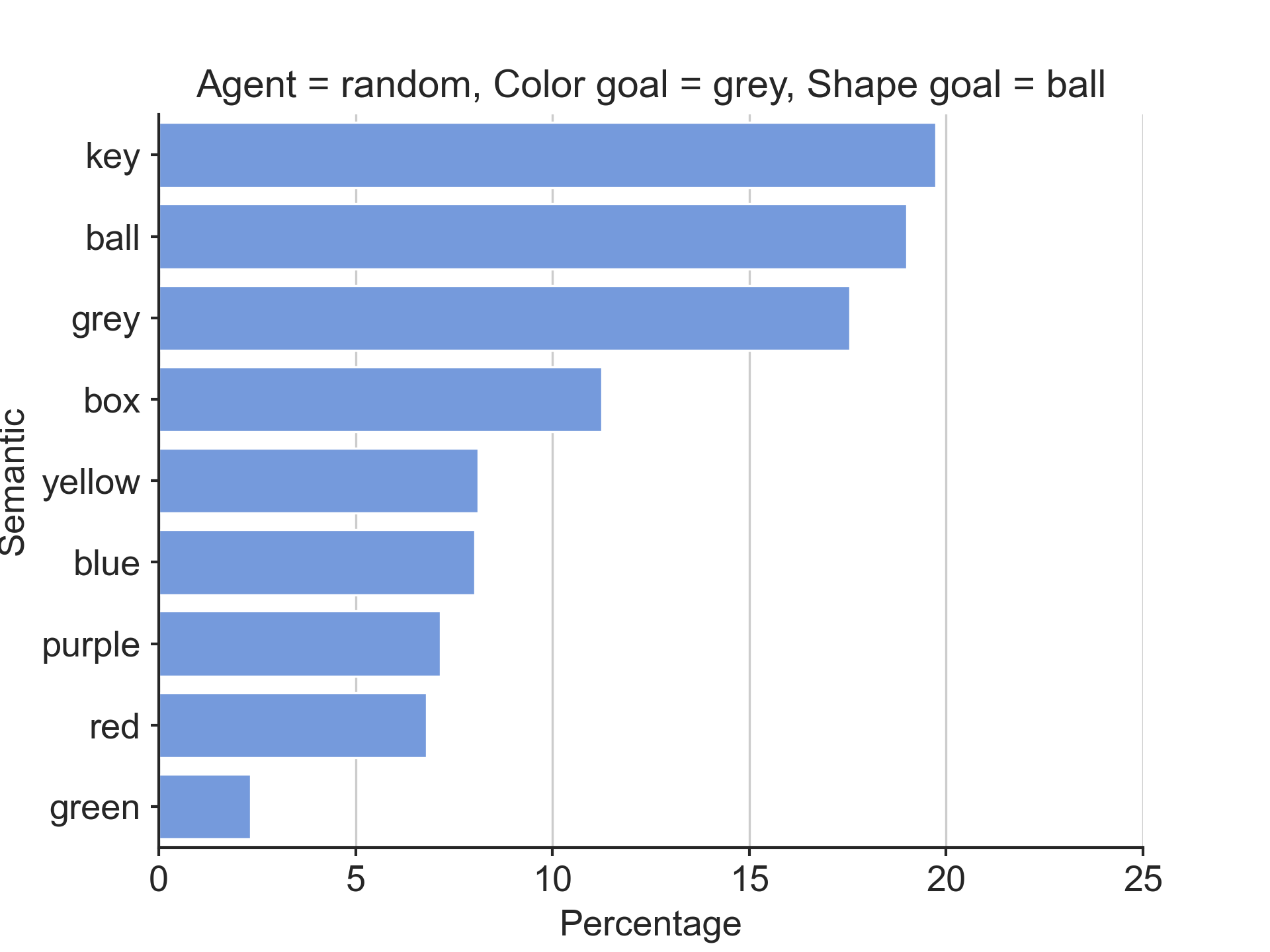}
    \end{subfigure}
    \begin{subfigure}{0.45\linewidth}
        \centering
        \includegraphics[width=\textwidth]{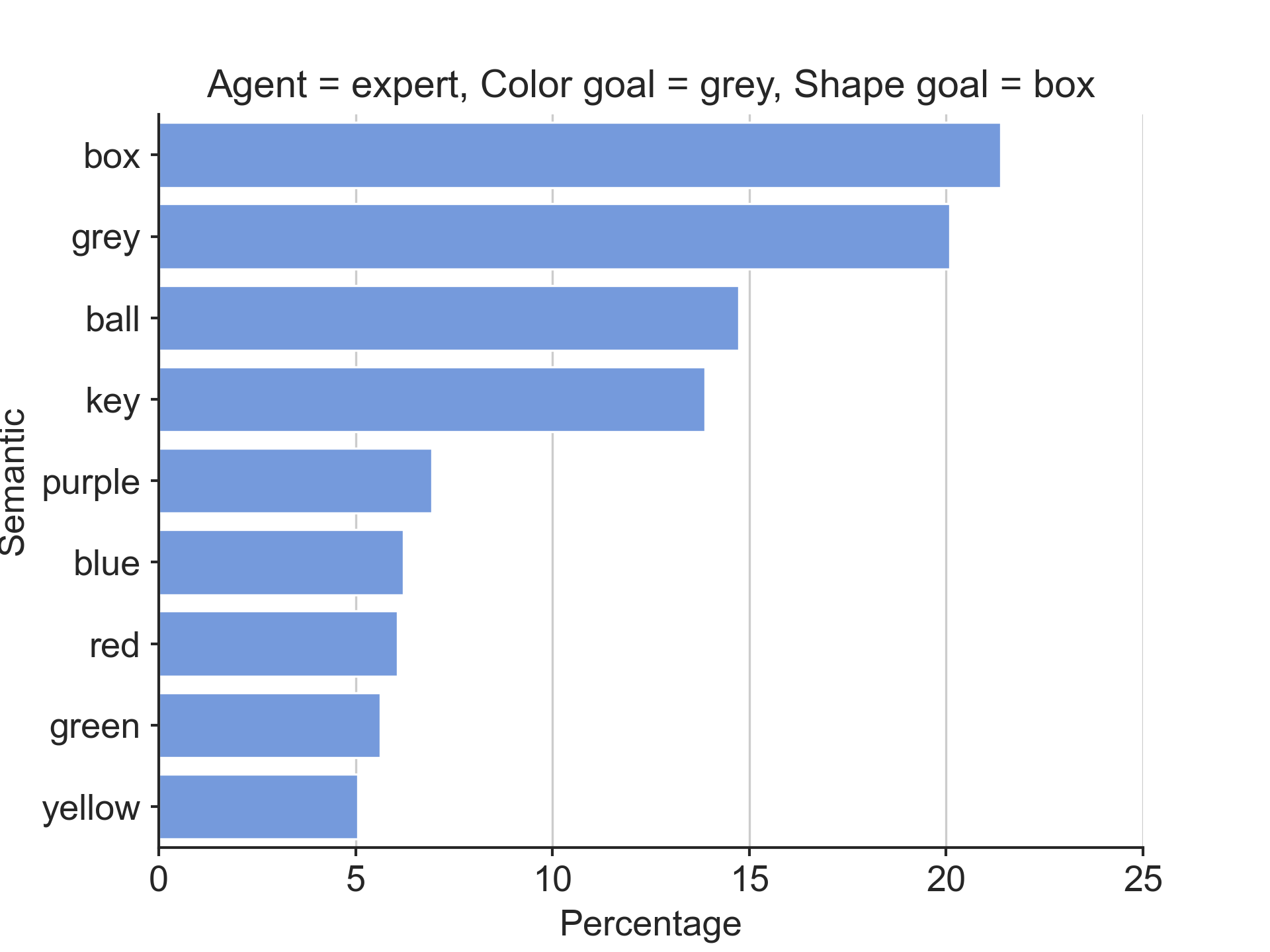}
    \end{subfigure}
    \begin{subfigure}{0.45\linewidth}
        \centering
        \includegraphics[width=\textwidth]{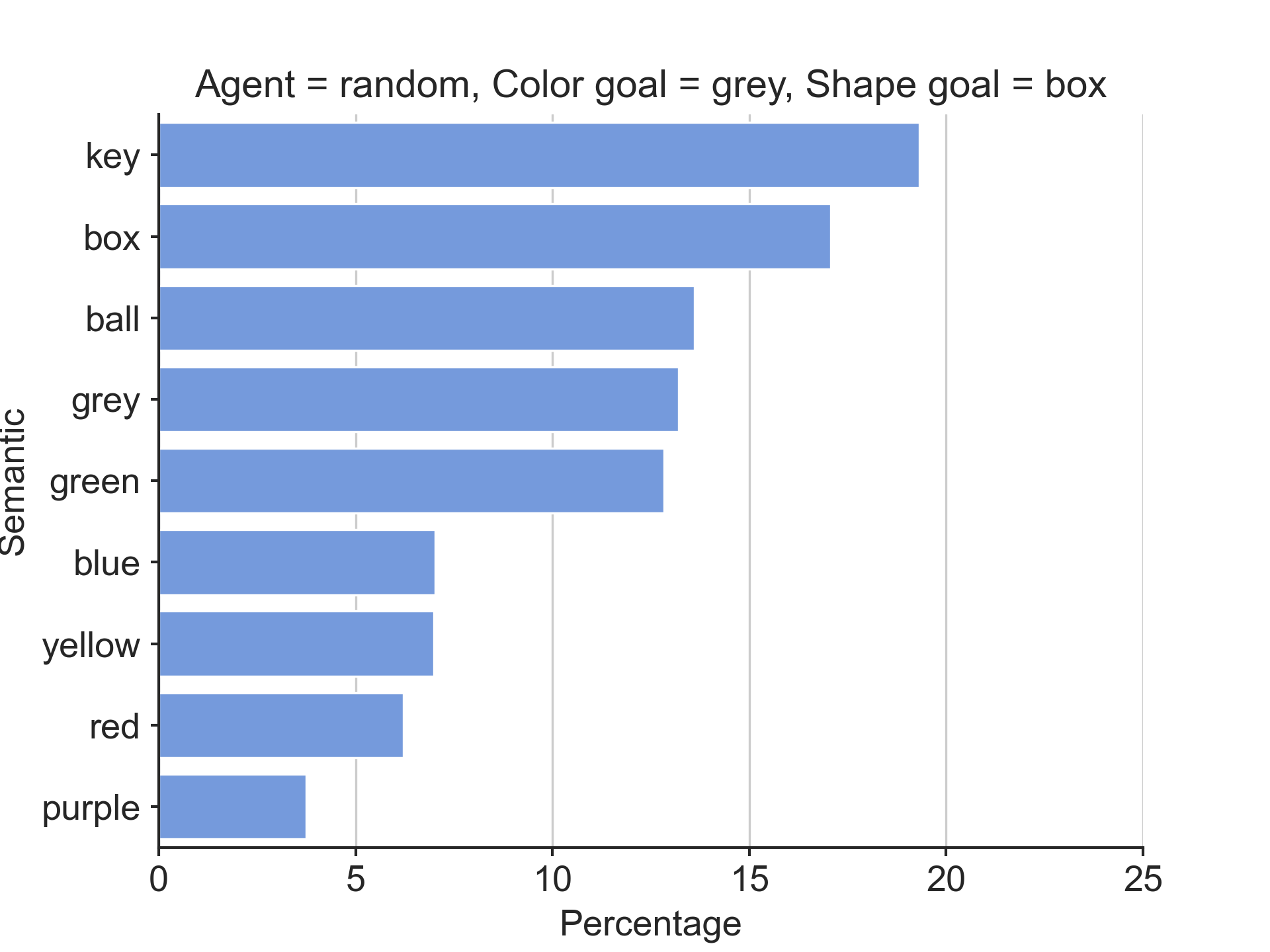}
    \end{subfigure}
    \begin{subfigure}{0.45\linewidth}
        \centering
        \includegraphics[width=\textwidth]{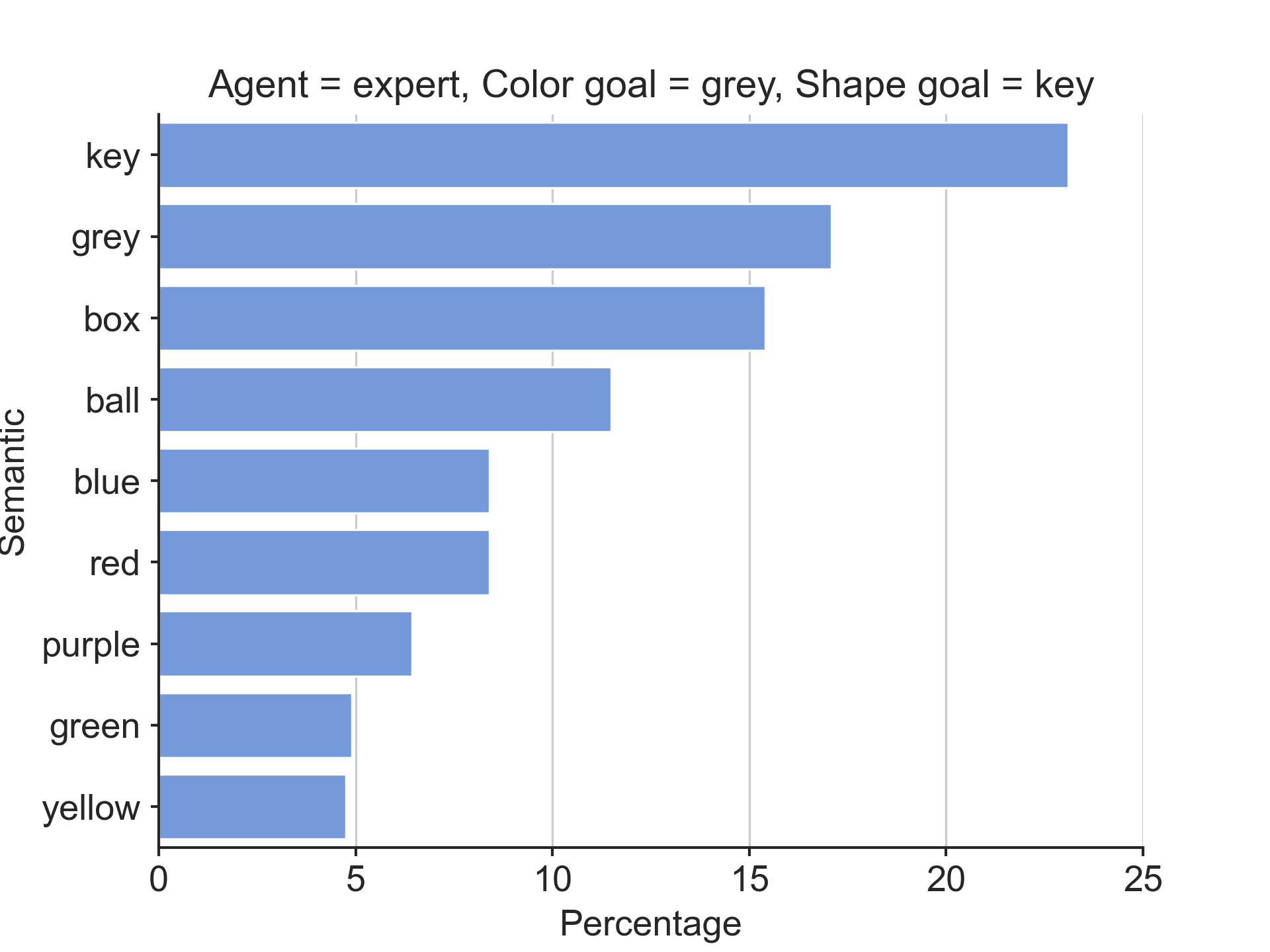}
    \end{subfigure}
    \begin{subfigure}{0.45\linewidth}
        \centering
        \includegraphics[width=\textwidth]{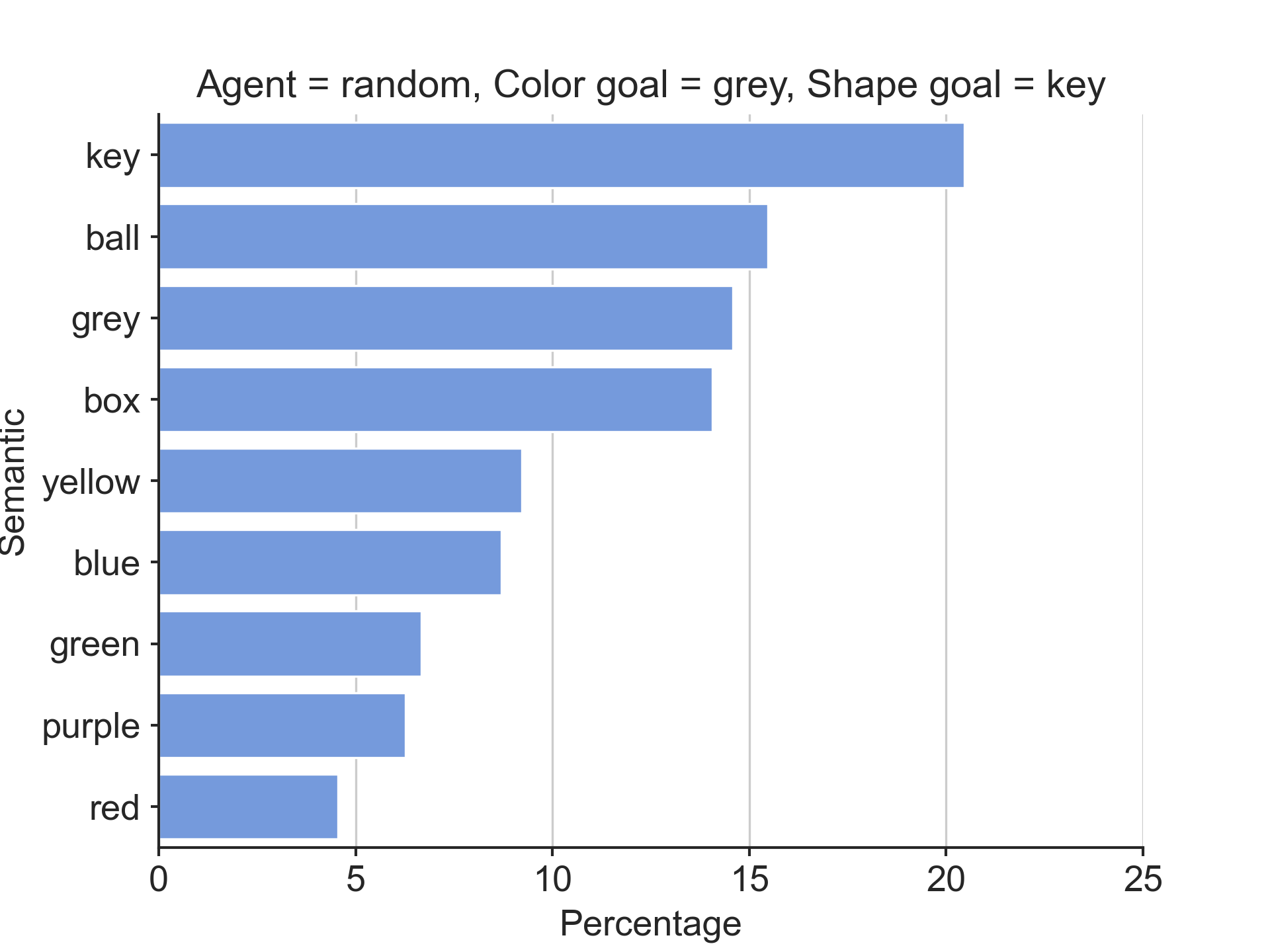}
    \end{subfigure}
    \begin{subfigure}{0.45\linewidth}
        \centering
        \includegraphics[width=\textwidth]{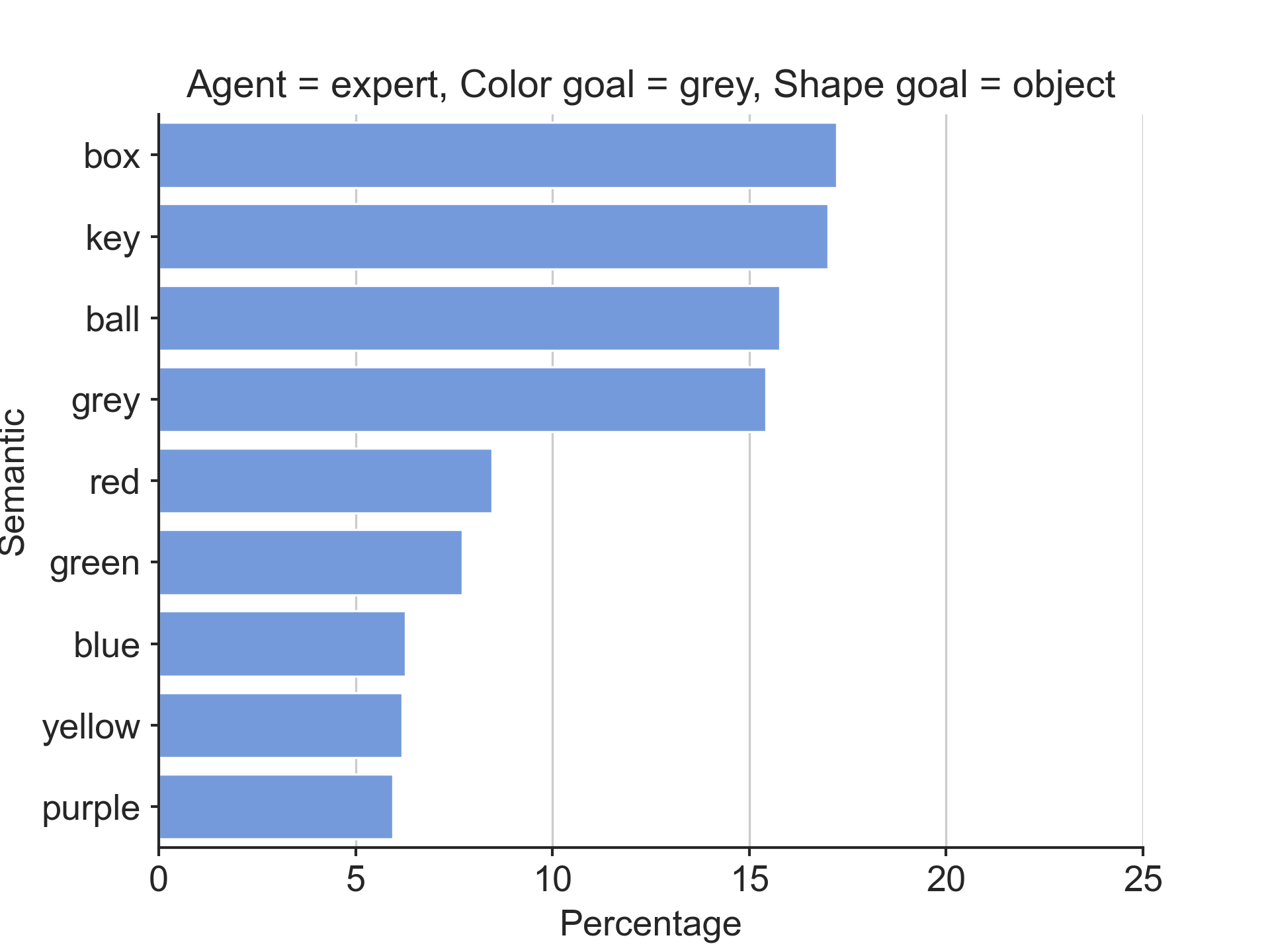}
        \caption{Expert agent}
    \end{subfigure}
    \begin{subfigure}{0.45\linewidth}
        \centering
        \includegraphics[width=\textwidth]{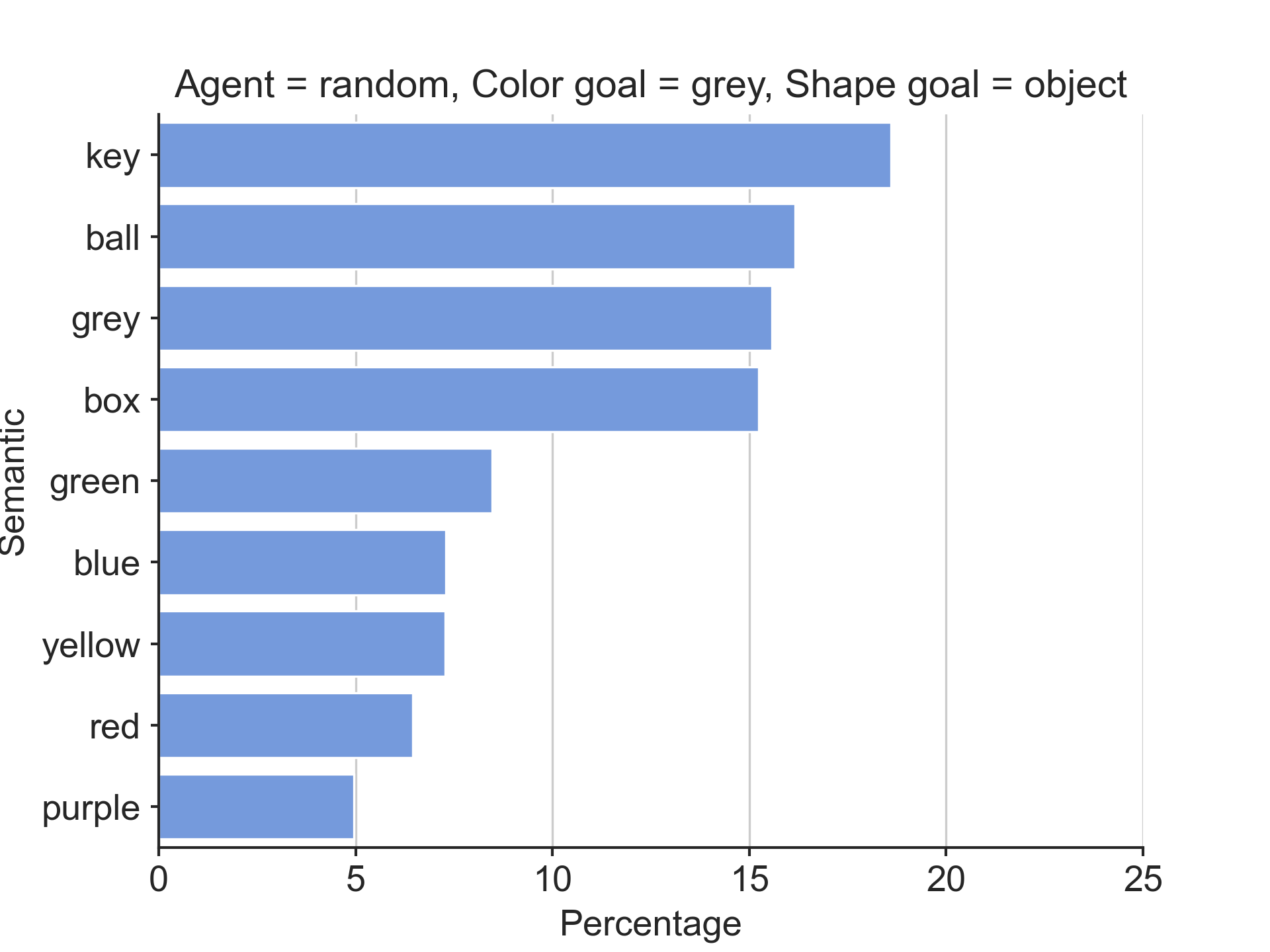}
        \caption{Random agent}
    \end{subfigure}
    \caption{\textbf{Left:} Trajectories for the grey color goal from BabyAI's built-in expert agent which always reaches the goal. \textbf{Right:} Random agent trajectories. In both cases the semantics of the goal are among the most observed semantic features for any given trajectory. This effect is less pronounced in the random agent.}
    \label{fig:appendix_co_occurrence_grey}
\end{figure}

\begin{figure}
    \centering
    \begin{subfigure}{0.45\linewidth}
        \centering
        \includegraphics[width=\textwidth]{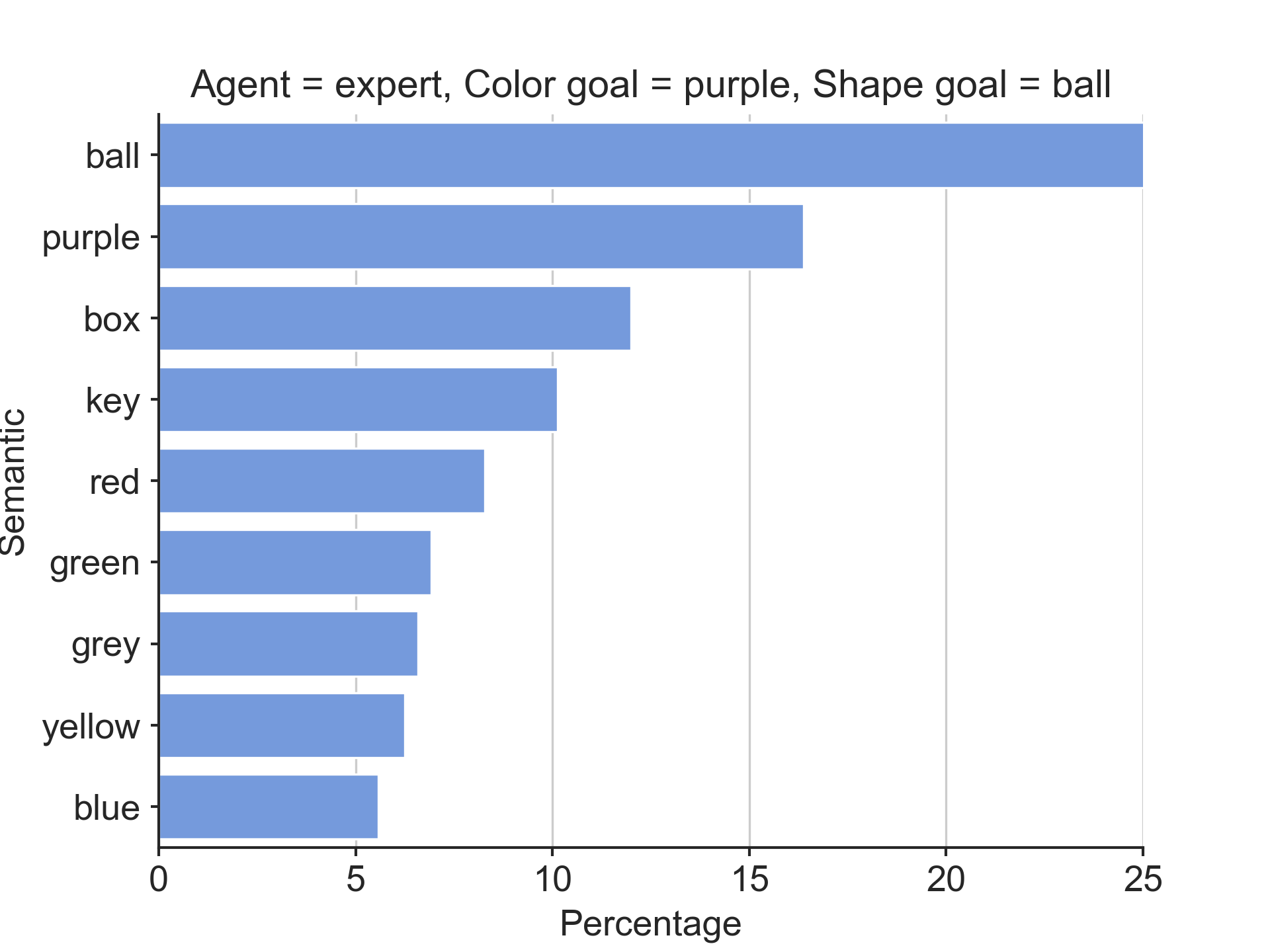}
    \end{subfigure}
    \begin{subfigure}{0.45\linewidth}
        \centering
        \includegraphics[width=\textwidth]{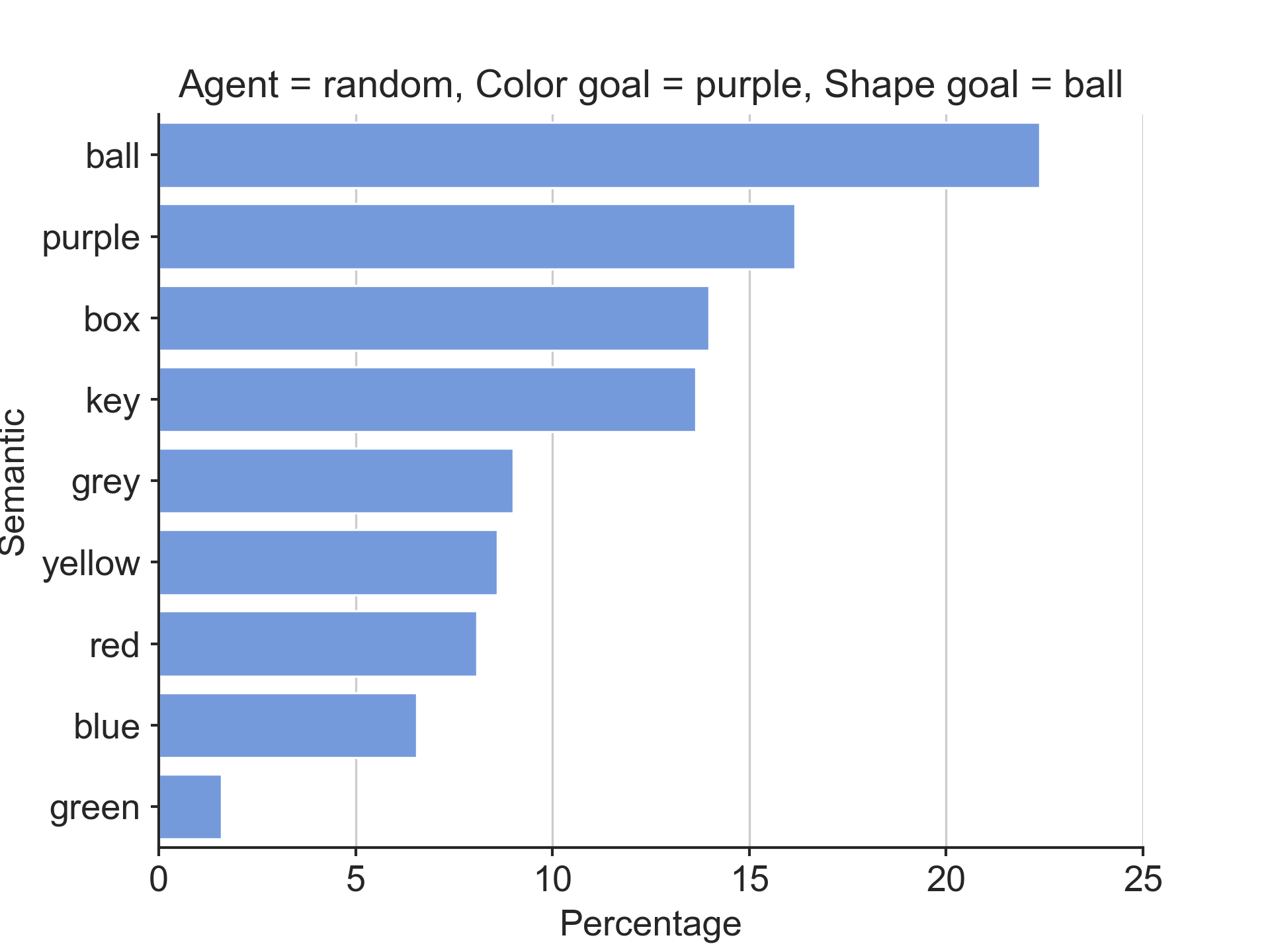}
    \end{subfigure}
    \begin{subfigure}{0.45\linewidth}
        \centering
        \includegraphics[width=\textwidth]{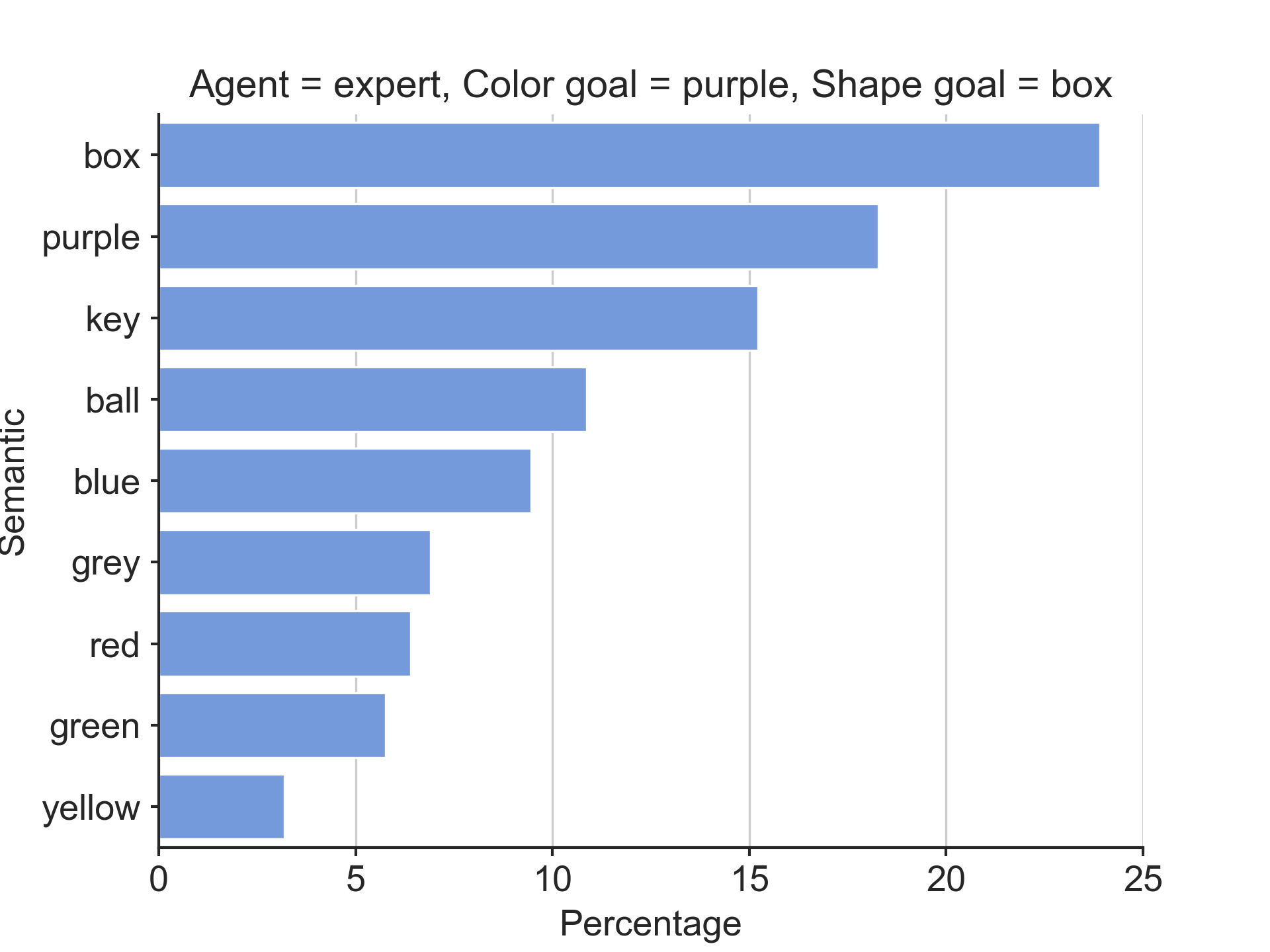}
    \end{subfigure}
    \begin{subfigure}{0.45\linewidth}
        \centering
        \includegraphics[width=\textwidth]{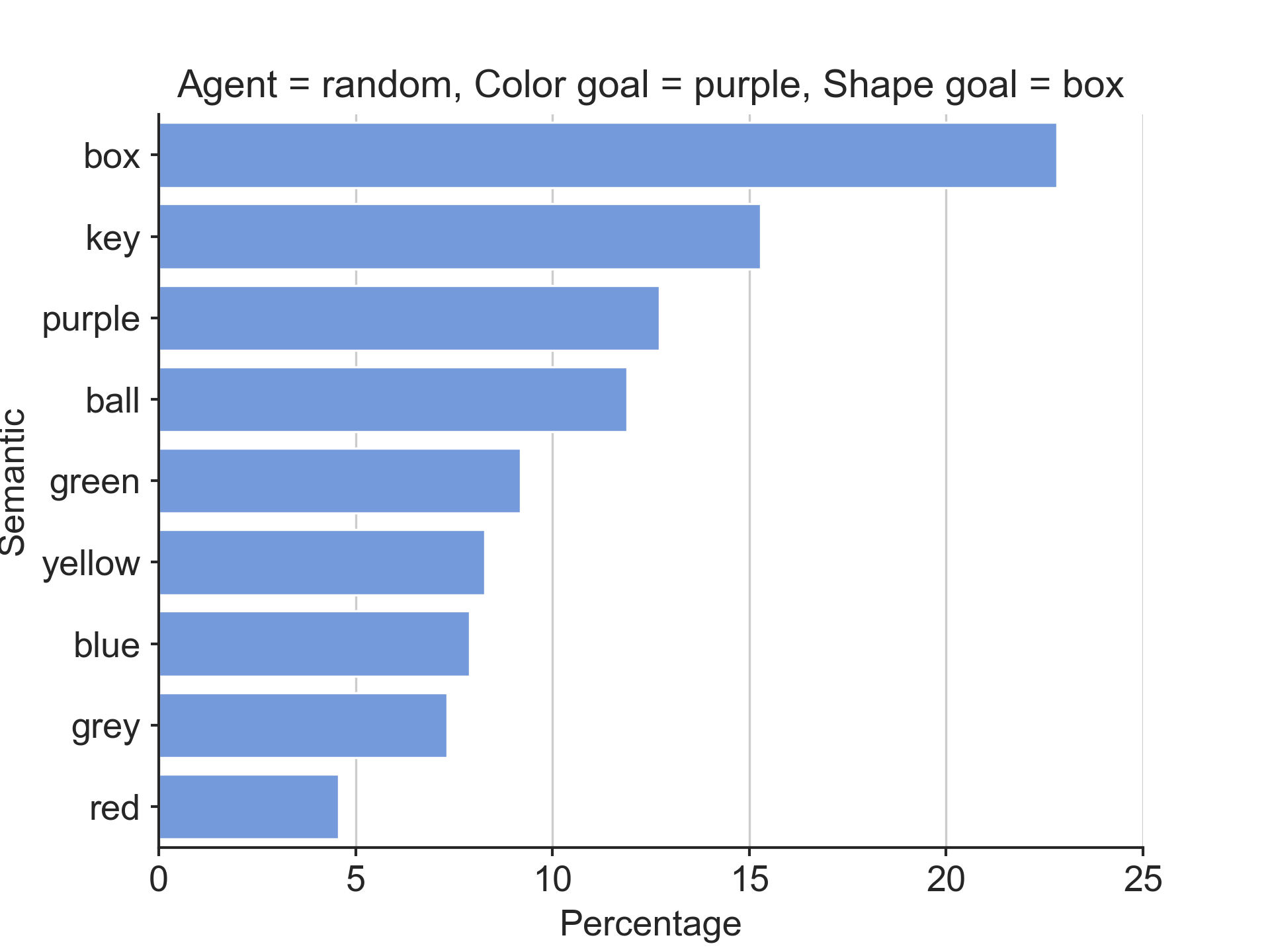}
    \end{subfigure}
    \begin{subfigure}{0.45\linewidth}
        \centering
        \includegraphics[width=\textwidth]{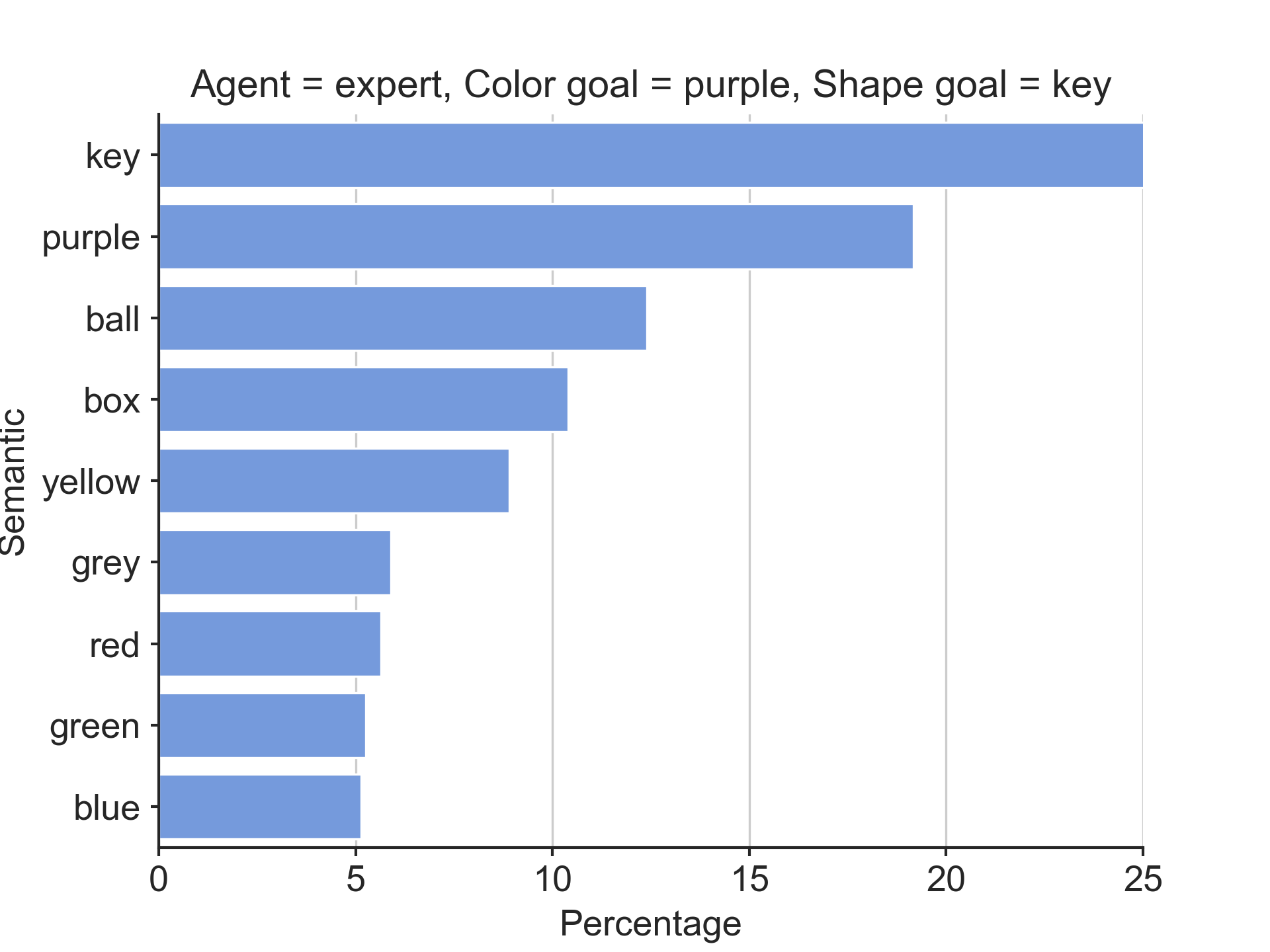}
    \end{subfigure}
    \begin{subfigure}{0.45\linewidth}
        \centering
        \includegraphics[width=\textwidth]{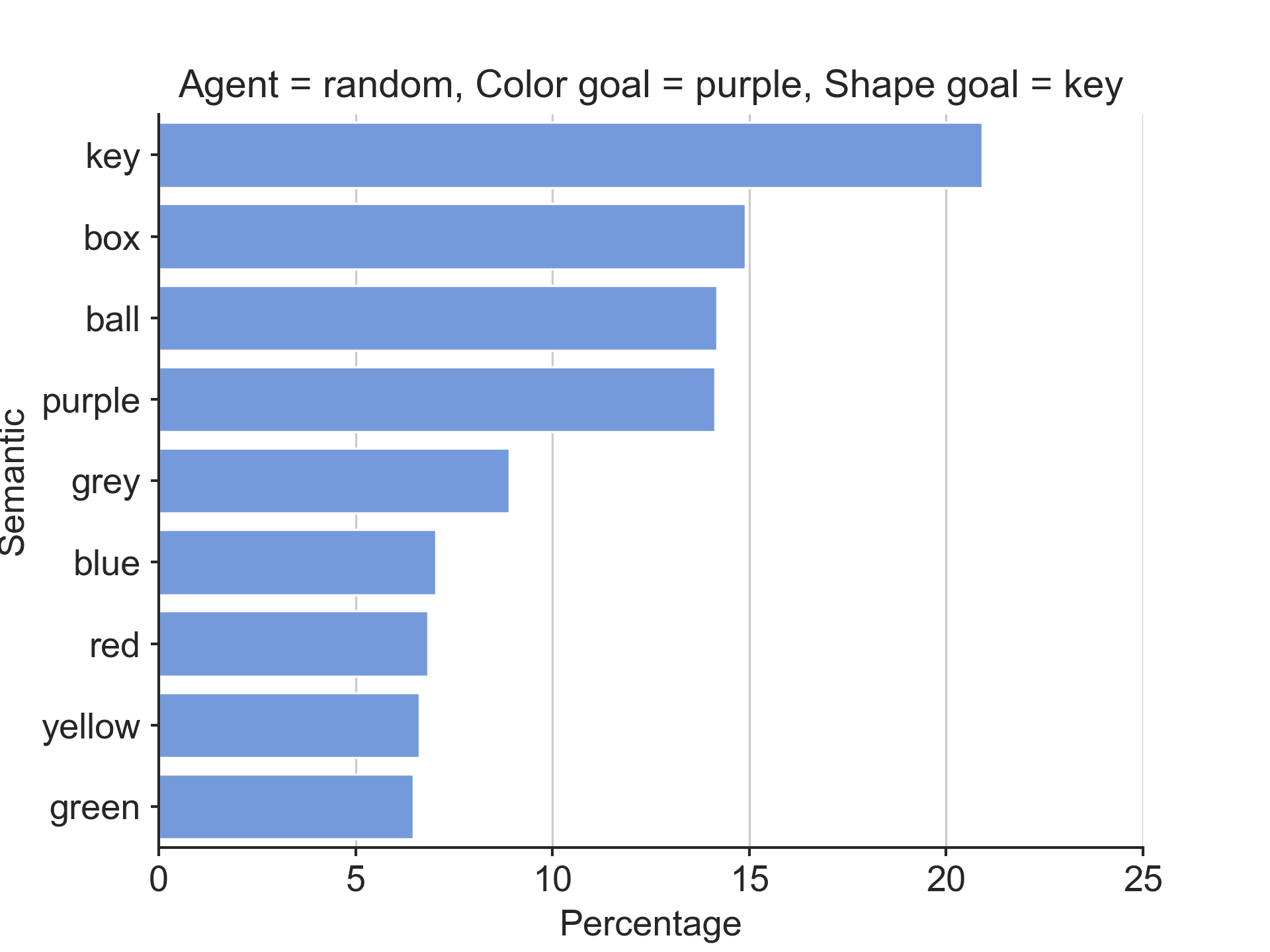}
    \end{subfigure}
    \begin{subfigure}{0.45\linewidth}
        \centering
        \includegraphics[width=\textwidth]{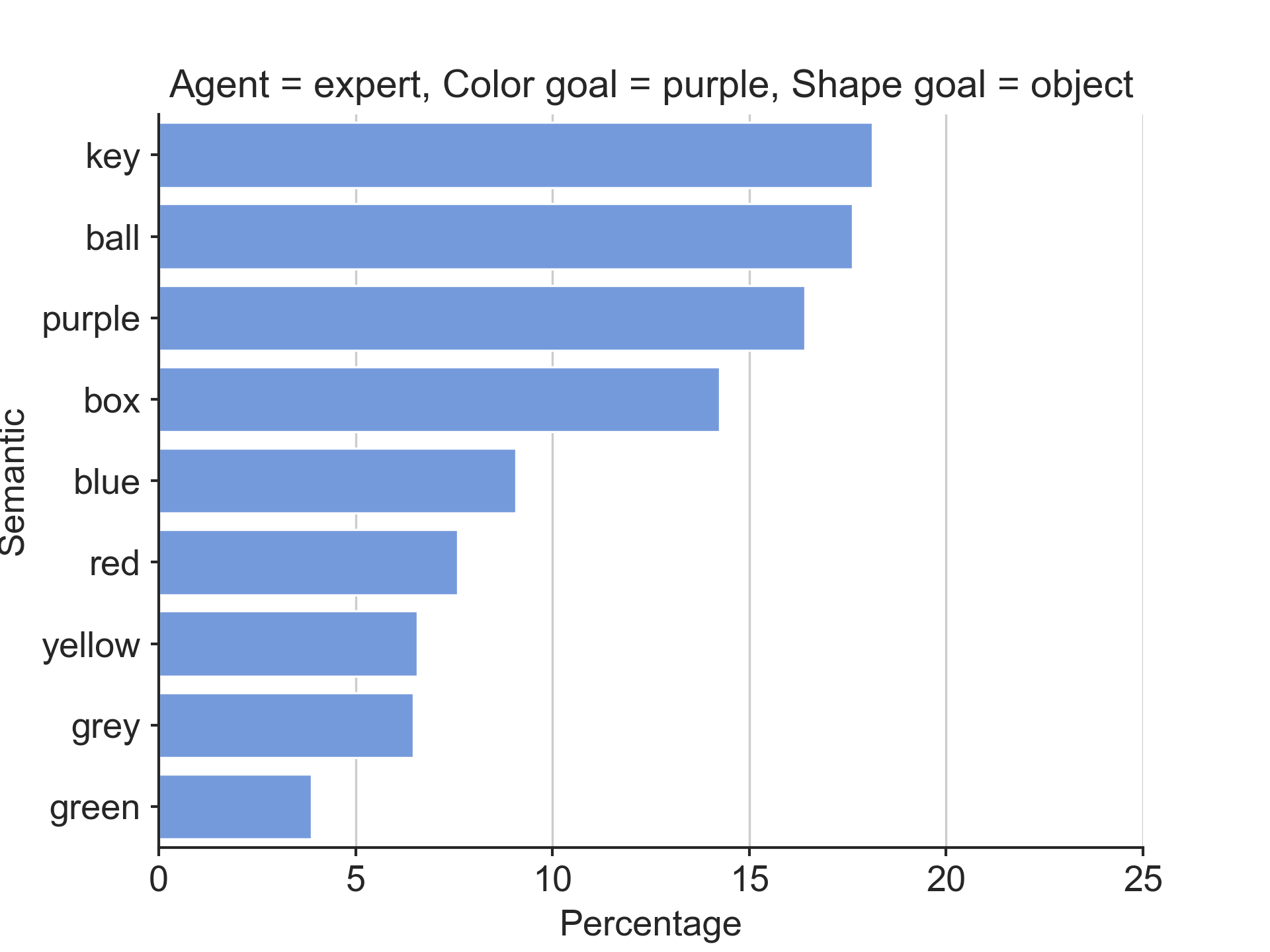}
        \caption{Expert agent}
    \end{subfigure}
    \begin{subfigure}{0.45\linewidth}
        \centering
        \includegraphics[width=\textwidth]{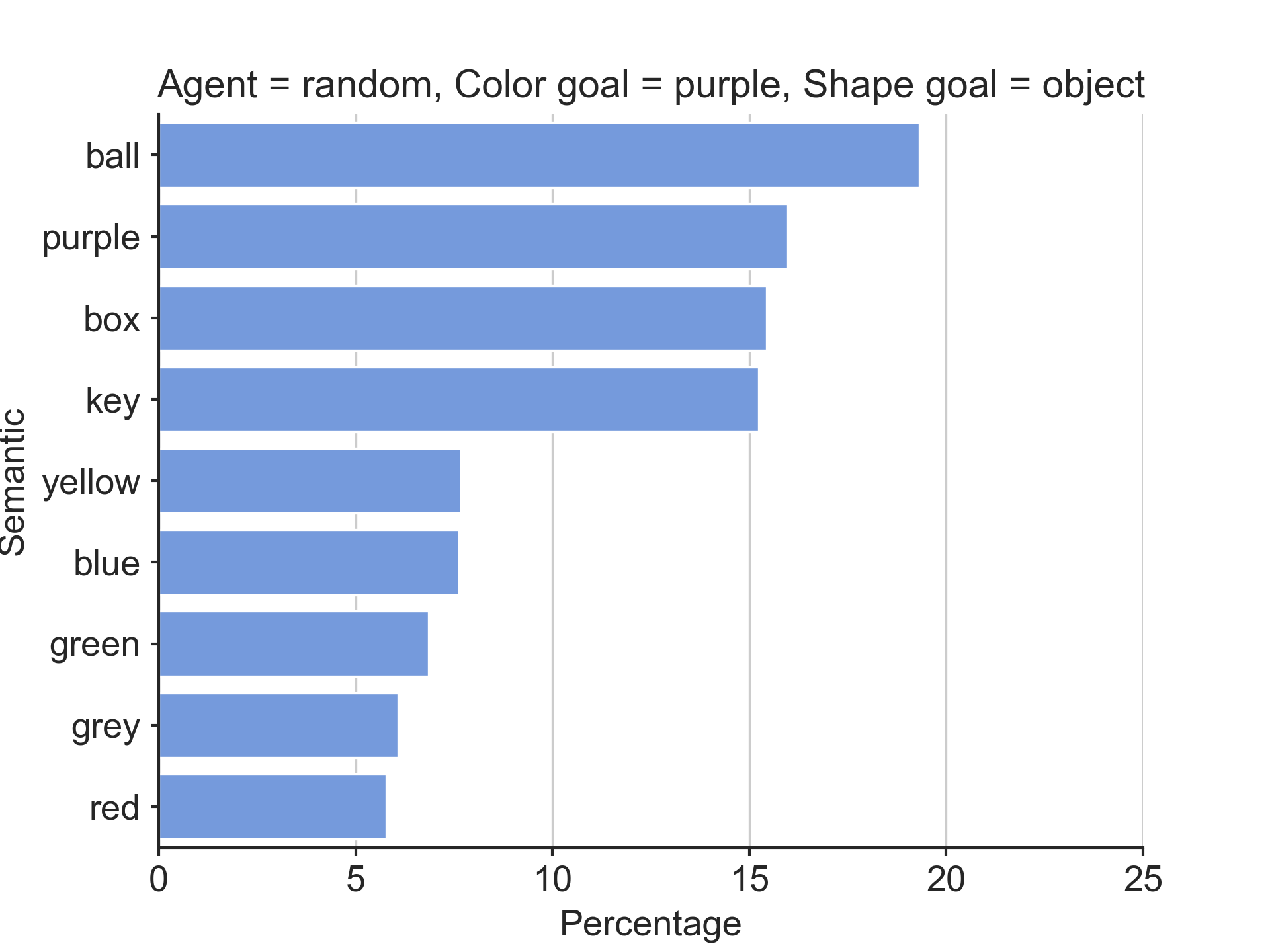}
        \caption{Random agent}
    \end{subfigure}
    \caption{\textbf{Left:} Trajectories for the purple color goal from BabyAI's built-in expert agent which always reaches the goal. \textbf{Right:} Random agent trajectories. In both cases the semantics of the goal are among the most observed semantic features for any given trajectory. This effect is less pronounced in the random agent.}
    \label{fig:appendix_co_occurrence_purple}
\end{figure}

\begin{figure}
    \centering
    \begin{subfigure}{0.45\linewidth}
        \centering
        \includegraphics[width=\textwidth]{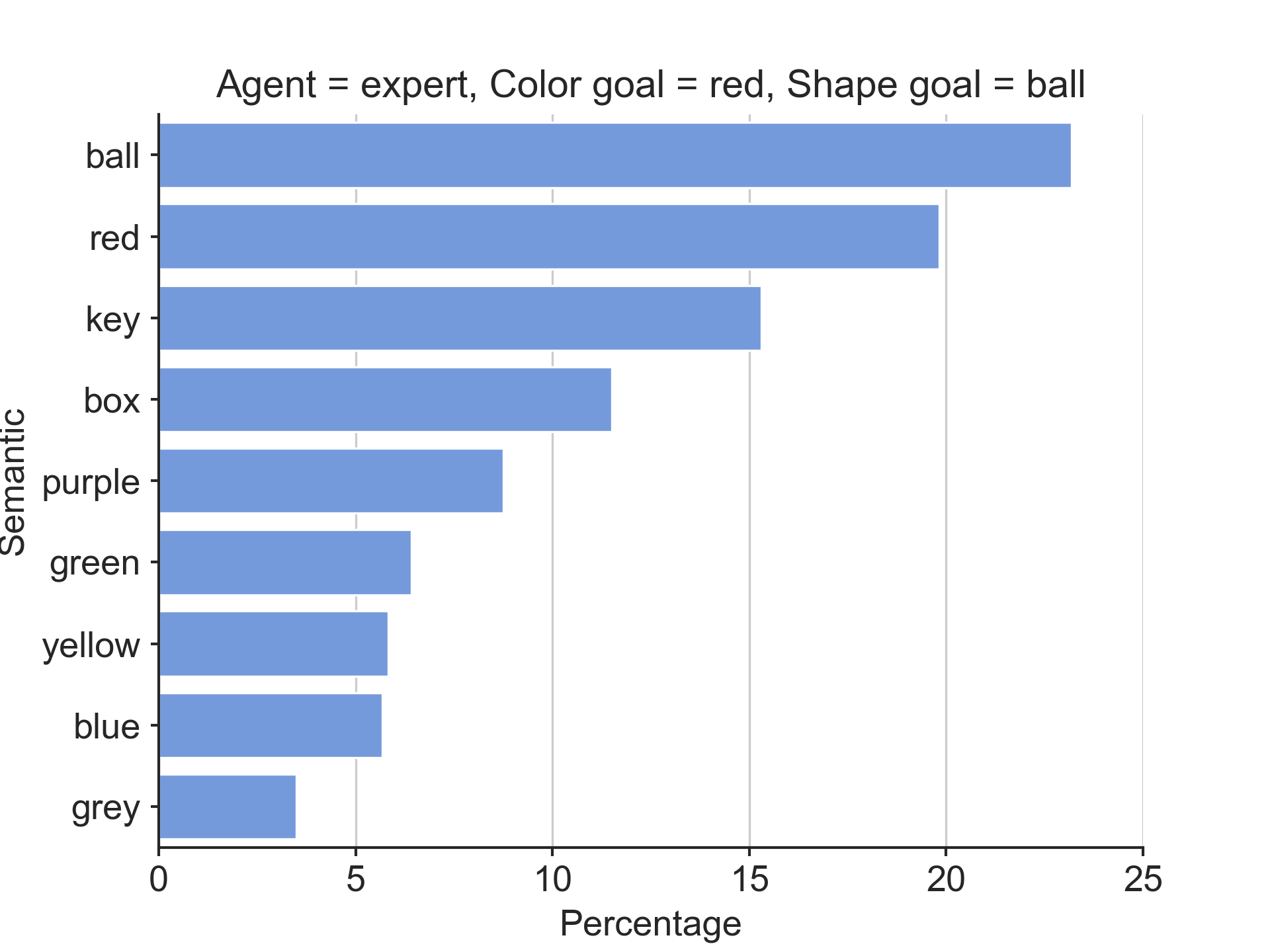}
    \end{subfigure}
    \begin{subfigure}{0.45\linewidth}
        \centering
        \includegraphics[width=\textwidth]{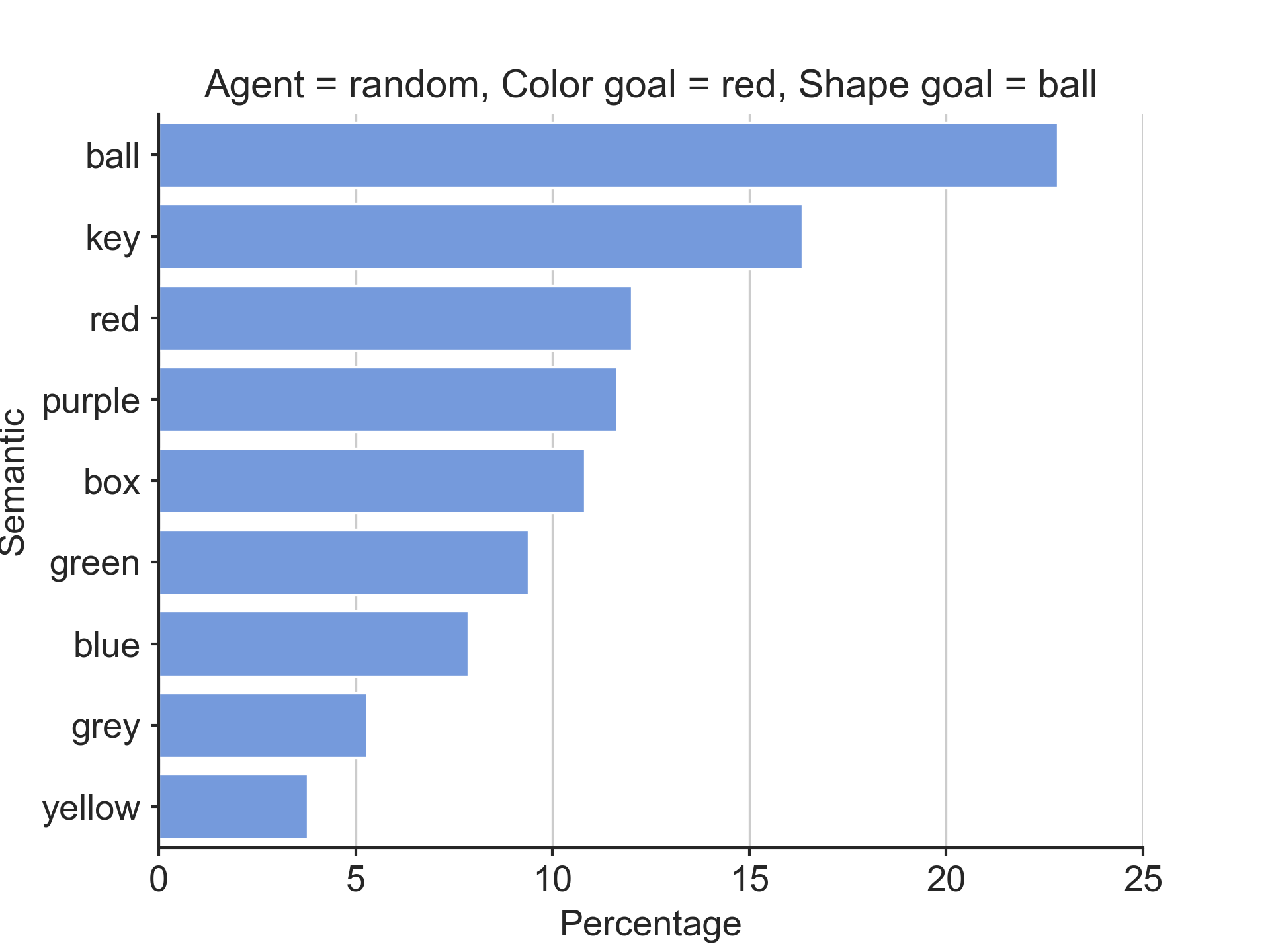}
    \end{subfigure}
    \begin{subfigure}{0.45\linewidth}
        \centering
        \includegraphics[width=\textwidth]{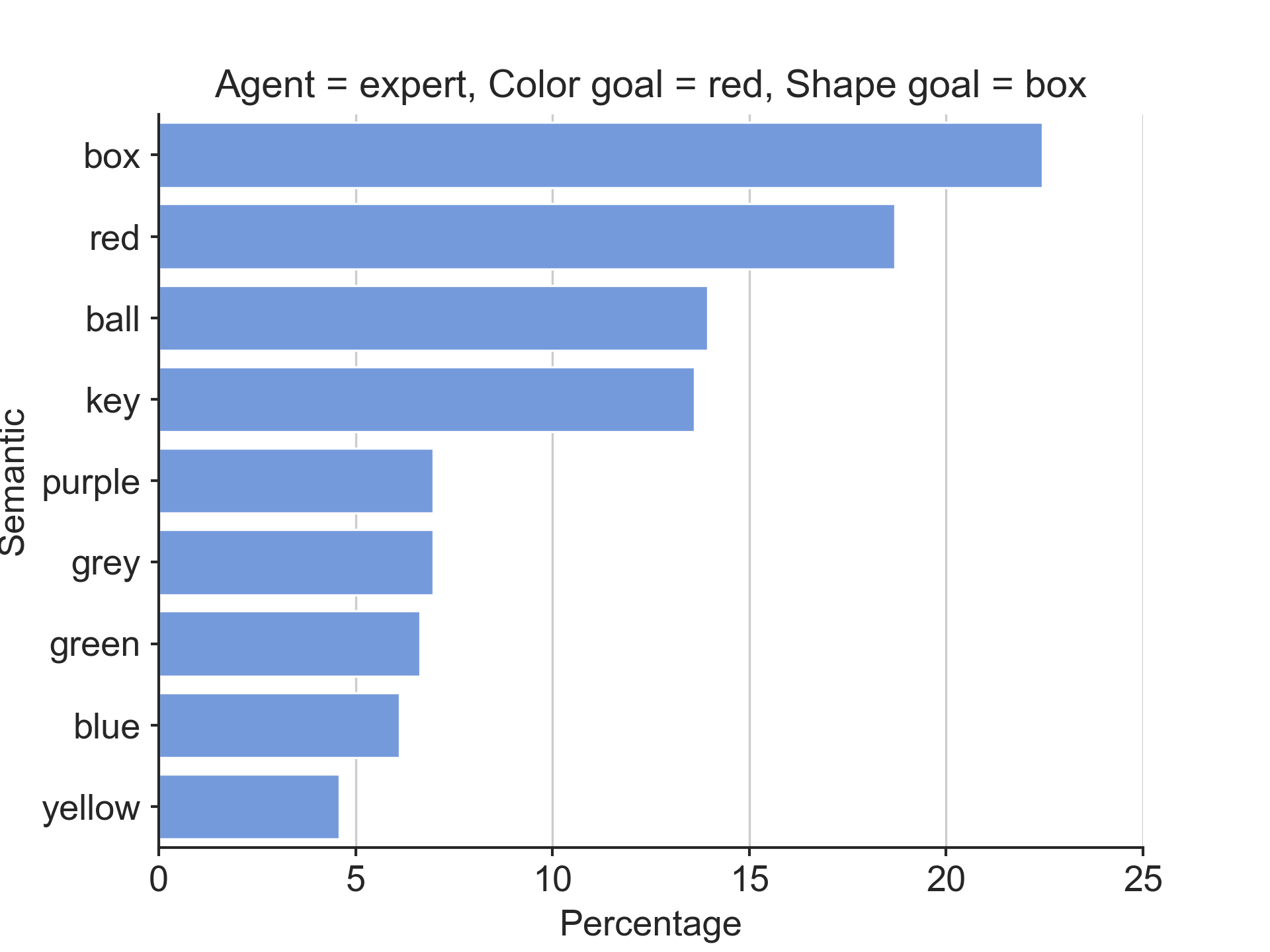}
    \end{subfigure}
    \begin{subfigure}{0.45\linewidth}
        \centering
        \includegraphics[width=\textwidth]{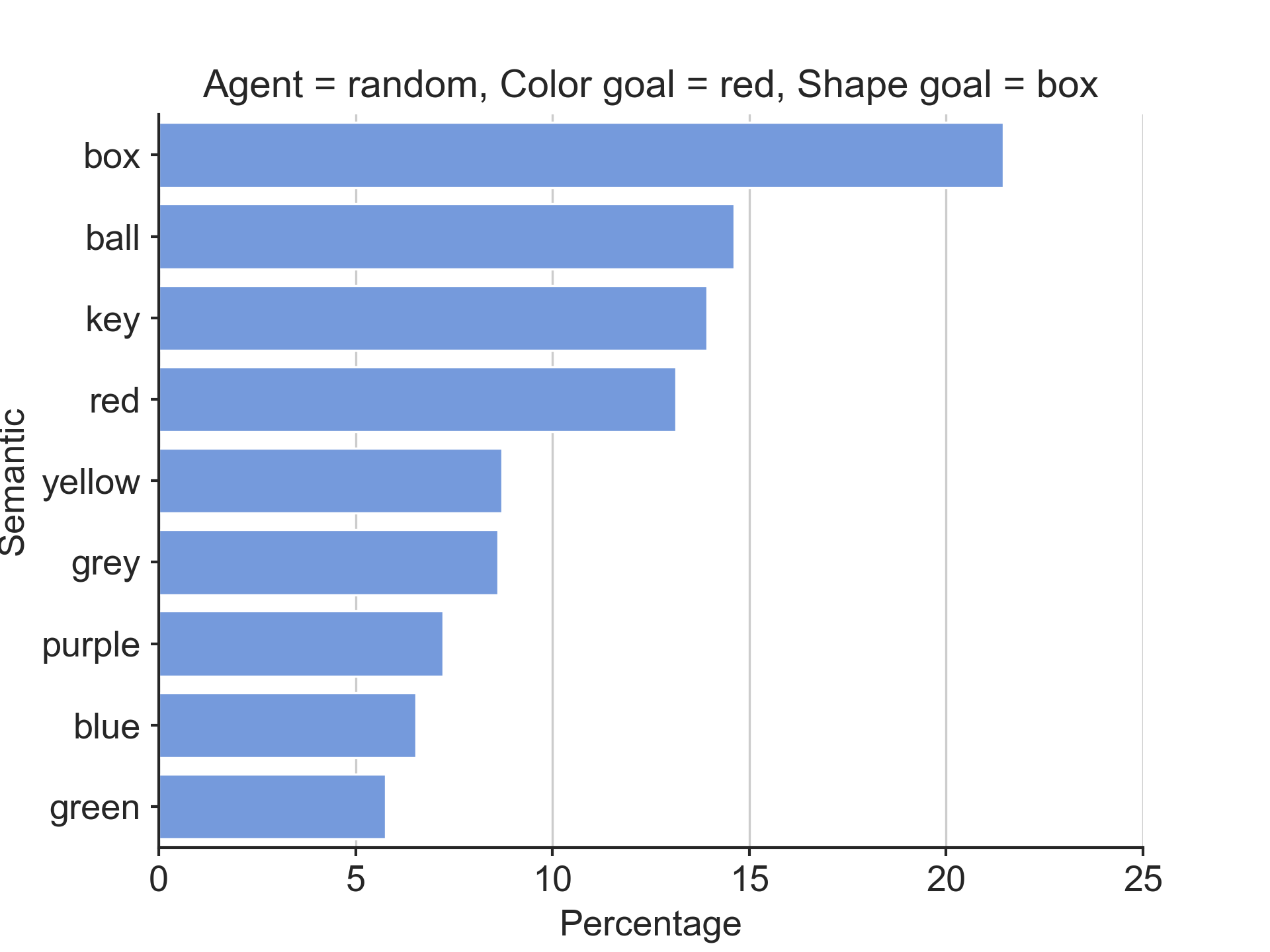}
    \end{subfigure}
    \begin{subfigure}{0.45\linewidth}
        \centering
        \includegraphics[width=\textwidth]{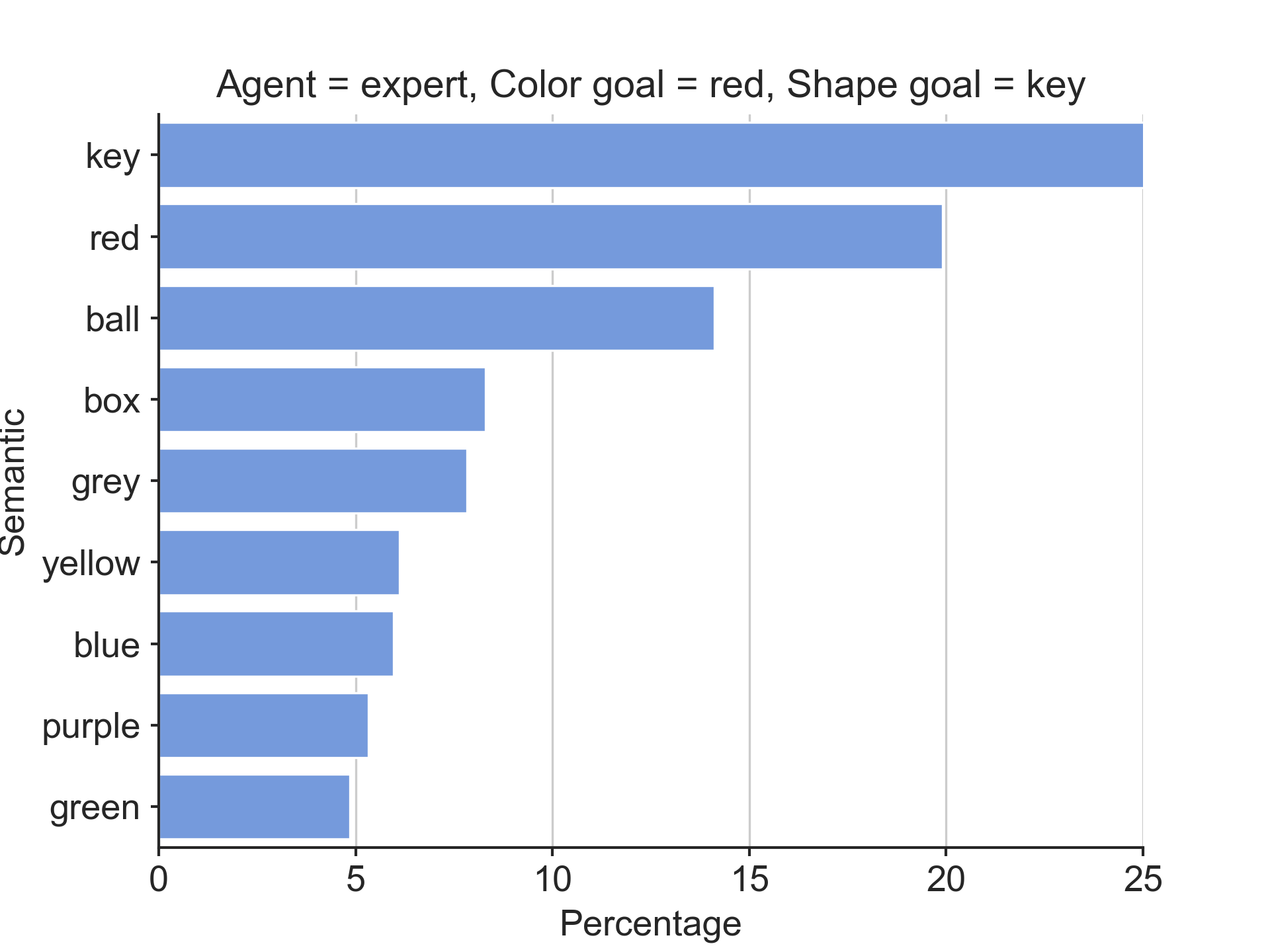}
    \end{subfigure}
    \begin{subfigure}{0.45\linewidth}
        \centering
        \includegraphics[width=\textwidth]{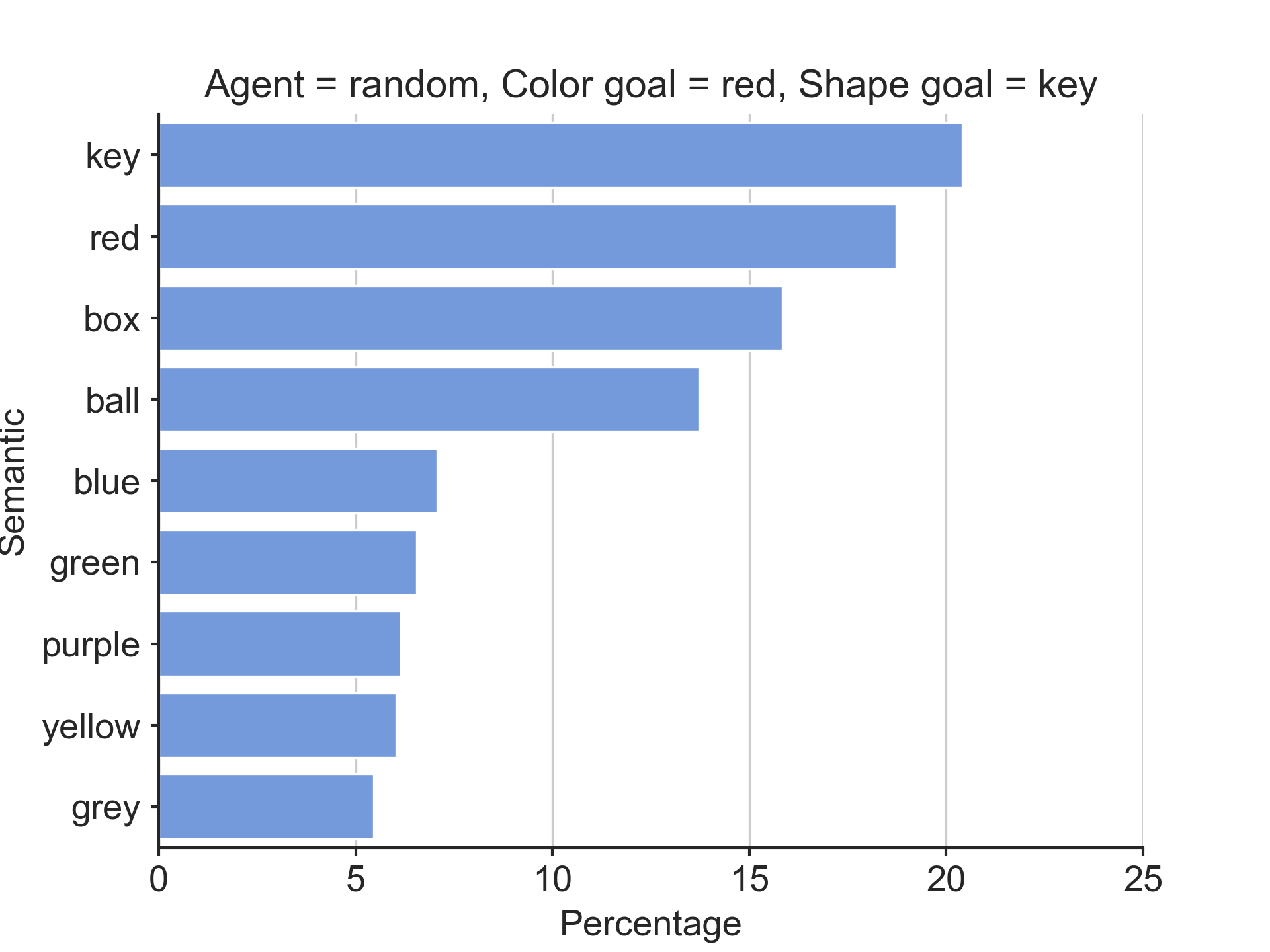}
    \end{subfigure}
    \begin{subfigure}{0.45\linewidth}
        \centering
        \includegraphics[width=\textwidth]{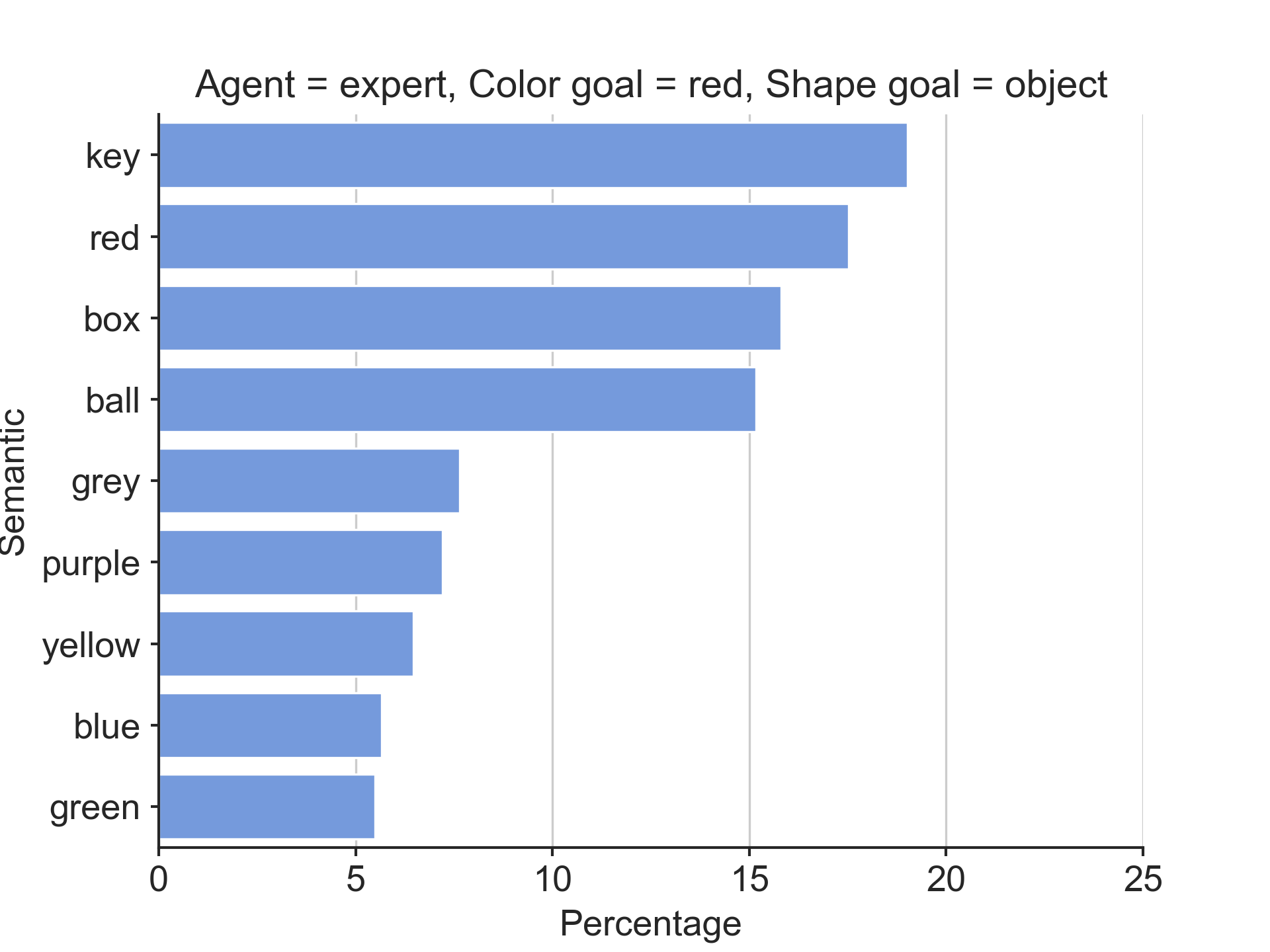}
        \caption{Expert agent}
    \end{subfigure}
    \begin{subfigure}{0.45\linewidth}
        \centering
        \includegraphics[width=\textwidth]{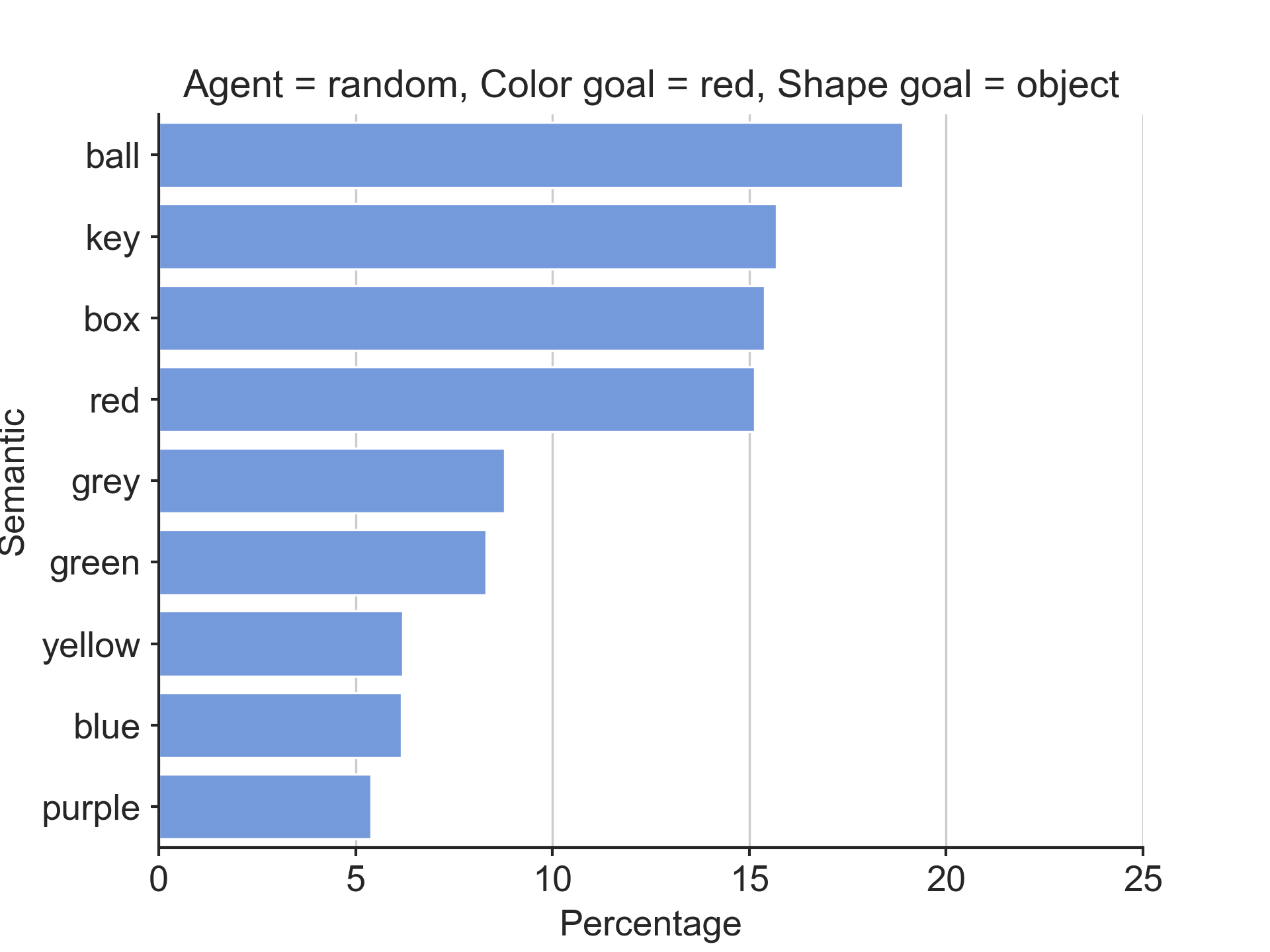}
        \caption{Random agent}
    \end{subfigure}
    \caption{\textbf{Left:} Trajectories for the red color goal from BabyAI's built-in expert agent which always reaches the goal. \textbf{Right:} Random agent trajectories. In both cases the semantics of the goal are among the most observed semantic features for any given trajectory. This effect is less pronounced in the random agent.}
    \label{fig:appendix_co_occurrence_red}
\end{figure}

\begin{figure}
    \centering
    \begin{subfigure}{0.45\linewidth}
        \centering
        \includegraphics[width=\textwidth]{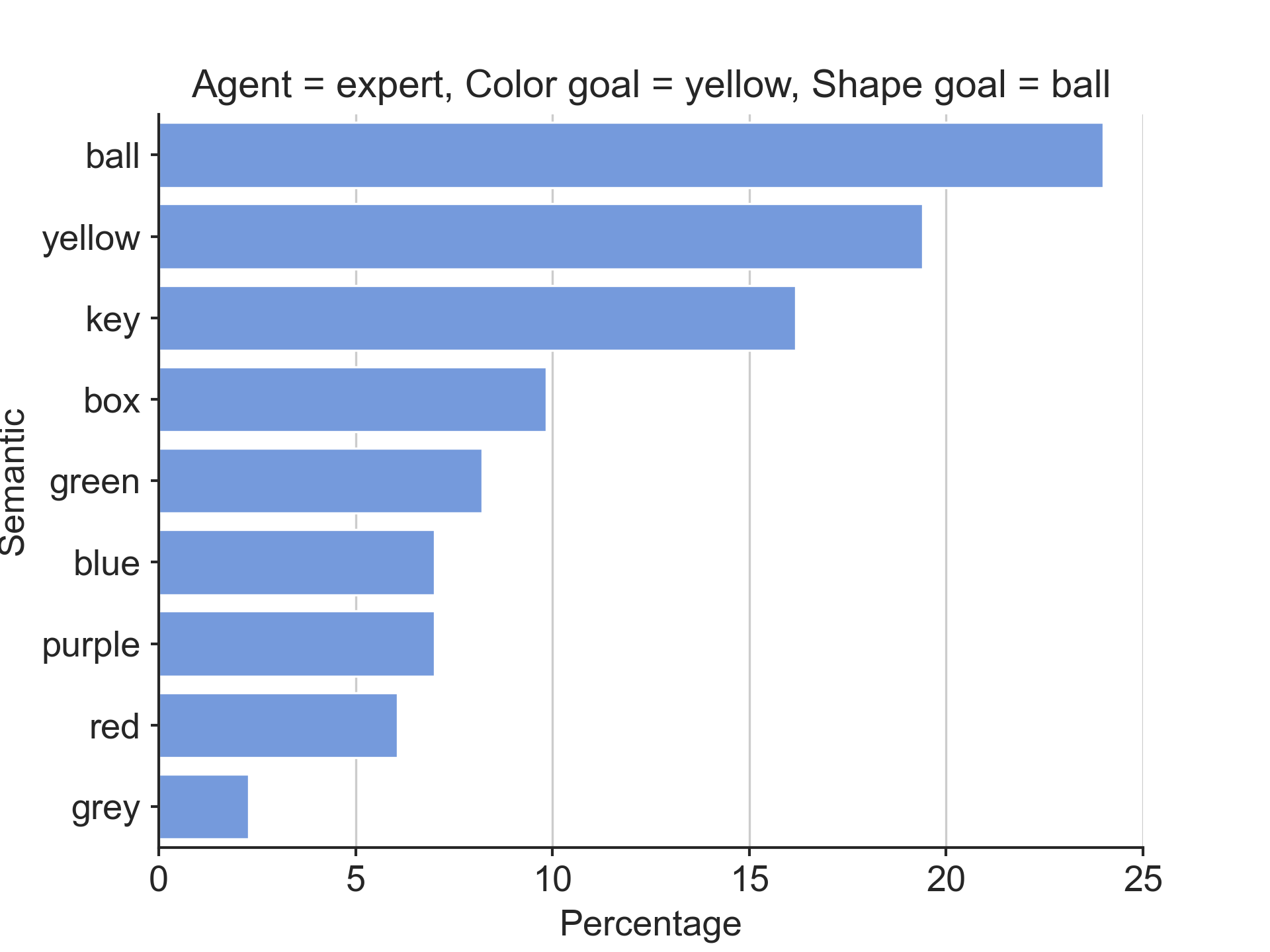}
    \end{subfigure}
    \begin{subfigure}{0.45\linewidth}
        \centering
        \includegraphics[width=\textwidth]{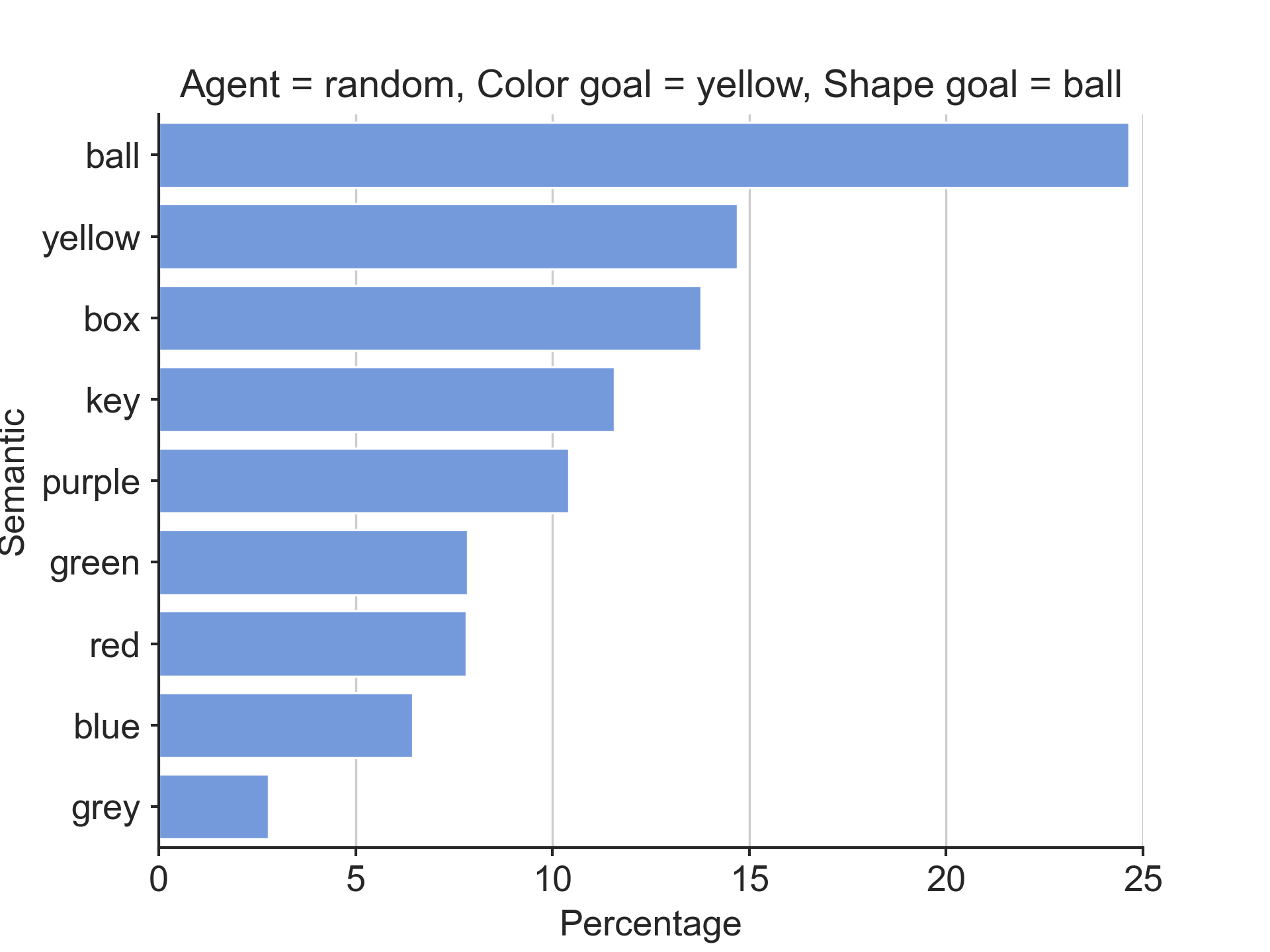}
    \end{subfigure}
    \begin{subfigure}{0.45\linewidth}
        \centering
        \includegraphics[width=\textwidth]{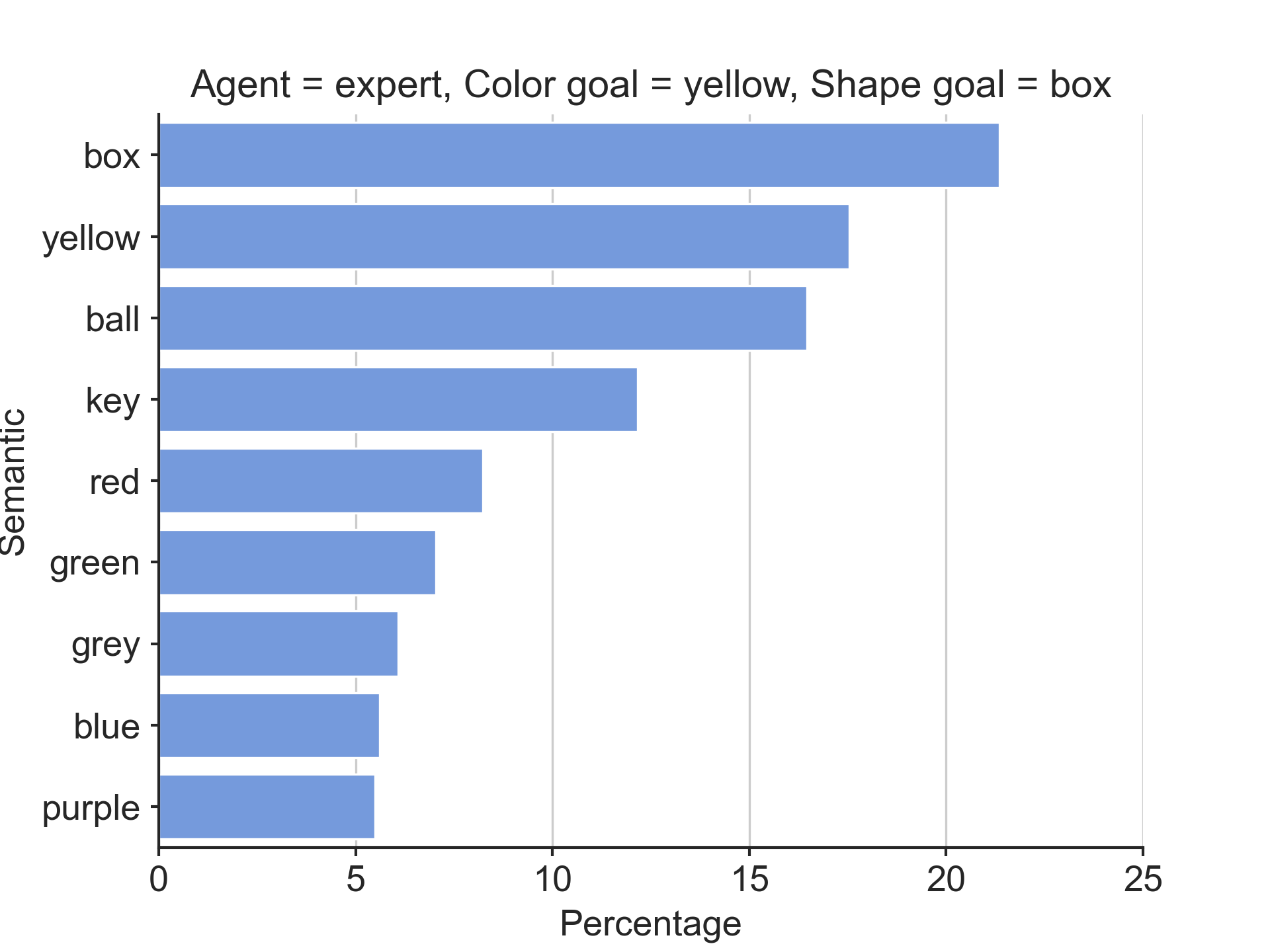}
    \end{subfigure}
    \begin{subfigure}{0.45\linewidth}
        \centering
        \includegraphics[width=\textwidth]{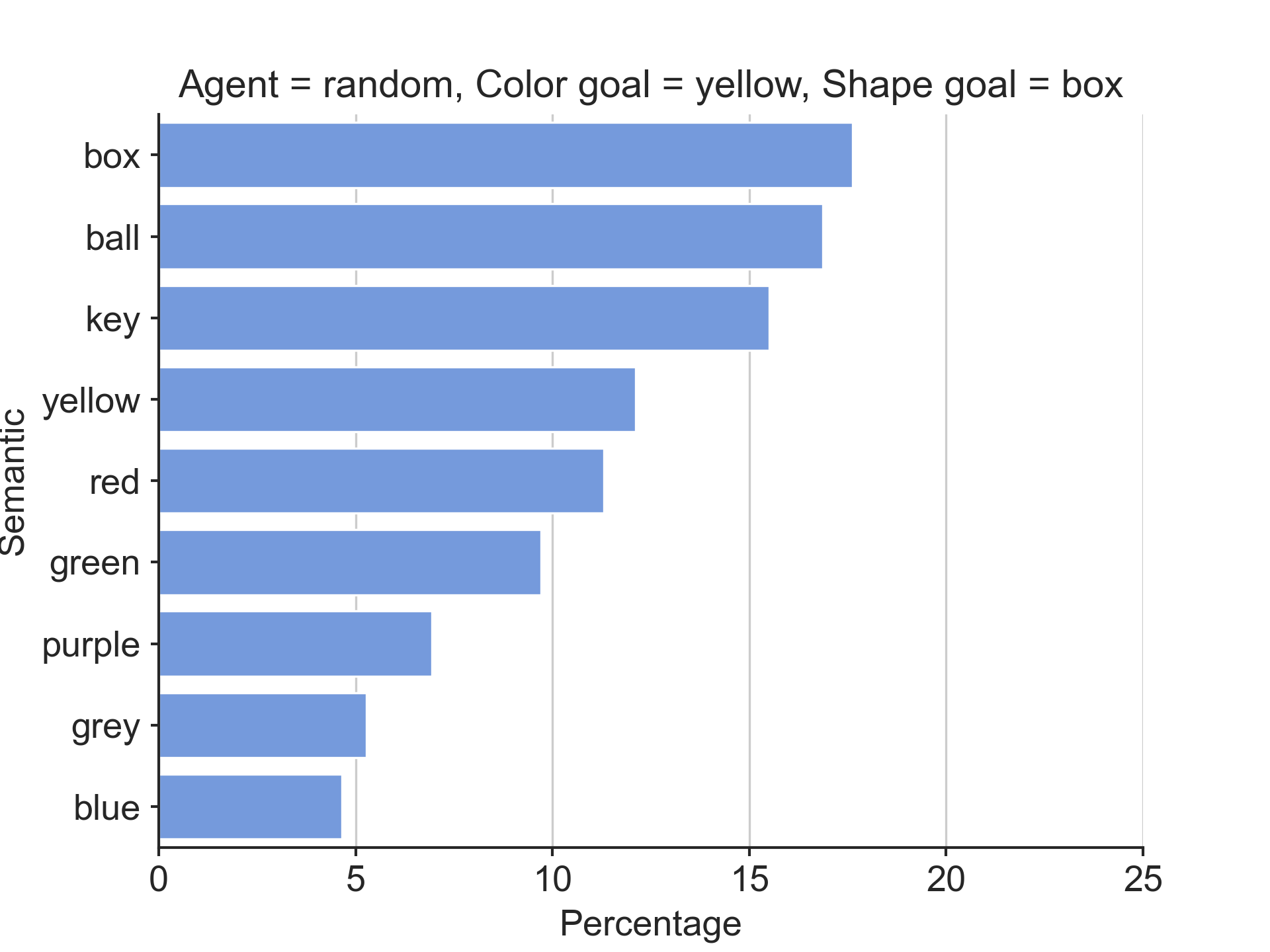}
    \end{subfigure}
    \begin{subfigure}{0.45\linewidth}
        \centering
        \includegraphics[width=\textwidth]{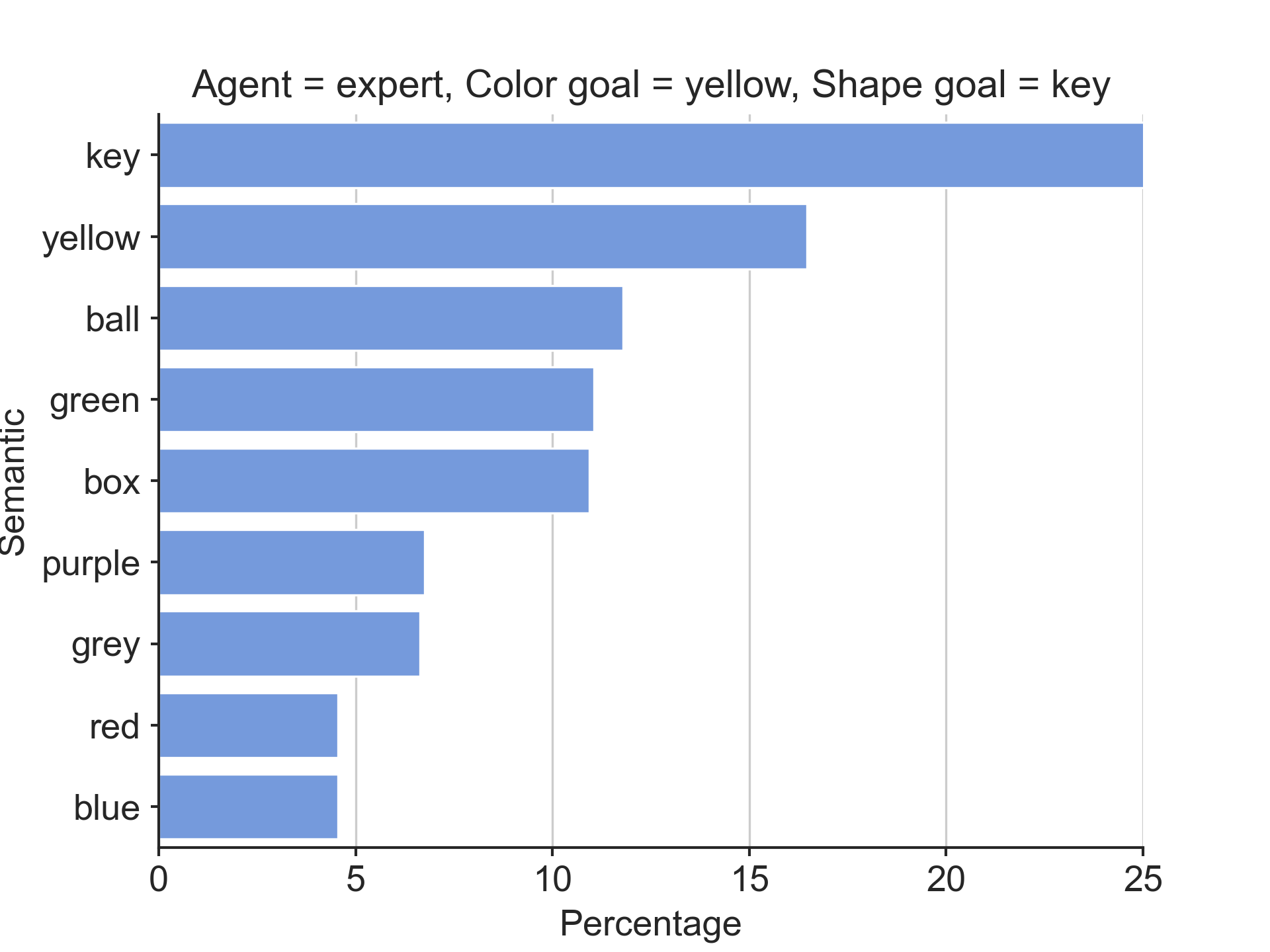}
    \end{subfigure}
    \begin{subfigure}{0.45\linewidth}
        \centering
        \includegraphics[width=\textwidth]{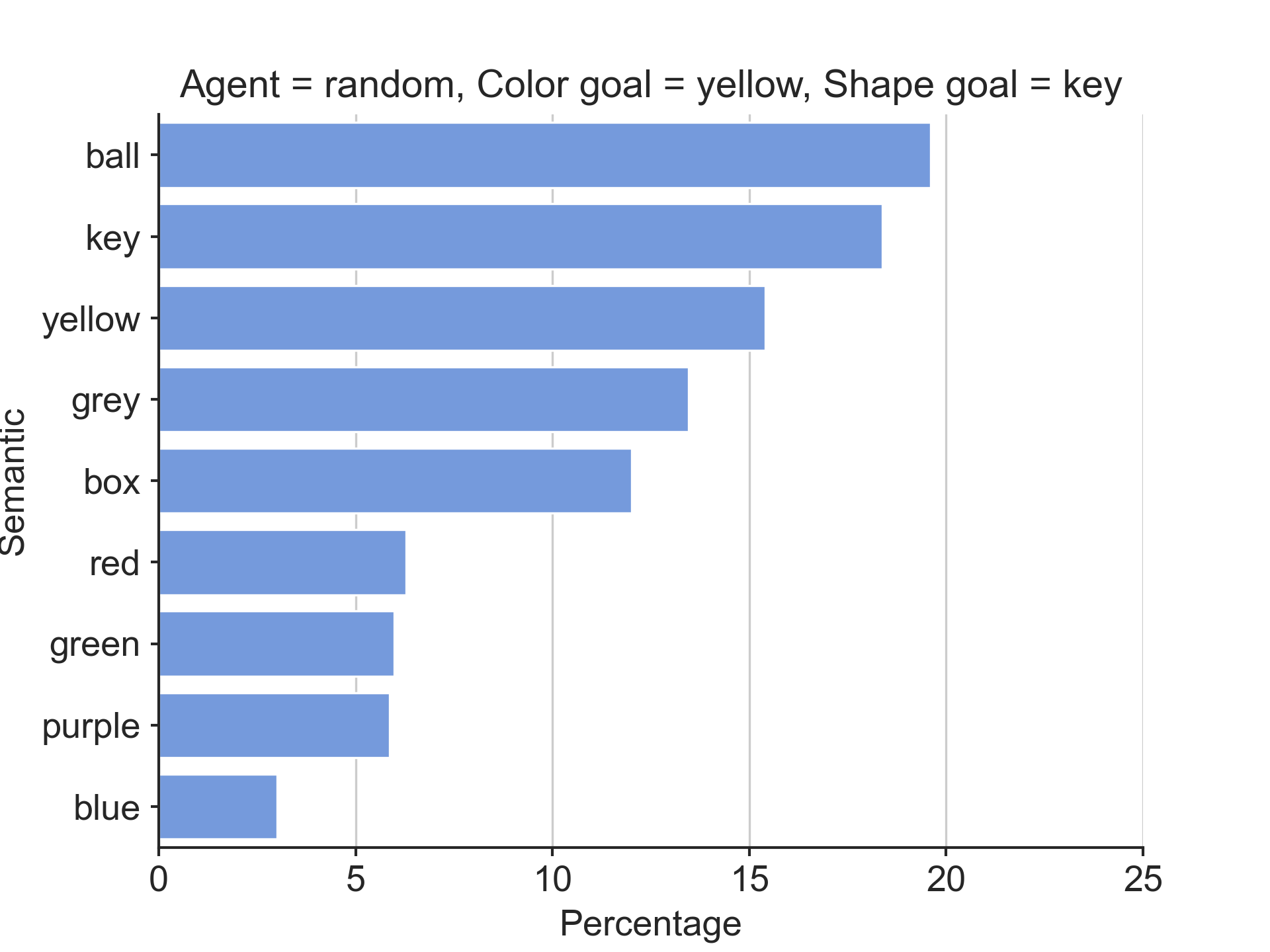}
    \end{subfigure}
    \begin{subfigure}{0.45\linewidth}
        \centering
        \includegraphics[width=\textwidth]{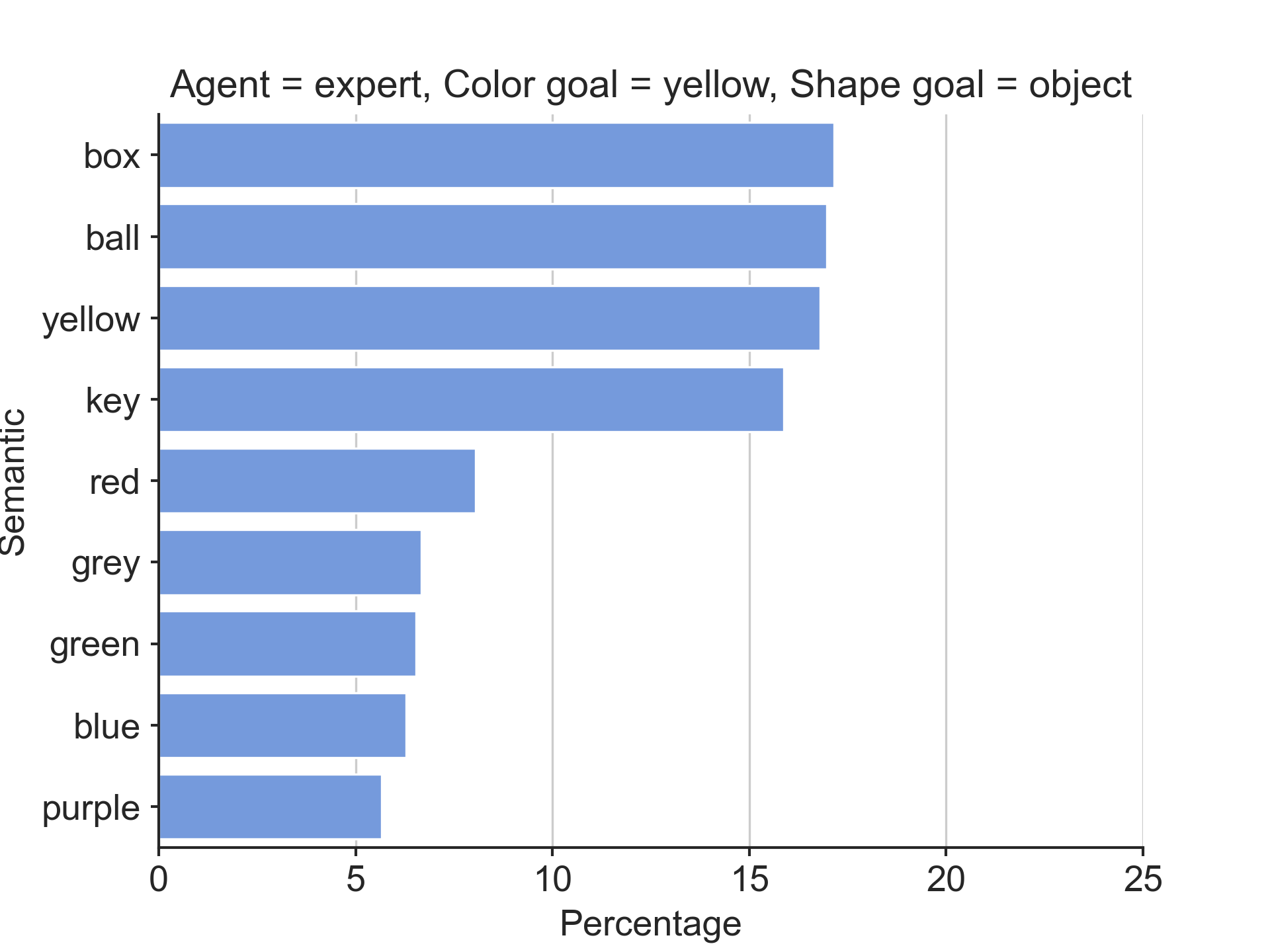}
        \caption{Expert agent}
    \end{subfigure}
    \begin{subfigure}{0.45\linewidth}
        \centering
        \includegraphics[width=\textwidth]{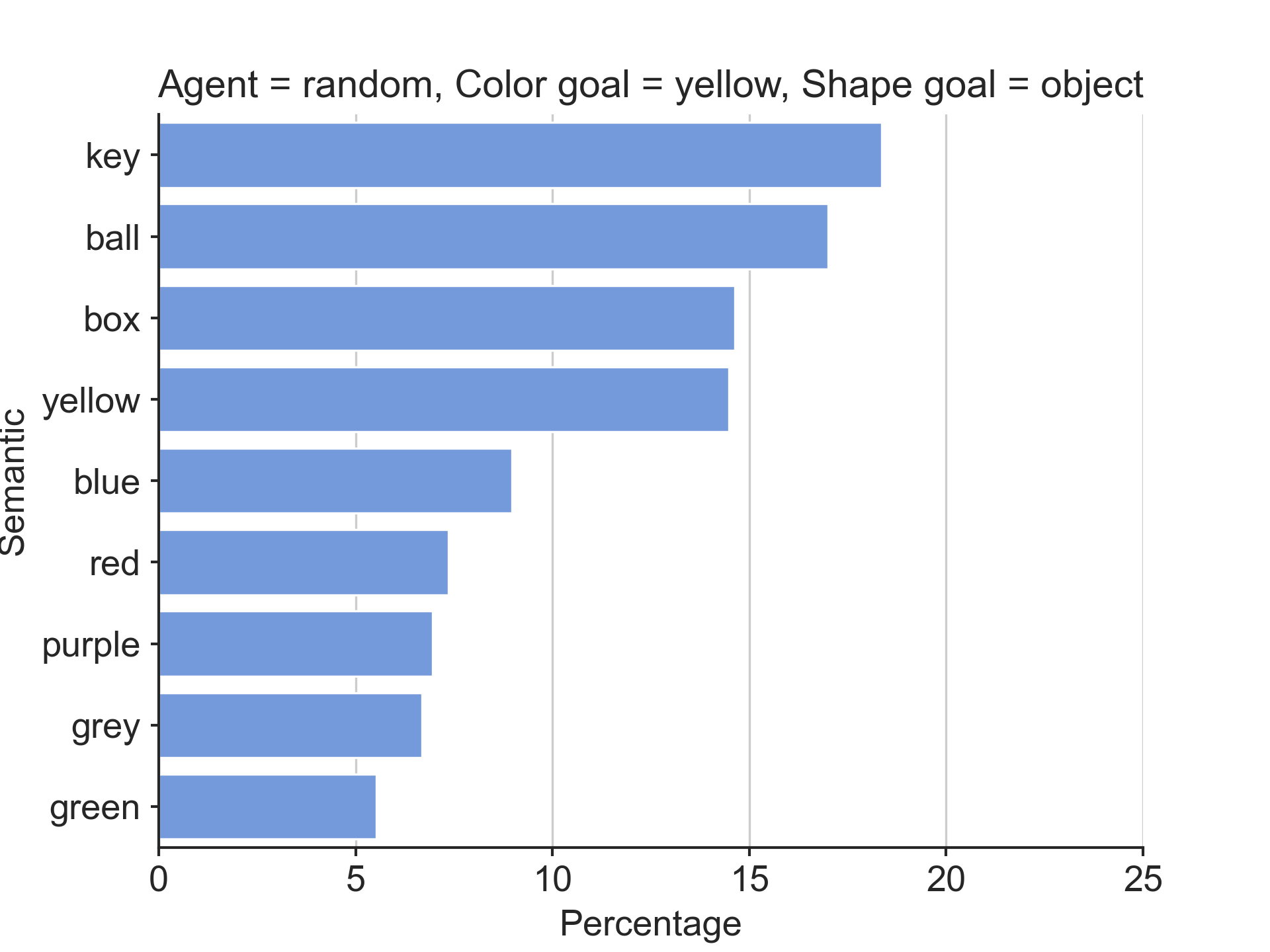}
        \caption{Random agent}
    \end{subfigure}
    \caption{\textbf{Left:} Trajectories for the yellow color goal from BabyAI's built-in expert agent which always reaches the goal. \textbf{Right:} Random agent trajectories. In both cases the semantics of the goal are among the most observed semantic features for any given trajectory. This effect is less pronounced in the random agent.}
    \label{fig:appendix_co_occurrence_yellow}
\end{figure}

Here, we provide further evidence that the linguistic goal description aligns with the observed semantics across different permutations of colour goal and shape goal. As described in the main body text, this is consistent across both agents, but more visible in the expert agent. Figures \ref{fig:appendix_co_occurrence_none}, \ref{fig:appendix_co_occurrence_green}, \ref{fig:appendix_co_occurrence_blue}, \ref{fig:appendix_co_occurrence_grey}, \ref{fig:appendix_co_occurrence_purple}, \ref{fig:appendix_co_occurrence_red}, and \ref{fig:appendix_co_occurrence_yellow} present histograms for each combination of colour and shape goal for both the expert and random agent. We note that semantic co-occurrence, while prevalent, is not always perfectly the case. For instance, Figure \ref{fig:appendix_co_occurrence_blue}, the most commonly observed semantic in the expert agent trajectories for the blue colour and ball shape was "box", as opposed to the expected "ball" semantic.

\newpage

% Blue
\FloatBarrier
%\todo[inline]{uncomment below}
% None

%\begin{comment}
% Green

% Grey

% Purple

% Red

% Yellow
%\end{comment}

\end{document}